\pdfoutput=1
\documentclass[11pt]{article}

\usepackage[final]{acl}

\usepackage{times}
\usepackage{latexsym}
\usepackage{times}
\usepackage{latexsym}
\usepackage{graphicx}
\usepackage{multicol}
\usepackage{multirow}
\usepackage{adjustbox}
\usepackage{comment}
\usepackage{caption}
\usepackage{mathtools}
\usepackage{amsmath}
\usepackage{algorithm}
\usepackage{float}
\usepackage{url}
\usepackage{hyperref}
\usepackage[justification=centering]{caption}
\usepackage{algpseudocode}
\algblock{Input}{EndInput}
\algnotext{EndInput}
\algblock{Paramters}{EndParameter}
\algnotext{EndParameter}

\usepackage[T1]{fontenc}

\usepackage[utf8]{inputenc}

\usepackage{microtype}
\usepackage{subcaption}

\usepackage{inconsolata}

\usepackage{graphicx}

\newcommand{\saransh}[1]{\textcolor{black}{#1}}
\newcommand{\sharma}[1]{\textcolor{black}{#1}}

\definecolor{M7B}{HTML}{060ba5} 
\definecolor{M7B-After}{HTML}{dcadff} 
\definecolor{MulT}{HTML}{f6a800}
\definecolor{MulT-After}{HTML}{FFFF84}
\definecolor{SDIF}{HTML}{017716} 
\definecolor{SDIF-After}{HTML}{85f7d7} 

\definecolor{te}{HTML}{0173B2} 
\definecolor{tva}{HTML}{f7cae9}
\definecolor{au}{HTML}{029E73}
\definecolor{vi}{HTML}{DE8F05}
\definecolor{ta}{HTML}{CC78BC} 
\definecolor{tv}{HTML}{D55E00} 
\definecolor{av}{HTML}{a36c0d}

\title{Text Takes Over: A Study of Modality Bias in Multimodal Intent Detection}

\author{$^1$Ankan Mullick \qquad $^2$Saransh Sharma \qquad $^3$Abhik Jana \qquad $^1$Pawan Goyal \\ \texttt{ankanm@kgpian.iitkgp.ac.in} \qquad \texttt{sarsharma@adobe.com} \\ \texttt{abhikjana@iitbbs.ac.in} \qquad \texttt{pawang@cse.iitkgp.ac.in}\\ $^1$IIT Kharagpur, India \qquad $^2$Adobe Research, India \\ $^3$IIT Bhubaneswar, India}

\begin{document}
\maketitle
\begin{abstract}
The rise of multimodal data, integrating text, audio, and visuals, has created new opportunities for studying multimodal tasks such as intent detection. This work investigates the effectiveness of Large Language Models (LLMs) and non-LLMs, including text-only and multimodal models, in the multimodal intent detection task. Our study reveals that Mistral-7B, a text-only LLM, outperforms most competitive multimodal models by approximately 9\% on MIntRec-1 and 4\% on MIntRec2.0 dataset. This performance advantage comes from a strong textual bias in these datasets, where over 90\% of the samples require textual input, either alone or in combination with other modalities, for correct classification. We confirm the modality bias of these datasets via human evaluation, too. Next, we propose a framework to debias the datasets, and upon debiasing, more than 70\% of the samples in MIntRec-1 and more than 50\% in MIntRec2.0 get removed, resulting in significant performance degradation across all models, with smaller multimodal fusion models being the most affected with an accuracy drop of over 50 - 60\%. Further, we analyze the context-specific relevance of different modalities through empirical analysis. Our findings highlight the challenges posed by modality bias in multimodal intent datasets and emphasize the need for unbiased datasets to evaluate multimodal models effectively.  We release both the code and the dataset used for this work.\footnote{\url{https://github.com/Text-Takes-Over-EMNLP-2025/MultiModal-Intent-EMNLP-2025}}

\end{abstract}

\section{Introduction}
\begin{figure}[t]
\centering
\vspace{-2mm}
  \fbox{\includegraphics[width=0.8\columnwidth]{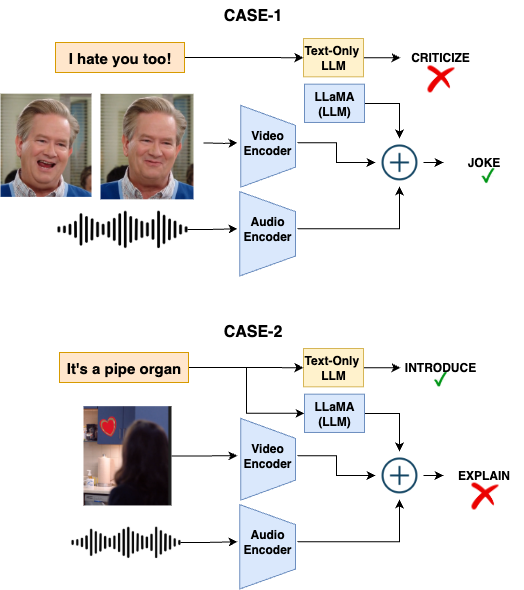}}
  \vspace{-2mm}
  \caption {Role of different modalities in Multimodal Intent Detection Task}
  \vspace{-8mm}
    \label{fig:intro-pic1}
\end{figure}


\sharma{Intent detection refers to classifying user queries into predefined categories and is a key component of systems such as chatbots, search engines, virtual assistants, and robots. While traditional approaches mainly rely on text, multimodal intent detection extends this by incorporating audio and visual inputs. By considering speech, tone, and facial expressions, these systems capture both what is said and how it is conveyed, making them essential for building more accurate, natural, and context-sensitive applications.}

For example, the phrase \textbf{``I hate you too!''} may seem negative if judged only by the text and be classified as \textit{Criticize}. However, if the tone or expression is playful, the actual intent is a \textit{Joke}, which can only be identified by considering non-verbal signals (Case-1 in Fig.~\ref{fig:intro-pic1}). On the other hand, there are cases where multimodal inputs may confuse the model. In Case-2, the phrase \textbf{``It's a pipe organ''} is correctly interpreted as \textit{Introduce} using text alone, but the multimodal model misclassifies it as \textit{Explain}. Hence, a multimodal intent detection model must dynamically balance multiple modalities and prioritize which one or ensemble approach to utilize.
To evaluate such a model, we would require gold-standard datasets having no bias towards a particular modality, and preparing such a dataset from scratch is a cumbersome job. 


In this paper, we pick existing gold-standard datasets, detect the bias towards any specific modality and finally propose a framework to debias the datasets. To validate our claim of bias in existing datasets, we conduct a comprehensive analysis on the performance of state-of-the-art models for the intent detection task using both multimodal and unimodal architectures on both biased and unbiased datasets. 
We evaluate large language models (LLMs) such as LLaMA, GPT-4o, and Claude, as well as non-LLMs like SDIF, BERT, and MAG-BERT, to assess their ability to capture nuanced intents from text, audio, and visual inputs. Our study includes models ranging from 100 million to 15 billion parameters and explores different training strategies, including full training, parameter-efficient fine-tuning, and zero-/few-shot learning. Our results show that \textbf{Mistral-7B}, a text-based LLM, achieves the best performance on both datasets, even surpassing GPT-4o. We evaluate the models on two benchmark multimodal intent detection datasets: MIntRec-1~\cite{10.1145/3503161.3547906} (\textbf{M-1}) and MIntRec2.0~\cite{zhang2024mintrec} (\textbf{M-2.0}). Our findings reveal a strong bias towards the textual modality: \textbf{70\%} of the samples in M-1 and about \textbf{55\%} of the samples in M-2.0 can be classified correctly using text alone, with text being necessary for correct classification in more than \textbf{90\%} of cases. 
We then re-evaluate the models using the debiased datasets, effectively removing heavily biased intent categories (6 in M-1 and 3 in M-2.0) and reducing the proportion of biased samples.

Re-evaluation of models' performance reveals that the initial results were \textbf{heavily inflated} due to textual bias in the dataset. After debiasing, all models struggle significantly on the task, with 60\%  performance drop observed in smaller multimodal models and more than 50\% performance drop in textual models. 
Human annotation also verifies that around \textbf{80\%} samples in both datasets are textually biased, which supports our findings. 




 To summarize, our contributions are as follows: (1) We perform a comprehensive evaluation of models with various sizes and fusion techniques for multimodal intent detection and find that the Mistral-7B textual model outperforms others, despite the task’s multimodal nature; (2) We identify the presence and impact of modality bias in existing datasets, analyzing how it affects model performance and capabilities. (3) We propose a debiasing framework that systematically detects and removes biased samples, enabling the construction of more balanced datasets for bias-free intent detection. (4) Through re-evaluation after debiasing, we identify a significant gap between the expected and actual performance of models, highlighting the need for robust 
 multimodal intent detection frameworks.

\section{Related Works}

The task of intent detection has been explored through various approaches discussed below:

\noindent \textbf{Generalized Intent Detection Approaches:} Researchers investigate several methodologies for intent detection, including few-shot techniques~\citep{xia2021incremental}, zero-shot approaches~\citep{xia2018zero}, and clustering-based frameworks~\citep{mullick2022framework}. Other studies~\citep{wang2020active,mu2017streaming,mu2017classification} focus on identifying new intents, often formulated as an outlier detection problem.  


\noindent \textbf{Intent Detection Applications:} There are various domain-specific applications of intent detection task in specialized contexts \citep{mullick2022evaluation, mullick2022fine,mullick2022framework,mullick2023exploring, mullick2023novel,mullick2024pointer}. For instance, \citep{mullick2023intent,mullick2024intent} and ~\citep{bharti2020medbot}(MedBot) focus on medical queries and conversational systems, while \citep{patwa2021overview} identifies fake and hostile posts related to COVID-19 in multilingual context. Unlike these works, our study investigates the comparative roles of unimodal and multimodal frameworks in intent detection.


\noindent \textbf{Existing Datasets:} Several datasets for intent detection tasks have been proposed. Text-only datasets include SNIPS~\citep{coucke2018snips}, ATIS~\citep{tur2010left}, HWU~\citep{liu2019benchmarking}, and BANKING~\citep{casanueva2020efficient}. For multimodal contexts, \citep{kruk2019integrating} curate MDID dataset with binary labels for Instagram posts' images, while \citep{9961847} proposes the EmoInt dataset featuring 32 emotions and 15 empathetic intents, although it is not publicly available. MIntRec-1~\citep{10.1145/3503161.3547906} and MIntRec2.0~\citep{zhang2024mintrec} datasets focus on multimodal intent detection, offering diverse contexts for evaluating models that integrate text, audio, and visual information. MIntRec2.0 focuses more on out-of-scope intents. These two datasets are used in our experiments. 


\noindent \textbf{Modality Bias:} Modality bias happens when models rely too much on one input, like image or video, while ignoring others. It is common in VQA and video reasoning tasks~\citep {guo2023modality,10572012}. ~\citep{10477535} uses counterfactual reasoning to reduce both visual and language shortcuts , while video-dropout training helps with missing frames but can cause audio bias~\citep{10658284}. Multimodal intent detection introduces additional challenges, as the dominant modality can vary, and there is no ground-truth modality supervision. But, the impact of modality bias in this setting remains largely unexplored.


\section{Dataset}


\sharma{We use two publicly available benchmark multimodal intent detection datasets: MIntRec-1 \cite{10.1145/3503161.3547906} \textbf{[M-1]} and MIntRec2.0 \cite{zhang2024mintrec} \textbf{[M-2.0]}, both under fine-grained settings with text, video, and audio modalities. M-1 contains more than 2000 annotated samples with 20 intent labels and was the first benchmark dataset for multimodal intent classification. M-2.0 is a larger extension, consisting of more than 9000 samples across 30 intent categories, making it the most comprehensive publicly available dataset in this domain. Further dataset details are provided in Table \ref{tab:datasets-stats} and Appendix section \ref{sec:multimodal-dataset}. }

\begin{table}[h]
\centering
\vspace{-1mm}
\begin{adjustbox}{width=0.75\linewidth}
\begin{tabular}{|c|c|c|c|c|}
\hline
\textbf{Dataset}   & \textbf{Train} & \textbf{Dev} & \textbf{Test}  & \textbf{\# Intents} \\
\hline
\textbf{M-1}  & 1334  & 445  & 445 & 20 \\
\hline
\textbf{M-2.0} & 6165  & 1106 & 2033 & 30 \\
\hline
\end{tabular}
\end{adjustbox}
\vspace{-2mm}
\caption{Statistics of Multimodal Intent Datasets}
\vspace{-6mm}
\label{tab:datasets-stats}

\end{table}

\section{Approach}
We consider a diverse set of models spanning different architectures and methodologies to ensure a comprehensive evaluation of multimodal intent detection. We include both state-of-the-art textual and multimodal models, covering both LLM-based and non-LLM approaches. We experiment with 
smaller-size Large Language Models (LLMs) such as LLaMA-2-7B [L2-7B] and LLaMA-2-13B [L2-13B]~\cite{touvron2023llama}, LLaMA-3-8B [L3-8B]~\cite{grattafiori2024llama3herdmodels}, Qwen2.5-7B [Q-7B]~\cite{yang2024qwen2} and Mistral-7B [M-7B]~\cite{jiang2023mistral} with fine-tuning, as well as larger-sized LLMs like Claude\footnote{\url{https://claude.ai}} and GPT-4o\footnote{\url{https://openai.com/gpt-4}} using prompting for intent-detection task. 
We compare against state-of-the-art (SOTA) multimodal models such as MulT~\cite{tsai2019multimodal}, MAG-BERT~\cite{rahman-etal-2020-integrating}, and MISA~\cite{hazarika2020misa}.
To systematically analyze model performance, we categorize the selected approaches into four groups:

\noindent \textbf{(A) Textual Non-LLM Approach:}  
This approach employs BERT~\cite{kenton2019bert} [bert-base-uncased] as a text encoder to detect intents using textual features.  

\noindent \textbf{(B) Textual LLM Approach:}  
We utilize several small-sized Large Language Models (LLMs), including L2-7B, L2-13B, L3-8B, Q-7B and M-7B.  

\noindent \textbf{(C) Multimodal Non-LLM Approach:}  
We examine multiple state-of-the-art non-LLM multimodal frameworks, including MISA, MulT, MAG-BERT and SDIF-DA~\cite{huang2023sdifda}, which integrates the Shallow-to-Deep Interaction Framework with Data Augmentation (SDIF).  \saransh{BERT, MISA, MulT, MAG-BERT }are tested by the authors of M-1 and M-2.0. SDIF, a newer method, outperforms others on M-1 and is thus used in this study.

\noindent \textbf{(D) Multimodal LLM (MLLM) Approach:}  
This approach leverages advanced multimodal LLMs, including:  
(i) Video-ChatGPT~\cite{Maaz2023VideoChatGPT} [VChatGPT], which combines a CLIP-like visual encoder with Vicuna LLM;  
(ii) Video-LLaVA~\cite{lin2023videollava} [VLLaVA] which uses an encoder-decoder architecture, where LanguageBind encoders align image and video features into a unified textual space for the LLM decoder (e.g., Vicuna) to process;  
(iii) GPT-4o and Claude-3 in two settings: text-only (\textit{ClaudeT, GPT-4T}) and text with six uniformly sampled video frames (\textit{ClaudeV, GPT-4V});  
(iv) Video-LLaMA~\cite{zhang2023video} [VLLaMA], which utilizes separate vision (Video Q-former for spatiotemporal visual features) and audio branches (pre-trained ImageBind and Q-former for audio representations).  

\begin{table}[!b]
\centering
\vspace{-4mm}
\begin{adjustbox}{width=0.75\linewidth}
\begin{tabular}{|c|c c|c c|}
\hline
\textbf{Model}& \multicolumn{2}{|c|}{\textbf{M-1}} & \multicolumn{2}{|c|}{\textbf{M-2.0}} \\
\cline{2-5}
 \textbf{[Approach]} & \textbf{Acc} & \textbf{F1} & \textbf{Acc} & \textbf{F1} \\ \hline 
 \textbf{BERT [A]}& 70.8 & 67.4 & 57.1 &49.3 \\ \hline \hline
 \textbf{M-7B [B]}& \textbf{82.9} & \textbf{82.5} & \textbf{65.2} &\textbf{64.4} \\\hline 
 \textbf{L2-7B [B]}& 79.3 & 79.5 & 57.1 &56.7 \\ \hline 
 \textbf{Q-7B [B]}& 72.4 & 64.6 & 61.6 & 62.4 \\ \hline
 \textbf{L3-8B [B]}& 77.3 & 77.1 & 61.3 &59.9 \\ \hline 
 \textbf{L2-13B [B]}& 80.7 & 80.4 & 54.8 &54.3 \\\hline \hline 
\textbf{Mult [C]}& 71.5 & 67.9 & 58.4 & 51.5\\
\hline
\textbf{MAG  [C]}& 72.7 & 68.6 & 58.2 & 49.4 \\
\hline
\textbf{MISA  [C]}& 71.8 & 69.1 & 57.8 & 51.9 \\
\hline
\textbf{SDIF  [C]}& 72.8 & 71.6 & 58.6 & 52.5 \\
\hline\hline
\textbf{ClaudeT  [D]}& 57.7 & 56.4 & 39.7 & 38.1 \\
\hline
\textbf{ClaudeV  [D]}& 59.1 & 55.9 & 40.9 & 39.2 \\
\hline
\textbf{GPT-4T  [D]}& 60.4 & 59.8 & 42.1 & 41.8 \\
\hline
\textbf{GPT-4V [D]}& 59.5 & 58.8 & 41.8 & 41.4 \\
\hline
\textbf{VLLaMA [D]}& 23.6 & 23.4 & 11.8 & 11.0 \\
\hline
\textbf{VLLaVA [D]} & 32.4 & 30.5 & 18.8 & 17.5 \\
\hline
\textbf{VChatGPT [D]} & 33.3 & 33.0 & 18.7 & 16.8 \\
\hline
\end{tabular}
\end{adjustbox}
\vspace{-2mm}
\caption{Performance of different models on original dataset (Acc and F1 in \%)} 
\vspace{-2mm}
\label{tab:results-summary}
\end{table}

\begin{table*}[htb]
\centering
\vspace{-4mm}
\begin{adjustbox}{width=\linewidth}
\begin{tabular}{| c   |c   |c c  c c c  |c c c c  |c c c c c |}
\hline
   &\multicolumn{15}{|c|}{\textbf{M-1 Performance} }\\
\hline
& \textbf{BERT} & \textbf{M-7B} & \textbf{L2-7B} &   \textbf{Q-7B} &\textbf{L3-8B} & \textbf{L2-13B} & \textbf{MulT} &  \textbf{MAG} & \textbf{MISA} & \textbf{SDIF}& \textbf{ClaudeT}& \textbf{ClaudeV}& \textbf{GPT-4T}&\textbf{GPT-4V}& \textbf{VChatGPT}\\ 
\hline
\textbf{Apologize}&  96.3& 96.3&\textbf{100.0}&     96.3&92.6&92.6&96.3&  97.3&97.4& 96.3& \textbf{100.0}& \textbf{100.0}& 85.2& 96.3&92.6\\
\hline
\textbf{Flaunt}&  48.0& 50.0&{70.0}&     \textbf{80.0}&50.0&{70.0}&46.0&  42.0&54.0& 60.0& 40.0& 40.0& 60.0& 60.0&58.3\\
\hline
\textbf{Joke}&  26.0& 70.0&80.0&     20.0&70.0&\textbf{90.0}&33.0&  32.0&40.0& 60.0& 30.0& 30.0& 50.0& 50.0&11.8\\

\hline
\textbf{Taunt}&  16.7& 8.3&16.7&     25.0&16.7&25.0&16.7&  11.1&26.7& 16.7& 16.7& 8.3& 33.3& 41.7&\textbf{84.0}\\
\hline
\textbf{Thank}&  98.0& \textbf{100.0}&\textbf{100.0}&     88.0&96.0&\textbf{100.0}&97.6&  99.5&99.6& 96.0& \textbf{100.0}& 84.0& 84.0& 92.0&60.0\\
\hline
   &\multicolumn{15}{|c|}{\textbf{M-2.0 Performance}}\\
\hline
 \textbf{Apologize}& 92.9&  97.0&94.0&  \textbf{100.0}&93.8& 96.9& 94.5& 94.5& 93.2& 93.2& {98.5}& {98.5}& 81.5& 84.6&81.5\\
\hline
\textbf{Flaunt}&  9.1& 36.4&9.1&     18.2&27.3&9.1&15.4&  8.2&21.8& 20.0& 36.4& 36.4& \textbf{73.7}& \textbf{73.7}&7.4\\
\hline
\textbf{Joke}&  4.8& 10.3&17.2&     6.9&6.9&3.4&4.1&  0.0&15.2& 12.4& 28.6& 28.6& \textbf{48.1}& 37.0&0.0\\
\hline
\textbf{Taunt}&  14.5& 31.0&18.9&     34.5&20.7&37.9&25.2&  4.8&23.8& 23.1& 20.7& 20.7& \textbf{37.0}& \textbf{37.0}&26.3\\
\hline
\textbf{Thank}&  96.5& \textbf{100.0}&98.3&     94.8&\textbf{100.0}&98.3&96.5&  96.5&96.2& 96.5& 94.8& 94.8& 81.0& 81.0&67.6\\
\hline
\textbf{Refuse} &	14.1& 58.8&	41.2&  0.0&29.4& 35.3& 	12.9& 	10.6& 	23.5& 	22.3& 	47.1& 	52.9& 	52.9& 	52.9& 	\textbf{84.5}\\
\hline
\end{tabular}
\end{adjustbox}
\vspace{-2mm}
\caption{MIntRec intent-wise results (Intent Accuracy in \%); Models grouped based on approach (A to D in order). Due to space constraints extended tables for M-1 and M-2.0 are in Appendix Tables \ref{tab:extended-results-mintrec-1}, \ref{tab:extended-results-mintrec-2}, \ref{tab:mllm-mintrec-1} and \ref{tab:mllm-mintrec2.0}}
\vspace{-3mm}
\label{tab:results-mintrec}
\end{table*}
The training strategy of a model is determined by its size and resource requirements. Smaller models, such as those based on BERT, are fully trained due to their manageable size ($\sim$ few hundred MBs). For textual LLMs, fine-tuning is performed using memory-efficient techniques, including quantization to 4-bit precision and low-rank adapter (LoRA) fine-tuning, to optimize resource usage. Class A and B models utilize only text modality, and classes C and D (except GPT-4T/V, and ClaudeT/V) utilize audio, video, and text modality. GPT-4V and ClaudeV utilize text and image modality. GPT-4T and ClaudeT utilize text only.


Fine-tuning MLLMs is challenging due to size, alignment issues, and modality-specific noise, making accuracy metrics across models non-comparable. This study focuses on analyzing multimodal intent detection methods rather than optimizing performance. Detailed configurations are provided in Appendix Section \ref{sec:approach-appendix}.

\section{Experiment and Results}

Our experiment consists of three phases. In \textbf{Phase-1}, we analyze overall and intent-wise model performance. \textbf{Phase-2} then finds the smallest set of modalities needed to classify each example correctly. Finally, \textbf{Phase-3} checks for modality biases in datasets and measures model performance after de-bias techniques.


\subsection {\textbf{Phase-1: Initial Performance Analysis 
}} 



We evaluate various strategies to handle multiple modalities on two datasets with overall and intent-wise accuracy and F1-score. To enhance multimodal LLM performance, we apply in-context learning by adding a few example input-output pairs in the prompt. The model's predictions are then compared with the true intent labels. In cases where the model generates an explanation instead of a label (about 6\% in M-1 and 15\% in M-2.0), we select the label with the highest BERTScore similarity to the output.


\noindent \textbf{a) Overall Results:} We compare several methods across four model categories on the M-1 and M-2.0 datasets using accuracy and F1-score, as shown in Table~\ref{tab:results-summary}. Text-only LLMs, especially M-7B, perform best overall with accuracy of 82.9\% on M-1, and 65.2\% on M-2.0. Traditional multimodal non-LLM models (C) like Mult, MAG, MISA, and SDIF do reasonably well, with SDIF being the strongest in this group. These models outperform multimodal LLMs (D), but still fall short of text-only LLMs (B). Among multimodal LLMs, GPT-4T performs the best, but remains lower than others. This is likely due to their general-purpose chatbot type design, which limits their ability to handle structured multimodal intent tasks. Note that we do not focus on achieving state-of-the-art results; instead, our goal is to understand per-class behavior and how different types of models perform under resource constraints. Further analysis is in the Appendix section \ref{sec:phase_1_appendix}.

\noindent \textbf{b) Intent-wise Performance:} Performances of different models across different intents and evaluations (accuracy) are shown in Table \ref{tab:results-mintrec} for both MIntRec-1 (Extended Appendix Table \ref{tab:extended-results-mintrec-1}) and 2.0 datasets (Extended Appendix Table \ref{tab:extended-results-mintrec-2}). Textual LLMs excel in certain intents like \textit{Agree}, \textit{Thank}, and \textit{Greet}, achieving 100\% accuracy with the textual modality alone. However, for intents such as \textit{Taunt}, \textit{Flaunt}, and \textit{Warn}, multimodal models outperform textual models due to the additional contextual information from visual or other modalities. Small-scale multimodal models perform better on intents like \textit{Care}, \textit{Apologize}, and \textit{Ask for help}, while larger multimodal LLMs excel in more complex intents like \textit{Taunt}, \textit{Praise}, \textit{Invite}, and \textit{Complain}. On the other hand, naive multimodal fusion techniques, which treat all modalities with equal importance, can degrade performance for intents where textual cues are dominant. Overall (Table \ref{tab:results-summary}), textual LLMs and small-scale multimodal models offer practical solutions for resource-constrained settings, while larger multimodal models deliver superior performance on more nuanced intents. 

\subsection{\textbf{Phase-2: Analysis of Modality Importance}}

Given that the textual model M-7B achieves the best overall performance on both datasets, and often surpasses their multimodal counterparts, we investigate the relative contribution of each modality by examining whether samples require multimodal or unimodal information for correct classification.


\noindent \textbf{a) Optimal Modality Combination Selection: }We conduct modality-wise ablation using \textit{SDIF} for M-1 and \textit{MulT} for M-2.0, as these are the best-performing Class-C models for their respective datasets. This analysis helps to determine which combination of modalities contributes most to accurate classification. To measure the importance of each modality, we perform ablation by masking individual modalities. Each masked modality’s feature vector is replaced with a zero tensor. The models are evaluated in 7 different settings: {T}ext (T), {V}ideo (V), {A}udio (A), Text+Video (T+V), {T}ext+{A}udio (T+A), {V}ideo+{A}udio (V+A), and {T}ext+{V}ideo+{A}udio (T+V+A). 

\noindent \textbf{b) Automated Annotation Framework: }We train \textit{SDIF} and \textit{MulT} under different masking setups to identify the minimal combination of modalities required for accurate classification. For each sample, the models produce seven output probabilities corresponding to the seven modality combinations. For example, the probability corresponding to the Text+Audio setting represents the model’s output probability for the ground label when provided with text and audio while masking video. Using these seven probability features, we determine the smallest modality combination that correctly classifies the sample or emits the highest probability for the ground label. This analysis provides insight into the optimal modality fusion required for accurate classification in multimodal intent detection.

\begin{table}
\centering
\begin{adjustbox}{width=0.75\columnwidth}
\begin{tabular}{| c | c | c |}
\hline
\textbf{Modality} & \textbf{M-1} & \textbf{M-2.0} \\
\cline{2-3}
\textbf{Combination} & \textbf{\% samples} & \textbf{\% samples} \\
\hline
\textbf{Text} (T) & \textbf{69.66} & \textbf{57.85} \\
\hline
\textbf{Video} (V) & 3.60 & 0.00 \\
\hline
\textbf{Audio} (A) & 4.72 & 4.53 \\
\hline
\textbf{T+V} & 1.35 & 7.18 \\
\hline
\textbf{T+A} & 3.37 & 0.54 \\
\hline
\textbf{V+A} & 0.67 & 1.48 \\
\hline
\textbf{T+V+A} & 16.63 & 28.43 \\
\hline
\textbf{$\Sigma \, \text{T}$} & \textbf{91.01} & \textbf{94.00} \\
\hline
\textbf{$\Sigma \, \text{V}$} & 22.25 & 37.09 \\
\hline
\textbf{$\Sigma \, \text{A}$} & 25.39 & 34.97 \\
\hline
\end{tabular}
\end{adjustbox}
\vspace{-2mm}
\caption{\% of samples for distribution of modality combinations in the M-1 and M-2.0 datasets. $\Sigma \, \text{T}$ denotes the total percentage of samples that require text for correct classification, i.e., the combined share of samples in the \textbf{T}, \textbf{T+V}, \textbf{T+V+A}, and \textbf{T+A} categories.}

\label{tab:modality_combination_split}
\end{table}

\noindent \textbf{c) Key Findings: }As shown in Table \ref{tab:modality_combination_split}, \textit{Text} is the most dominant modality for intent detection, followed by the combination of \textit{Text+Audio+Video}. When combining all modality combinations that include text, approximately 90\% of the samples of both datasets require text for correct classification. This observation suggests that the dataset is not only textually biased, given that a high fraction of samples can be correctly classified using text alone, but also that text serves as the primary and most significant modality. This finding underscores the crucial role of text in intent recognition, establishing it as the foundational modality in multimodal models (relative to MIntRec datasets). Further exploratory analysis details are in the Appendix Section \ref{subsection:phase_2_appendix}.

\subsection{Phase-3: Debiasing MIntRec Datasets}


\saransh{Given that over 90\% of samples in both datasets rely on text for classification, we further analyze the nature of these samples. Specifically, we characterize textually biased examples, those correctly classified using only text, and those benefiting from additional modalities. This sheds light on the dataset's modality sensitivity and helps identify the true utility of multimodal learning in this task.}

\begin{table}
\centering
\begin{adjustbox}{width=0.75\linewidth}  

\begin{tabular}{| c | c | c | c | c |}
\hline
 & \textbf{T+V+A}& \textbf{T}& \textbf{V}& \textbf{A}\\
\hline
\textbf{Acc}& \textbf{72.8}& 69.9& 17.8& 28.5\\
\hline
\textbf{Agree}& \textbf{100.0}& 63.2& 61.4& 40.4\\
\hline
\textbf{Apologize}& \textbf{96.3}& 81.4& 11.6& 48.8\\
\hline
\textbf{Arrange}& 68.2& \textbf{96.3}& 7.4& 63.0\\
\hline
\textbf{Ask Help}& 50.0& \textbf{100.0}& 0.0& 64.0\\
\hline
\textbf{Care}& \textbf{84.2}& 39.1& 0.0& 13.0\\
\hline
\textbf{Complain}& 66.7& \textbf{91.7}& 0.0& 8.3\\
\hline
\textbf{Greet}& \textbf{83.3}& 40.0& 0.0& 10.0\\
\hline
\textbf{Inform}& 70.2& \textbf{72.2}& 0.0& 16.7\\
\hline
\textbf{Introduce}& 57.1& \textbf{89.5}& 0.0& 5.3\\
\hline
\textbf{Joke}& 60.0& \textbf{61.4}& \textbf{61.4}& 40.4\\
\hline
\textbf{Praise}& \textbf{79.1}& 66.7& 0.0& 4.8\\
\hline
\textbf{Thank}& \textbf{96.0}& 66.7& 16.7& 50.0\\
\hline

\end{tabular}
\end{adjustbox}
\vspace{-2mm}
\caption{Ablation analysis for M-1. SDIF is the best-performing small multimodal model}
\label{tab:mintrec-1-disection-mini}
\vspace{-5mm}
\end{table}

\noindent \textbf{a) Defining Textual Bias}: Text plays a dominant role in determining intents, with textual models achieving 100\% accuracy on specific intents such as \textit{Thank}, \textit{Greet}, and \textit{Apologize}. Further details are in Appendix Fig. \ref{fig:mintrec_modality_importance}. Table \ref{tab:mintrec-1-disection-mini} (Extended Appendix Tables \ref{tab:mintrec-1-disection} and \ref{tab:mintrec-2-disection}) indicates that models relying solely on text perform significantly better than those using other modalities alone. This suggests the presence of textual bias in the dataset. Upon manual inspection, we identify two types of bias: \underline{explicit} and \underline{implicit}.
Explicit bias occurs when a statement directly contains words associated with a specific intent. For instance, phrases such as ``I am sorry" or ``Thank you" contain words like ``sorry" and ``thank", which are directly linked to corresponding intent labels (Appendix Fig. \ref{fig:wordcloud}). Implicit bias refers to cases where the intent is not explicitly stated but can be inferred from the context. For example, the sentence ``Hannah, can you update the SIM card on the robot?" does not contain words directly tied to an intent category but can be inferred as a request based on the model's understanding of the context.

\noindent \textbf{b) Automated Debiasing Framework: } We propose a systematic framework for debiasing the dataset by identifying a subset of samples with reduced textual bias. Our goal is to focus on statements where textual models may struggle due to the absence of other modalities. To achieve this, we select three models: BERT (text-only), L2-7B (LLM), and SDIF (a small multimodal model). BERT is chosen for its strong textual understanding, L2-7B is selected instead of a more capable model like M-7B to ensure that even a relatively basic LLM can identify obvious textual biases, and SDIF evaluates the impact of multimodal inputs with masked audio and video cues.

The identification of textually biased samples follows a structured process. Since both datasets maintain a training, development, and test set ratio of approximately 3:1:1, we divide the training dataset into three equal parts randomly. A round-robin approach is applied where the first three parts form the training set, the fourth part serves as the development set, and the last part is used as the test set. This process is repeated for all five splits. Using these sets, three models are trained: LLaMA-2-7B is fine-tuned with QLoRA, BERT is fully trained on the training set, and the small multimodal model is trained with masked audio and video inputs. During training, models are evaluated on the development set, and early stopping is used to prevent overfitting. Once trained, they predict labels for the test set. To identify textually biased samples, any test sample that is correctly classified by the majority of these three models is labeled as textually biased. This iterative process ensures a systematic and controlled identification of textual bias while maintaining a balanced dataset for multimodal intent detection.

After debiasing, we observe that approximately 70\% of the samples in M-1 and 50\% in M-2.0 exhibit textual bias (more details in Appendix Fig. \ref{fig:debiasing}). Intent categories such as \textit{Agree}, \textit{Apologize}, \textit{Care}, \textit{Greet}, \textit{Praise}, \textit{Thank} in M-1, and \textit{Apologize}, \textit{Thank}, \textit{Greet} in M-2.0 (more details in Appendix Tables \ref{tab:mintrec-1-debiasing} and \ref{tab:mintrec-2-debiasing}) show over 85\% biased samples. In contrast, categories such as \textit{Taunt} and \textit{Flaunt} exhibit less than 25\% bias, indicating that these statements cannot be inferred by textual models alone. Further analysis in Appendix Section \ref{sec:phase_1_appendix}.

\noindent \textbf{c) Debiased Dataset Creation and Filtering:} To obtain the debiased datasets, all textually biased samples are removed. However, after this step, the six intent categories (\textit{Agree}, \textit{Apologize}, \textit{Care}, \textit{Greet}, \textit{Praise}, \textit{Thank}) of M-1 and the three of M-2.0 (\textit{Apologize}, \textit{Thank}, \textit{Greet}) with high bias are left with very few samples across different splits. Due to this limited representation, we exclude these categories. The final debiased dataset consists of 14 intent categories in M-1 and 27 in M-2.0.

The debiased dataset for M-1 consists of 290 train, 107 dev, and 108 test samples. For M-2.0, the dataset includes 2,826 train, 475 dev, and 891 test samples (Fig. \ref{fig:debiasing}). To validate the consistency of the observations, prior experiments are re-conducted on the debiased datasets.

\begin{table}
\centering
\begin{adjustbox}{width=0.75\columnwidth}
\begin{tabular}{| c  |c  |c|c |c|}
\hline
 \textbf{Model} & \multicolumn{2}{|c|}{\textbf{M-1}}& \multicolumn{2}{|c|}{\textbf{M-2.0}}\\
\cline{2-5}
\textbf{[Approach]}& \textbf{Acc}&\textbf{F1}& \textbf{Acc}&\textbf{F1}\\
\hline
\textbf{BERT [A]}& 16.7&23.6& 18.5&22.6\\
\hline
\hline
\textbf{M-7B [B]}& 32.4&38.7& \textbf{33.6}&34.7\\
\hline
\textbf{L2-7B [B]}& 30.6&33.8& 30.1&33.4\\
\hline
\textbf{Q-7B [B]} & 35.2 & 35.2 & 31.3 & 30.9 \\
\hline
\textbf{L3-8B [B]}& 25.0&30.0& {32.2}&\textbf{35.2}\\
\hline
\textbf{L2-13B [B]}& 37.0&38.0& 31.5&34.1\\
\hline
\hline
\textbf{Mult [C]}& 27.8&29.4& 22.7&25.4\\
\hline
\textbf{MAG [C]}& 19.4&25.8& 21.0&24.1\\
\hline
\textbf{MISA [C]}& 22.2&26.6& 21.4&24.7\\
\hline
\textbf{SDIF [C]}& 22.2&25.6& 19.4&23.1\\
\hline
\hline
\textbf{ClaudeT [D]}& 34.3&32.9& 24.5&22.6\\
\hline
\textbf{ClaudeV [D]}&29.6 & 27.6& 25.8&23.6\\
\hline
\textbf{GPT-4T [D]}& 44.9&40.2& 26.0&26.1\\
\hline
\textbf{GPT-4V [D]}& \textbf{52.9}&\textbf{54.1}& 30.5&31.0\\
\hline
\textbf{VLLaMA [D]}& 9.3& 7.4& 4.8& 4.8 \\
\hline
\textbf{VLLaVA [D]} & 11.1& 9.6& 9.9& 8.8\\
\hline
\textbf{VChatGPT [D]}& 18.0&19.7& 12.3&15.5\\
\hline

\end{tabular}
\end{adjustbox}
\vspace{-2mm}
\caption{Results of different models after Phase-3 i.e. Debiasing (Acc and F1 in \%)}
\vspace{-5mm}
\label{tab:results-summary-debias}
\end{table}

\begin{table*}
\centering
\begin{adjustbox}{width = \textwidth}
\begin{tabular}{| c  |c  |c  c  lc  c  |c  c  c c  |c  c  c c  c |}
\hline
\multicolumn{16}{|c|}{\textbf{M-1  Performance After Debiasing}} \\
\hline
& \textbf{BERT}& \textbf{M-7B}& \textbf{L2-7B} &\textbf{Q-7B}& \textbf{L3-8B}& \textbf{L2-13B}& \textbf{MulT}& \textbf{MAG}& \textbf{MISA}& \textbf{SDIF}& \textbf{ClaudeT}& \textbf{ClaudeV}& \textbf{GPT-4T}& \textbf{GPT-4V}& \textbf{VChatGPT}\\
\hline
\textbf{Arrange}& 0.0& 0.0& 14.3 &\textbf{42.9}& {28.6}& 14.3& {28.6}& 0.0& 0.0& 14.3& 14.3& {28.6}& 20.0& 14.3& 0.0\\
\hline
\textbf{Flaunt}& 0.0& 0.0& \textbf{50.0} &16.7& 33.3& 33.3& 16.7& 0.0& 0.0& 16.7& \textbf{50.0}& 33.3& 33.3& \textbf{50.0}& 40.0\\
\hline
\textbf{Joke}& 33.3& 33.3& \textbf{83.3} &66.7& 50.0& 50.0& 33.3& 16.7& 16.7& 33.3& 50.0& 16.7& 50.0& 62.5& 16.7\\
\hline
\textbf{Taunt}& 0.0& 0.0& 30.0 &30.0& 0.0& 0.0& 0.0& 20.0& 20.0& 10.0& 10.0& 10.0& \textbf{100.0}& \textbf{100.0}& 14.3\\
\hline
\textbf{Apologize, Thank}& \multicolumn{14}{c}{\textbf{Removed}} &\\
\hline
\multicolumn{16}{|c|}{\textbf{M-2.0  Performance After Debiasing}} \\
\hline
\textbf{Arrange}& 5.9& 5.9& 5.9 &11.8& 0.0& \textbf{23.5}& 5.9& {17.7}& {17.7}& 5.9& 6.3& 6.3& 5.9& 5.1& 0.0\\
\hline
\textbf{Flaunt}& 0.0& {38.9}& 27.8 &\textbf{44.4}& 5.6& 33.3& 11.1& 0.0& 5.6& 11.1& 27.8& 27.8& 25.5& 30.6& 0.0\\
\hline
\textbf{Joke}& 12.0& 28.0& 12.0 &24.0& 4.0& 8.0& 4.0& 4.0& 12.0& 8.0& 29.2& 29.2& 40.0& \textbf{44.0}& 0.0\\
\hline
\textbf{Taunt}& 8.7& 43.5& 8.7 &54.4& 23.9& 23.9& 17.4& 28.3& 15.2& 17.4& 17.4& 17.4& 66.7& \textbf{72.7}& 7.7\\
\hline
\textbf{Refuse}& 0.0& 21.4& 28.6 &\textbf{71.4}& 35.7& 21.4& 7.1& 0.0& 0.0& 7.1& 35.7& 42.9& 58.8& {64.3}& 9.1\\
\hline
\textbf{Apologize, Thank}& \multicolumn{14}{c}{\textbf{Removed}} &\\
\hline
\end{tabular}\end{adjustbox}
\vspace{-2mm}
\caption{MIntRec intent-wise results post debiasing (Intent Accuracy in \%); Models grouped on the basis of approach; Due to space constraints, extended tables for M-1 and M-2.0 are in Tables \ref{tab:extended-mintrec-1-debiased-result}, \ref{tab:extended-mintrec-2-debiased-result}, \ref{tab:mllm-mintrec-1-debias} and \ref{tab:mllm-mintrec2.0-debias}}
\vspace{-5mm}
\label{tab:results-mintrec-debiased}
\end{table*}

\noindent \textbf{d) Performance Analysis on Debiased Datasets:} Table \ref{tab:results-summary-debias} shows model performance after debiasing, with all models dropping compared to Table \ref{tab:results-summary}. 1) On M-1, GPT-4V performs best and outperforms GPT-4T on both datasets, showing the need for visual cues. 2) Textual LLMs show a sharp decline, especially in the \textit{Joke}, \textit{Comfort}, and \textit{Flaunt} (Table \ref{tab:results-mintrec-debiased}) intents where visual cues like facial expressions or tonal cues play a crucial role. 3) Small multimodal models, which were competitive before, now perform similarly to BERT, highlighting the limits of BERT-based multimodal approaches and needs for better fusion mechanisms to effectively balance the contributions of different modalities. 4) MLLMs like VChatGPT achieve better results in categories such as \textit{Acknowledge} (Table \ref{tab:results-mintrec-debiased}) despite the absence of fine-tuning. These findings suggest that advanced techniques improve the recognition of specific intents. More detailed and intent-specific results are in Appendix Section \ref{subsection:phase_3_appendix} and Table \ref{tab:combined_mintrec_1_table} and \ref{tab:combined_mintrec_2_table}.


\noindent \textbf{e) Analyzing the Impact of Dataset Size vs. Textual Bias on Performance: } We analyze whether the performance drop observed in different classes of models is due to the samples being textually unbiased or a reduction in sample size. We use a random subset of the original dataset. We then train several models on this subset to compare their performance with the debiased dataset. The 
train, test, and dev splits of the random sample and the debiased sample are the same. Table \ref{tab:subset_debias} indicates that the observed performance drop across various model categories is primarily due to their inability to accurately classify textually unbiased samples, rather than a reduction in dataset size.

\subsection{Experimental Settings} All our experiments are conducted on a single A100 GPU with 80 GB RAM and 2*960 GB SSD. The training of small multimodal models takes $\sim$1 hour. Text-LLMs require $\sim$3-5 hours for training and 1-2 hours for inference. MLLMs need 1-4 hours for inference. Small multimodal models are trained using a cross-entropy loss with a learning rate of 5e-6 over $100$ epochs. The Text-LLMs are fine-tuned using prompts and 4-bit quantization, supported by QLoRA \cite{NEURIPS2023_1feb8787}, 
with a learning rate of 2e-4. AdamW optimizer, enabled with early termination, is used in all cases. We use the NLTK, spaCy, and NumPy packages for further analysis and ablations. Further details are in Appendix section \ref{sec:experiment}.

\section{Human Annotation Verification}
To assess automated annotation accuracy, we randomly sample 15-20\% of M-1 (444 samples) and M-2.0 (1,489 samples) for manual verification. Three Annotators are selected to classify samples into five modality categories: Text, Video, Audio+Text, Video+Text, or Audio+Video+Text, excluding audio-only samples. Initial labeling is done by
two annotators and any annotation discrepancy is
checked and resolved by the third annotator after discussing with others. Annotators are chosen after discussions and conditions of fulfilling many criteria like domain
knowledge expertise along with a good working
proficiency in English. Inter-annotator agreement 85.7\%.

Results indicate strong textual bias, with over 80\% of samples in both datasets primarily relying on text. In M-1, the distribution was: T (82.46\%), T+V (5.69\%), T+V+A (5.01\%), T+A (4.56\%), and V (2.28\%). In M-2.0, it was: T (80.63\%), T+V+A (7.60\%), T+V (6.50\%), T+A (4.31\%), and V (0.96\%).
Findings suggest that humans rely more on visual cues than audio for interpretation. A possible approach to address bias is training a classifier to predict the required modality subset for each sample, ensuring decisions are based only on the most relevant modalities. Detailed guidelines are in Appendix section \ref{sec : human_annotations}.

\begin{table}
\centering
\begin{adjustbox}{width=0.93\columnwidth}
\begin{tabular}{|c |c c |c c |c c|c c|}
\hline
\multirow{3}{*}{\textbf{Model}} & \multicolumn{4}{|c|}{\textbf{M-1}} & \multicolumn{4}{|c|}{\textbf{M-2.0}} \\
\cline{2-9}
& \multicolumn{2}{|c|}{\textbf{Random}} & \multicolumn{2}{|c|}{\textbf{Debiased}} & \multicolumn{2}{|c|}{\textbf{Random}} & \multicolumn{2}{|c|}{\textbf{Debiased}} \\
\cline{2-9}
& \textbf{Acc} & \textbf{F1} & \textbf{Acc} & \textbf{F1} & \textbf{Acc} & \textbf{F1} & \textbf{Acc} & \textbf{F1} \\
\hline
\textbf{BERT [A]}& 67.6& 56.4& 16.7& 23.6& 57.7& 52.7& 18.5&22.6\\
\hline
\textbf{M-7B [B]}& 72.2& 73.2& 32.4& 38.7& 59.3& 57.7& 33.6&34.7\\
\hline
\textbf{SDIF [C]}& 59.4& 59.0& 22.2& 25.6& 59.2& 52.1& 19.4&23.1\\
\hline
\textbf{GPT-4T [D]}& 73.1& 75.3& 44.9& 40.2& 50.7& 49.6& 26.0&26.1\\
\hline
\textbf{GPT-4V [D]}& 63.0& 66.7& 52.9& 54.1& 50.7& 49.5& 30.5&31.0\\
\hline

\end{tabular}
\end{adjustbox}
\vspace{-2mm}
\caption{Performance Comparison: Random Subset vs. Debiased Data for MIntRec Datasets}
\label{tab:subset_debias}
\vspace{-6mm}
\end{table}

\begin{figure*}[t]
\centering
  \includegraphics[width=0.9\linewidth]{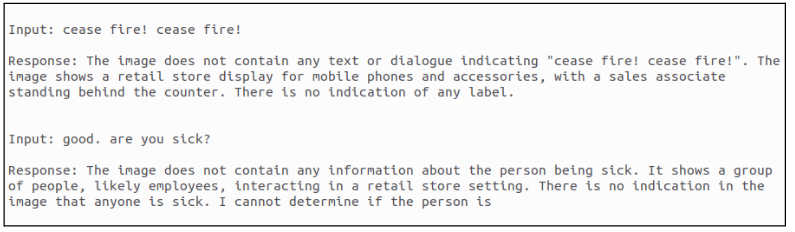}
  \vspace{-3mm}
  \caption {Multimodal model confusion caused due to image frames provided}
  \vspace{-3mm}
  \label{fig:claude-mis}
\end{figure*}

\section{Ablation Analysis}

\noindent \textbf{Modality Contributions: }
Table~\ref{tab:mintrec-1-disection-mini} highlights the impact of individual modalities. Audio column exhibits greater variation than the video column, which predominantly classifies samples into broad categories (e.g., positive, negative, neutral). This suggests that tonal variations in speech provide richer information than visual features. Over 80\% of intents like \textit{Apologize}, \textit{Thank}, and \textit{Praise} show strong textual bias, whereas intents such as \textit{Joke}, \textit{Taunt}, and \textit{Asking for opinions} require video for accurate classification (Appendix Fig.~\ref{fig:text_needed} and~\ref{fig:video_needed}). For further details refer to Appendix \ref{sec:few-shot-results}.

\noindent \textbf{Effectiveness of Multimodal Models: }Multimodal models like MulT, SDIF perform better by using text to guide attention over audio and video on textually biased original dataset. Large MLLMs (e.g., VLLaMA, VLLaVA, VChatGPT) struggle due to limited fine-tuning and high computational requirements. Smaller multimodal models achieve comparable or better performance, making them more practical for real-world applications. \saransh{Further details on fusion techniques in Appendix \ref{sec:fusion-comparison}.}

\sharma{\noindent \textbf{Generalizability Analysis:}
Generalizability is a key concern for intent detection since user intents evolve and datasets differ in label space. Traditional multimodal and BERT-based models, with fixed output sizes, often need retraining to adapt. Large language models, however, show stronger transfer across datasets. For instance, Mistral-7B fine-tuned on MIntRec2.0 reached 72.58\% accuracy and 72.85\% F1 on MIntRec-1, while a model trained on MIntRec-1 achieved 47.47\% accuracy and 40.42\% F1 on MIntRec2.0. The drop mainly comes from label mismatches, but overall performance remains competitive with major baselines. These findings highlight the promise of LLMs for handling evolving intents with less retraining. Full analysis is provided in Appendix~\ref{sec:model-general}.}

\sharma{\noindent \textbf{Analysis of Fair Comparisons:}Due to resource limits, textual LLMs are fine-tuned with QLoRA, while multimodal LLMs are tested in few-shot settings. This raises concerns about fairness, since textual models may appear to perform better largely because they benefit from fine-tuning. We argue that the performance gap is not only due to differences in model size or training strategies, but is also strongly influenced by text-biased samples in the dataset, which naturally favor textual LLMs over multimodal ones. To study this, we run an ablation with LLaMA-3-8B Instruct and Mistral-7B Instruct, checking if fine-tuning explains the advantage of textual LLMs. We use the same few-shot prompts and system messages for fairness for all models. }

\begin{table}[h]
\centering
\begin{adjustbox}{width=0.98\columnwidth}
\begin{tabular}{|c|c|c|c|c|c|}
\hline
\textbf{Dataset} & \textbf{Model} & \textbf{M-1 (Acc)} & \textbf{M-1 (F1)} & \textbf{M-2.0 (Acc)} & \textbf{M-2.0 (F1)} \\
\hline
\multirow{2}{*}{Original} 
& LLaMA-3-8B-Instruct & 44.49 & 38.34 & 31.19 & 27.40 \\
\cline{2-6}
& Mistral-7B-Instruct & 64.27 & 59.80 & 39.80 & 34.27 \\
\hline
\multirow{2}{*}{Debiased} 
& LLaMA-3-8B-Instruct & 26.85 & 15.93 & 25.64 & 5.73 \\
\cline{2-6}
& Mistral-7B-Instruct & 26.85 & 19.78 & 28.65 & 5.00 \\
\hline
\end{tabular}
\end{adjustbox}
\caption{Performance of Instruction-tuned models on MIntRec-1 and 2.0 datasets (Original vs Debiased)}
\vspace{-5mm}
\label{tab:combined}
\end{table}

\begin{table*}[!h]
  \centering
    \begin{adjustbox}{width=0.88\textwidth}
    \begin{tabular}{|c|c|c|c|}
        \hline
        \textbf{Input} & \textbf{Textual LLM} & \textbf{Multimodal Model} & \textbf{Type of Issue} \\
        \hline
        Ugh, calm down & \textcolor{red}{Comfort} & \textcolor{teal}{Taunt} & Missing facial expression cues \\
        \hline
        It's a pipe organ & \textcolor{teal}{Introduce} & \textcolor{red}{Explain} & Object not visible in frames \\
        \hline
        Great job & \textcolor{red}{Praise} & \textcolor{teal}{Taunt} & Missing tonal cues \\
        \hline
        Anyway, Zephra asked me to set up a call & \textcolor{red}{Arrange} & \textcolor{red}{Arrange} & Confusion in similar labels (\textcolor{teal}{Inform}) \\
        \hline
    \end{tabular}
    \end{adjustbox}
    \vspace{-2mm}
    \caption{Erroneous cases in textual and multimodal models, highlighting intent recognition limitations}
    \vspace{-3mm}
    \label{tab:error_cases}
\end{table*}

\sharma{Table: \ref{tab:combined} shows that few-shot textual LLMs perform worse than their fine-tuned counterparts, confirming that fine-tuning provides task-specific benefits. However, the observed gap between textual and multimodal LLMs cannot be attributed solely to fine-tuning. While fine-tuning alters the absolute accuracy levels, it does not fundamentally change the relative behavior; textual LLMs consistently outperform open-source multimodal models even without fine-tuning on the original dataset. This suggests that the benchmarks themselves are predominantly text-heavy, with only a limited contribution from other modalities. Furthermore, our debiasing step reduces the dataset’s over-reliance on textual content, causing both fine-tuned and non-fine-tuned textual models to drop in performance, thereby narrowing the gap. This underscores that the original dataset’s strong text bias is a substantial factor, but not the only reason for the performance differences observed.}

\section{Discussion and Error Analysis}
We closely examine the mistakes made by the evaluated models and find the following key insights:

\noindent \textbf{a) Multimodal LLMs:} \sharma{These models often struggle to understand speaker intent in situations involving emotions or abstract meanings. While they use both visual and audio cues, they get confused when the visual information does not match the spoken words. For example, if someone says \textit{``It is a pipe organ''} (correct label: \textit{Introduce}), the model may fail when the pipe organ is not visible in the video. In such cases, it may rely on unrelated signals like facial expressions or the background, leading to errors, as shown in Table~\ref{tab:error_cases}. Image or audio quality can also play a major role in performance. When a video is converted into a small set of frames (sampled uniformly or randomly), the model may become confused because it loses the temporal information linking the frames. This issue is shown in Figure~\ref{fig:claude-mis}. In addition, these models require considerable computational resources for fine-tuning and have difficulty combining multiple modalities when those signals are weak or noisy.}

\noindent \textbf{b) Textual LLMs:} These models work well for intent classes that are clear from text alone but struggle when body language or tone is needed. For example, \textit{``Ugh, calm down''} is often mislabeled as \textit{Comfort} instead of \textit{Taunt}, or \textit{``Great job''} is labeled as \textit{Praise} instead of \textit{Taunt}. These errors show the limitations of relying solely on text for tasks that require a deeper understanding of tone or context.

\noindent \textbf{c) Small Multimodal Models:} These models with BERT backend act like a noisier version of BERT and only give small improvements, if any. When textually biased samples are removed, their performance drops significantly. They also suffer from similar issues as larger multimodal models, misalignment of modalities, and difficulty handling abstract or emotion-driven samples.

\noindent \textbf{Correlated Labels:} All models face challenges when intent classes are highly similar. For example, \textit{``Anyway, Zephra asked me to set up a call''} is incorrectly labeled as \textit{Arrange} instead of \textit{Inform}. Though both labels might seem valid, a closer look shows the true intent is to inform. Both textual and multimodal models miss fine distinctions like these. More details are provided in Appendix~\ref {sec:error-analysis}.

Our analysis shows that the modality needed for correct classification varies across samples. Some samples require only text, others need audio or visual cues. Any fixed rule at the dataset or label level fails to capture this variation. To handle the sample-specific nature of modality use, future work should explore adaptive models that can route and process inputs based on their modality needs. Mixture-of-experts or routing-based frameworks can help focus compute on the relevant parts of the input. This also requires datasets that are better balanced across modalities and robust fusion methods that can work well even when only certain modalities are informative. Refer to Appendix Section~\ref{sec:input-adaptive} for more discussion.

\section{Conclusion}

In this work, we study multimodal intent detection using benchmark datasets like MIntRec. We observe that Mistral-7B, a text-only model, achieves the best performance due to the strong text bias in the datasets. While around 50\% and 70\% of the samples can be accurately classified using only textual information, over 90\% of the samples require textual input, either alone or in combination with other modalities, for correct classification, which limits the usefulness of these benchmarks for evaluating multimodal models. To address this, we propose a debiasing strategy that removes text-dominant samples (around 70\% and 55\% samples in M-1 and M-2.0, respectively). After debiasing, model performances drop significantly. The drop exceeds 60\% in smaller multimodal models, while textual models also see a drop of more than 50\%, showing that current models depend heavily on text features. 
Our comprehensive analysis shows the need for unbiased datasets, more effective fusion frameworks, and adaptive approaches that can select relevant modalities for each sample. 

\section*{Limitations}

Our work has the following limitations: 

(I) While our experiments focus on the MIntRec datasets, we have not conducted experiments on domain-specific data due to the lack of publicly available datasets, restricting the scope of our findings. Future research can explore intent detection in specific domains such as education and healthcare. 

(II) The experimental datasets used in this study are limited to English, and extending this work to a multilingual setting, particularly for low-resource languages, would provide a broader understanding of multimodal intent detection and address language-related biases. 

(III) Fine-tuning MLLMs is challenging due to their large size, modality alignment issues, and modality-specific noise. \saransh{These challenges lead to variations in training approaches; however, since our objective is to gain deeper insights into model behavior rather than achieve state-of-the-art performance, the majority of our observations should remain unaffected.}

(IV) Due to hardware limitations, we were unable to fine-tune the open-source MLLMs. While fine-tuning could have improved the performance of the best model, it is unlikely to change key findings such as the impact of textual bias or the relative performance of different models. 

\noindent Addressing these limitations in future studies would provide a more comprehensive evaluation of the models and methods explored in this work.

\section*{Ethical Concerns} 
We use publicly available datasets and approaches, ensuring compliance with all ethical guidelines. The datasets do not contain personally sensitive information, avoiding ethical concerns.

\bibliography{main_latex}

\newpage
\section*{Appendix}
\subsection{Motivation for Multimodal Intent Detection}

\saransh{Multimodal intent detection is chosen as the focus of this study because it is interpretable, involves dynamic selection of input types, and provides a compact but challenging setup to evaluate fusion strategies. Unlike tasks such as Visual Question Answering (VQA), Crystal property prediction~\citep{das2023crysmmnet,das2025periodic}, where the main input modality is usually fixed (e.g., vision), intent detection requires the model to decide which combination of audio, video, and text is most useful for each example. This makes the task more complex and better aligned with real-world scenarios, where input relevance can vary significantly.}

\saransh{The task has a wide range of practical applications, including robotics, conversational agents, and task-driven systems. Robots interacting with humans need to understand intent through gestures, tone, and spoken content. Chatbots and virtual assistants also benefit from multimodal understanding to respond more effectively. Although the study focuses on intent detection, the proposed tools and findings are general and can be applied to other multimodal tasks such as visual dialogue, video-based agents, and embodied interaction.}

\saransh{This work takes a systematic approach to understanding multimodal intent detection by analyzing a variety of models and datasets. The analysis reveals strong textual biases in existing benchmarks, 70\% in the M-1 dataset and 50\% in M-2.0 (see Fig.~\ref{fig:debiasing}), indicating that models often rely heavily on text while ignoring other modalities. In the second phase of the study, a detailed modality-wise analysis shows that certain intents like \textit{Apologize} and \textit{Thank} are easy to detect from text alone, while others, such as \textit{Complain}, \textit{Flaunt}, and \textit{Inform}, benefit significantly from audio or video input (Figs.~\ref{fig:video_needed}, \ref{fig:audio_needed}).  These findings highlight the need for more balanced and less biased datasets to evaluate multimodal models fairly.} This concept of intent is different from the notion of entity-phrases~\citep{mullick2021rte, guha2021matscie, mullick2022using, mullick2024matscire} and opinion~\citep{mullick2016graphical, mullick2017extracting, mullick2017generic,mullick2018identifying, mullick2018harnessing, mullick2019d}. 

\saransh{To address this issue, the study introduces a method to reduce textual bias in datasets and evaluates model performance under these more controlled conditions. In addition, it proposes tools to measure how different modalities contribute to predictions and how models adapt when a modality is missing or misleading. These tools provide deeper insights into model behavior beyond simple accuracy. While the experiments focus on intent detection, the approach is general and can support the development of robust systems for any task involving multiple input modalities. The dynamic nature of modality relevance in intent detection makes it an ideal testbed for studying adaptive multimodal learning in real-world scenarios.}

\subsection{Dataset}
\label{sec:multimodal-dataset}
MIntRec-1~\textbf{[M-1]}~\citep{10.1145/3503161.3547906} contains 2,224 annotated samples and is introduced as the first benchmark dataset for multimodal intent classification with 20 intent labels across text, audio, and video modalities.

The MDID dataset~\citep{kruk2019integrating} provides binary intent labels using Instagram posts with image context. EmoInt-Trans~\citep{9961847} presents the EmoInt dataset with annotations for 32 emotions and 15 empathetic intents. However, the dataset is currently not publicly accessible due to technical issues.

MIntRec2.0~\textbf{[M-2.0]}~\citep{zhang2024mintrec} is the largest available dataset for multimodal intent detection, labeled under a 30-class intent taxonomy, with inputs from text, audio, and video sources.

In our study, we use both MIntRec-1 and MIntRec2.0 under the fine-grained intent classification setting, which consists of 20 and 30 intent labels, respectively.

The authors of the MIntRec datasets have evaluated various baseline models, including BERT, MAG-BERT, MISA, MulT, and SDIF\_DA. They claim that extensive experiments on their datasets show that using multimodal data leads to better performance compared to only using text. However, based on our observations, we find that both MIntRec datasets contain a significant number of samples that are heavily biased towards the textual modality.

\begin{figure*}
    \centering
    \fbox{\includegraphics[width=0.9\linewidth]{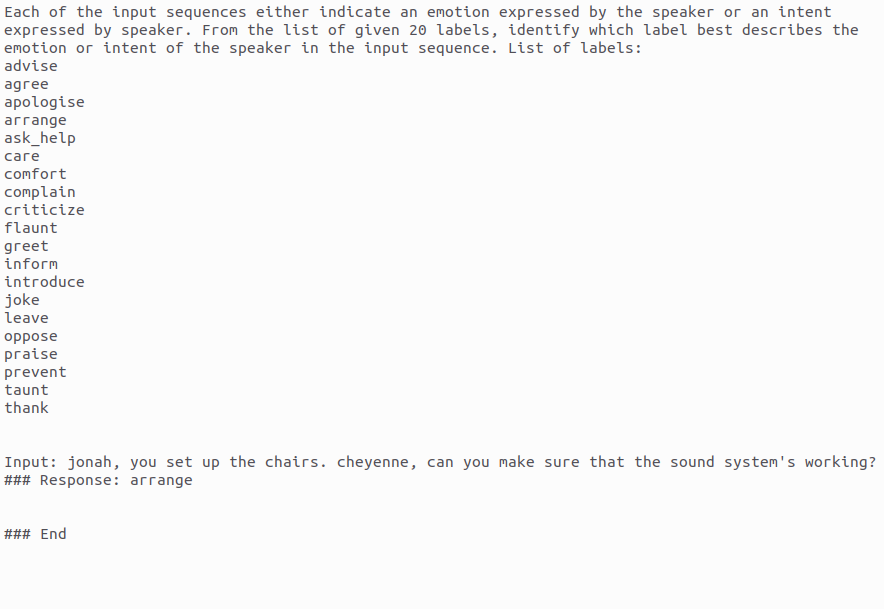}}
  \caption {Textual LLM prompt for finetuning}
  \label{fig:llm-prompt}
\end{figure*}

\begin{figure*}
\centering
  \fbox{\includegraphics[width=0.9\linewidth]{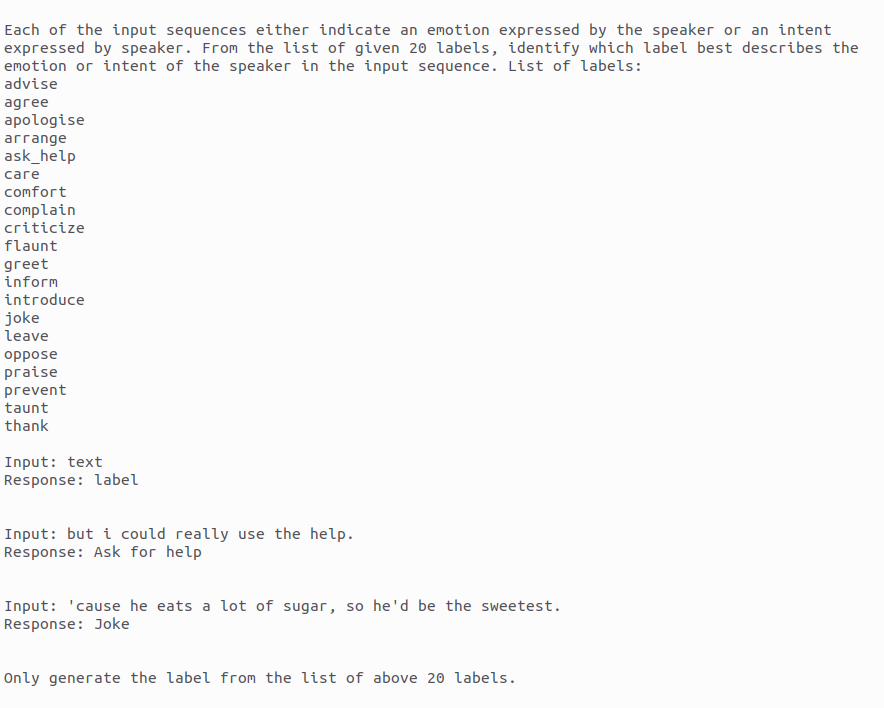}}
  \caption {Inferencing prompt (for models used without training, few-shot)}
  \label{fig:gpt-prompt}
\end{figure*}

\subsection{Approach}
\label{sec:approach-appendix}

\subsubsection{Textual LLM Approach}
\saransh{We chose LLaMA and Mistral models because they cover a range of practical sizes, their weights are openly available, and they have strong community support. By working with models in the 7 billion to 13 billion parameter range, we avoid the engineering overhead of very large models while still observing how capacity affects performance. This size span lets us study small-to-mid models that can run on single GPUs without distributed setups.}

\saransh{According to Hugging Face download statistics, the LLaMA2-7B \footnote{https://huggingface.co/meta-llama/Llama-2-7b-hf} model has significantly more (last month) downloads compared to the relatively newer Qwen2-7B \footnote{https://huggingface.co/Qwen/Qwen2-7B} and Qwen2.5-7B \footnote{https://huggingface.co/Qwen/Qwen2.5-7B} models. The relatively recent Qwen2.5-7B shows growing interest, but it remains well behind LLaMA2-7B. These figures show that any insights we draw from LLaMA models will be broadly relevant to the research community and practitioners already using these checkpoints.}

\saransh{Our experiments include four models: LLaMA2-7B, LLaMA2-13B, LLaMA3-8B, and Mistral-7B, alongside Qwen2.5-7B for comparison as a newer 7 billion-parameter alternative. Comparing LLaMA3-8B to Mistral-7B holds the size constant to highlight differences in architecture and training data. Comparing LLaMA2-7B to LLaMA2-13B isolates the effect of roughly doubling parameters within the same family. Including Qwen2.5-7B adds perspective on a newer entrant at the 7 billion-parameter scale.}

\saransh{Our goal is not to declare one model universally \textbf{\textit{best}}, but to understand how design choices and size affect behavior along two key dimensions. First, we measure intent-specific performance: how accurately each model follows different task types, from direct factual queries to multi-step instructions. Second, we test bias mitigation by probing for unwanted associations and stereotyping, then comparing how those biases change with model size and training source. By focusing on relative performance within each model class and size, we ensure our observations reflect architectural and training differences rather than raw capacity alone. }

\subsubsection{Multimodal Non-LLM Approach}
\label{sec:mllm}
The following multimodal baselines are utilized for our study:

\textbf{MISA~\citep{hazarika2020misa}}: The fundamental idea behind MISA (Multimodal Invariant and Specific Representations) lies in learning two distinct subspaces for each modality: a modality-invariant subspace and a modality-specific subspace. In essence, MISA takes multiple modalities into account by learning representations that capture both shared features and modality-specific nuances, thus enabling a more comprehensive understanding of the underlying affective content in multimodal data. 

\textbf{MulT~\citep{tsai2019multimodal}}: The Multimodal Transformer (MulT) model has a crossmodal attention module, which allows the model to attend to crossmodal interactions across entire utterances. This module dynamically adapts features from one modality to another, facilitating the modeling of long-range crossmodal contingencies without the need for manual word alignment preprocessing.

\textbf{MAG-BERT~\citep{rahman-etal-2020-integrating}}: MAG-BERT handles multiple modalities by leveraging its Multimodal Adaptation Gate (MAG) component, which enables BERT to process multimodal input data effectively. This framework allows BERT to adapt seamlessly to the integration of nonverbal information, such as visual and acoustic cues, alongside linguistic input. During the fine-tuning process, MAG-BERT utilizes attention mechanisms conditioned on nonverbal behaviors to map informative features from different modalities to a vector representation.

\textbf{SDIF-DA~\citep{huang2023sdifda}}: The Shallow-to-Deep Interaction Framework with Data Augmentation (SDIF-DA) utilizes a hierarchical structure composed of shallow and deep interaction modules. Initially, the shallow interaction module preliminarily aligns video and audio features with text features. Subsequently, a KQV attention mechanism is employed to obtain bimodal representations of video and audio aligned with text. These aligned features are then concatenated with the original text features to create a trimodal representation, facilitating shallow integration across modalities. SDIF-DA also incorporates a ChatGPT-based data augmentation approach to fine-tune the text encoder, aiming to enhance the model's performance.

\begin{table}
\centering
\begin{adjustbox}{width=0.8\columnwidth} 
\begin{tabular}{| c | c | c |}
\hline
 & \textbf{Input} & \textbf{Output} \\
\hline
\textbf{GPT-4o} & \$5.00 / 1M tokens & \$15.00 / 1M tokens \\
\hline
\textbf{Claude-3-Haiku} & \$0.25 / 1M tokens & \$1.25 / 1M tokens \\
\hline
\end{tabular}
\end{adjustbox}
\caption{Close Source API Cost}
\label{tab:close-price}
\end{table}

\subsubsection{Multimodal LLM Approach}
\label{sec:mllm-1}
With the introduction of advanced multimodal large language models, we also evaluate their capabilities in constrained generation tasks such as intent detection. Additionally, we assess their performance beyond traditional applications like object detection and question answering.

\begin{table}[h]
\centering
\begin{adjustbox}{width=0.7\linewidth} 
\begin{tabular}{| c  |c  |c  |}
\hline
\textbf{Model} & \textbf{Num Param} & \textbf{Memory} \\
\hline
\textbf{MAG} & 112.5M & 429.20 MB \\
\hline
\textbf{MISA} & 140M & 534.15 MB \\
\hline
\textbf{MulT} & 136M & 519.39 MB \\
\hline
\textbf{BERT} & 109.5M & 417.70 MB \\
\hline
\hline
\textbf{M-7B} & 7B & 15 GB \\
\hline
\textbf{L2-7B} & 7B & 13 GB \\
\hline
\textbf{L3-8B} & 8B & 16 GB \\
\hline
\textbf{L2-13B} & 13B & 26 GB \\
\hline
\hline
\textbf{VLLaMA} & 15B & 30 GB \\
\hline
\textbf{VLLaVA} & 11B & 22 GB \\
\hline
\textbf{VChatGPT} & 11B & 22 GB \\
\hline

\end{tabular}
\end{adjustbox}
\caption{Model Parameter and Memory Requirement Details}
\label{tab:model-details}
\end{table}

\textbf{Video-ChatGPT~\citep{Maaz2023VideoChatGPT}[VChatGPT]}: Video-ChatGPT is a novel multimodal model that combines a video-adapted visual encoder with a large language model (LLM). The main idea is to enable detailed video understanding and generation of meaningful conversations about videos. It leverages a pretrained visual encoder like CLIP to extract spatiotemporal video features, which are then integrated with an LLM like Vicuna. 

\textbf{Video-LLaMA~\citep{zhang2023video}[VLLaMA]}: Video-LLaMA enables large language models to simultaneously understand both visual and auditory content in videos. Its unique idea is to have separate vision and audio branches that process the video frames and audio signals in parallel. The vision branch uses techniques like a Video Q-former to extract spatiotemporal visual features, while the audio branch leverages pre-trained audio models like ImageBind and an Audio Q-former to obtain audio representations. These multimodal representations are then aligned to the language model's embedding space through cross-modal pre-training on video/audio captioning data. 

\begin{table}
\centering
\begin{adjustbox}{width=0.5\columnwidth} 

\begin{tabular}{| c | c | c |}
\hline
 & \textbf{M-1} & \textbf{M-2.0} \\
\hline
\textbf{M-7B} & 140 & 180 \\
\hline
\textbf{L2-7B} & 120 & 150 \\
\hline
\textbf{L3-8B} & 150 & 200 \\
\hline
\textbf{L2-13B} & 180 & 230 \\
\hline
\end{tabular}
\end{adjustbox}
\caption{Finetuning-LLM Timings in min after quantization}
\label{tab:finetuning-time}
\vspace{-3mm}
\end{table}

\textbf{Video-LLaVA~\citep{lin2023videollava}[VLLaVA]}: Video-LLaVA employs an encoder-decoder architecture where the encoder maps raw image and video inputs into a unified visual feature space, while the decoder is a large language model like Vicuna. Specifically, it utilizes LanguageBind encoders pretrained on vision-language data to extract visual features from images and videos. These visual encoders are designed to align the representations of different modalities into a common textual feature space. This allows the image and video features to be mapped into a single unified visual representation. 

\textbf{GPT-4o, Claude-3-Haiku}: Closed-source chatbots such as GPT and Claude have shown impressive capabilities to understand multiple modalities. Despite their APIs not directly supporting video input, we utilize their functionality by including 6 uniformly sampled frames from the video along with the text utterance in the input. We test them under two settings: text-only \textbf{[ClaudeT, GPT-4T]} and text with 6 frames \textbf{[ClaudeV, GPT-4V]}.

\begin{table*}[h]
\centering
\begin{adjustbox}{width=\linewidth}
\begin{tabular}{| c  |c  |c c  c c c  |c c c c   |c c c c c |}
\hline
 &  \textbf{BERT} &   \textbf{M-7B} &\textbf{L2-7B}  &\textbf{Q-7B}&  \textbf{L3-8B} &\textbf{L2-13B} &\textbf{MulT} &  \textbf{MAG} &\textbf{MISA} & \textbf{SDIF} & \textbf{ClaudeT} & \textbf{ClaudeV} & \textbf{GPT-4T} & \textbf{GPT-4V}  & \textbf{VChatGPT} \\ 
\hline
\textbf{Advise} &  74.8 &   84.0 &88.0  &60.0&  \textbf{96.0} &80.0 &74.4 &  75.5 &72.8 & 80.0 & 48.0 & 52.0 & {76.0} & 40.0  &12.0\\
\hline
\textbf{Agree} &  95.8 &   \textbf{100.0} &91.7  &91.7&  91.7 &\textbf{100.0} &96.7 &  95.0 &96.7 & \textbf{100.0} & \textbf{100.0} & \textbf{100.0} & \textbf{100.0} & 83.3  &0.0\\
\hline
\textbf{Apologize} &  96.3 &   96.3 &\textbf{100.0}  &96.3&  92.6 &92.6 &96.3 &  97.3 &97.4 & 96.3 & \textbf{100.0} & \textbf{100.0} & 85.2 & 96.3  &92.6\\
\hline
\textbf{Arrange} &  66.8 &   {86.4} &81.8  &\textbf{95.5}&  63.6 &77.3 &71.4 &  67.6 &66.8 & 68.2 & 22.7 & {54.5} & 31.8 & 18.2  &0.0\\
\hline
\textbf{Ask Help} &  64.0 &   {80.0} &70.0  &70.0&  70.0 &60.0 &63.0 &  66.7 &63.0 & 50.0 & 80.0 & 40.0 & \textbf{100.0} & 90.0  &15.8\\
\hline
\textbf{Care} &  89.5 &   \textbf{94.7 }&89.5  &\textbf{94.7}&  84.2 &89.5 &\textbf{94.7} &  89.5 &89.5 & 84.2 & 15.8 & 31.6 & 15.8 & 15.8  &27.8\\
\hline
\textbf{Comfort} &  78.3 &   \textbf{88.9} &\textbf{88.9}  &66.7&  83.3 &83.3 &73.3 &  81.9 &78.9 & 77.8 & 44.4 & 72.2 & 66.7 & {77.8}  &24.6\\
\hline
\textbf{Complain} &  61.8 &   \textbf{82.5} &75.4  &80.7&  75.4 &71.9 &63.5 &  65.7 &58.6 & 66.7 & 64.9 & 59.6 & {66.7} & 57.9  &43.5\\
\hline
\textbf{Criticize} &  47.8 &   \textbf{78.3} &69.6  &73.9&  65.2 &\textbf{78.3} &50.0 &  51.3 &52.2 & 39.1 & 69.6 & {73.9} & 30.4 & 69.6  &0.0\\
\hline
\textbf{Flaunt} &  48.0 &   50.0 &{70.0}  &\textbf{80.0}&  50.0 &{70.0} &46.0 &  42.0 &54.0 & 60.0 & 40.0 & 40.0 & 60.0 & 60.0  &58.3\\
\hline
\textbf{Greet} &  77.5 &   {91.7} &83.3  &83.3&  {91.7} &75.0 &74.2 &  85.0 &71.7 & 83.3 & \textbf{100.0} & \textbf{100.0} & 83.3 & 91.7  &24.6\\
\hline
\textbf{Inform} &  68.6 &   78.9 &73.7  &35.1&  75.4 &\textbf{80.7} &70.2 &  68.2 &72.3 & 70.2 & 54.4 & 47.4 & 64.9 & {66.7}  &14.3\\
\hline
\textbf{Introduce} &  68.6 &   \textbf{85.7} &66.7  &\textbf{85.7}&  71.4 &76.2 &71.0 &  71.1 &68.1 & 57.1 & 14.3 & 23.8 & 23.8 & 19.0  &60.0\\
\hline
\textbf{Joke} &  26.0 &   70.0 &80.0  &20.0&  70.0 &\textbf{90.0} &33.0 &  32.0 &40.0 & 60.0 & 30.0 & 30.0 & {50.0} & {50.0}  &11.8\\
\hline
\textbf{Leave} &  72.4 &   76.5 &76.5  &58.8&  \textbf{88.2} &\textbf{88.2} &71.2 &  75.3 &72.4 & 82.4 & 41.2 & 47.1 & 41.2 & {47.1}  &10.0\\
\hline
\textbf{Oppose} &  33.0 &   60.0 &\textbf{80.0}  &50.0&  60.0 &70.0 &32.0 &  37.3 &35.0 & 40.0 & 50.0 & \textbf{80.0} & 40.0 & 60.0  &48.8\\
\hline
\textbf{Praise} &  84.9 &   \textbf{90.7} &86.0  &\textbf{90.7}&  81.4 &83.7 &83.3 &  87.8 &85.8 & 79.1 & {72.1} & {72.1} & 65.1 & 62.8  &40.0\\
\hline
\textbf{Prevent} &  80.0 &   93.3 &66.7  &80.0&  73.3 &\textbf{100.0} &79.3 &  84.0 &78.0 & 80.0 & 40.0 & 33.3 & {73.3} & 46.7  &0.0\\
\hline
\textbf{Taunt} &  16.7 &   8.3 &16.7  &25.0&  16.7 &25.0 &16.7 &  11.1 &{26.7} & 16.7 & 16.7 & 8.3 & 33.3 & 41.7  &\textbf{84.0}\\
\hline
\textbf{Thank} &  98.0 &   \textbf{100.0} &\textbf{100.0}  &88.0&  96.0 &\textbf{100.0} &97.6 &  99.5 &99.6 & 96.0 & \textbf{100.0} & 84.0 & 84.0 & 92.0  &60.0\\
\hline
\end{tabular}
\label{tab:mintrec-results}
\end{adjustbox}
\caption{Extended MIntRec-1 Results (Accuracy in \%; MulT, MISA, MAG and SDIF are small multimodal models; L2-7B, L2-13B, M-7B, Q-7B and L3-8B are textual LLMs; VChatGPT is Multimodal LLM)(Extended version of Main Table \ref{tab:results-mintrec})}
\label{tab:extended-results-mintrec-1}
\end{table*}

\subsection{Experimental Setup}
\label{sec:experiment}

All small multimodal models, including SDIF, MAG, MulT, and MISA, are trained with a learning rate of 5e-6 for a maximum of 100 epochs with early termination enabled. On average, these models are trained for 20 epochs, and their results are averaged across 5 seed values. {Textual LLMs are trained and evaluated using carefully curated prompts, as shown in Figure \ref{fig:llm-prompt}, for fine-tuning. These prompts are refined through experimentation to determine the most effective format for intent classification. The best-performing prompt is then adapted across different models to ensure consistency in evaluation. For inference, similar prompts are used with the label removed from the response for both fine-tuned and multimodal LLMs, ensuring a uniform testing strategy across all models.}. The textual LLMs are fine-tuned using the AdamW optimizer with a learning rate of 2e-4, 4-bit quantization, and are modeled as a text-generation task. {Table \ref{tab:model-details} shows that smaller models like MAG, MISA, MulT, and BERT have fewer parameters (around 100–140 million) and require very little memory (less than 1 GB), which makes them easier to deploy on less powerful systems. In contrast, models with billions of parameters need significantly more memory, up to 30 GB, highlighting the trade-off between model size and the computational resources required.}
\begin{table}[h]
\centering
\begin{adjustbox}{width=0.5\columnwidth} 
\begin{tabular}{| c | c | c |}
\hline
 & \textbf{M-1} & \textbf{M-2.0} \\
\hline
\textbf{M-7B} & 80 & 100 \\
\hline
\textbf{L2-7B} & 65 & 100 \\
\hline
\textbf{L3-8B} & 80 & 100 \\
\hline
\textbf{L2-13B} & 90 & 120 \\
\hline
\hline
\textbf{VChatGPT} & 90 & 180 \\
\hline
\textbf{VLLaVA} & 100 & 240 \\
\hline
\textbf{VLLaMA} & 100 & 240 \\
\hline

\end{tabular}
\end{adjustbox}
\caption{Inferencing-LLM Timings in min (First four are after quantization, meanwhile the remaining three are without quantization)}
\label{tab:inference-time}
\vspace{-3mm}
\end{table}

\begin{table*}[h]
\centering
\begin{adjustbox}{width=\linewidth}
\begin{tabular}{| c  |c  |c c  c c c   |c c c c   |c c c c c |}
\hline
 &       \textbf{BERT} & \textbf{M-7B} &\textbf{L2-7B}  &\textbf{Q-7B}&\textbf{L3-8B} &\textbf{L2-13B} &\textbf{MulT} &\textbf{MAG} & \textbf{MISA} & \textbf{SDIF} & \textbf{ClaudeT} & \textbf{ClaudeV} & \textbf{GPT-4T} & \textbf{GPT-4V}  & \textbf{VLLaVA}\\
\hline
\textbf{Acknowledge} &       72.3 & 55.9 &76.5  &8.8&\textbf{86.8} &66.2 &65.9 &71.2 & 69.4 & 70.9 & 76.5 & 75.0 & 76.5 & 76.5  &29.4\\
\hline
\textbf{Advise} &       58.6 & 67.0 &67.0  &69.1&\textbf{77.3} &46.4 &60.2 &61.6 & 53.6 & 60.2 & 41.2 & 41.2 & 55.9 & 53.8  &8.2\\
\hline
\textbf{Agree} &       63.1 & {75.0} &64.0  &71.9&62.5 &70.3 &66.6 &68.7 & 61.9 & 64.1 & 51.6 & \textbf{79.7} & 38.1 & 34.9  &0\\
\hline
\textbf{Apologize} &       92.9 & {97.0} &94.0  &\textbf{100.0}&93.8 &96.9 &94.5 &94.5 & 93.2 & 93.2 & {98.5} & {98.5} & 81.5 & 84.6  &81.5\\
\hline
\textbf{Arrange} &       67.1 & \textbf{82.0} &65.0  &66.7&60.8 &45.1 &65.1 &67.1 & 62.3 & 59.6 & 14.6 & 14.6 & 10.0 & 10.0  &5.9\\
\hline
\textbf{Ask Help} &       52.4 & {70.0} &49.0  &73.0&21.6 &32.4 &57.8 &51.3 & 63.8 & 64.9 & 83.8 & \textbf{86.5} & {72.9} & {72.9}  &2.1\\
\hline
\textbf{Ask Opinions} &       56.2 & 50.0 &39.6  &\textbf{79.2}&27.1 &35.4 &57.9 &60.0 & 54.2 & {61.7} & 50.0 & 50.0 & 59.6 & 59.6  &34.7\\
\hline
\textbf{Care} &       51.1 & \textbf{61.7} &40.4  &21.3&31.9 &27.7 &55.3 &52.8 & 51.5 & 50.2 & 2.2 & 2.2 & 8.5 & 8.5  &10.3\\
\hline
\textbf{Comfort} &       42.0 & \textbf{48.9} &46.9  &34.7&38.8 &44.9 &44.5 &45.7 & 46.5 & 42.9 & 29.2 & 29.2 & 22.9 & 25.0  &0\\
\hline
\textbf{Complain} &       42.2 & {55.1} &51.4  &\textbf{71.0}&54.2 &53.3 &45.8 &51.2 & 43.7 & 51.4 & 52.4 & 52.4 & {57.8} & {57.8}  &23.4\\
\hline
\textbf{Confirm} &       46.7 & 44.1 &38.7  &15.1&\textbf{75.3} &44.1 &49.9 &47.9 & 48.4 & 47.7 & 8.9 & 8.9 & 18.3 & 18.3  &7.4\\
\hline
\textbf{Criticize} &       40.8 & 38.3 &34.0  &40.4&34.0 &40.4 &40.9 &37.4 & 36.6 & {44.7} & \textbf{58.7} & \textbf{58.7} & 36.4 & 36.4  &17.6\\
\hline
\textbf{Doubt} &       62.4 & \textbf{80.0} &64.4  &72.6&60.7 &53.3 &64.0 &64.6 & 67.6 & 61.8 & 20.0 & 20.0 & 40.1 & 40.1  &2.1\\
\hline
\textbf{Emphasize} &       2.3 & {23.5} &5.9  &0.0&11.8 &17.6 &2.3 &1.2 & 5.9 & 3.5 & 23.5 & 23.5 & \textbf{31.2} & \textbf{31.2}  &9.1\\
\hline
\textbf{Explain} &       57.2 & \textbf{76.2} &61.1  &68.9&63.7 &60.6 &63.2 &59.3 & 60.8 & 61.4 & 12.4 & 12.9 & 17.0 & 17.0  &46.9\\
\hline
\textbf{Flaunt} &       9.1 & {36.4} &9.1  &18.2&27.3 &9.1 &15.4 &8.2 & 21.8 & 20 & 36.4 & 36.4 &\textbf{ 73.7 }& \textbf{73.7}  &7.4\\
\hline
\textbf{Greet} &       82.7 & 86.4 &83.3  &80.3&84.8 &\textbf{90.9} &86.4 &85.1 & 81.8 & 80.6 & 83.3 & 83.3 & 70.3 & 70.3  &13.3\\
\hline
\textbf{Inform} &       60.9 & 64.1 &60.8  &68.7&\textbf{71.4} &53.9 &55.0 &56.8 & 58.9 & 54.3 & 45.2 & 46.7 & 46.9 & 46.9  &0\\
\hline
\textbf{Introduce} &       44.0 & 41.7 &33.3  &43.3&40.0 &23.3 &40.7 &\textbf{49.7} & 42.3 & 45.0 & 11.8 & 11.8 & 20.7 & 20.7  &41.4\\
\hline
\textbf{Invite} &       34.1 & 52.9 &52.9  &11.8&52.9 &{76.5} &38.8 &24.7 & 29.4 & 34.1 & 76.5 & 76.5 & \textbf{82.3} & \textbf{82.3}  &7.6\\
\hline
\textbf{Joke} &       4.8 & 10.3 &{17.2}  &6.9&6.9 &3.4 &4.1 &0 & 15.2 & 12.4 & 28.6 & 28.6 & \textbf{48.1} & 37.0  &0\\
\hline
\textbf{Leave} &       45.3 & 50.9 &45.3  &37.7&49.1 &47.2 &50.9 &49.8 & 44.9 & \textbf{55.1} & 30.2 & 30.2 & 34.6 & 34.6  &2.6\\
\hline
\textbf{Oppose} &       73.7 & 71.8 &69.2  &\textbf{83.8}&71.8 &64.9 &73.8 &{74.9} & 66.5 & 69.1 & 12.9 & 13.8 & 9.6 & 9.6  &49.6\\
\hline
\textbf{Plan} &       56.3 & \textbf{71.0} &50.0  &60.5&47.4 &57.9 &46.3 &55.3 & 60.5 & 53.7 & 33.3 & 33.3 & 19.4 & 19.4  &9.7\\
\hline
\textbf{Praise} &       75.9 & \textbf{85.8} &67.3  &84.1&77.9 &73.4 &77.8 &78.7 & 76.5 & 75.6 & 59.5 & 59.5 & 56.7 & 56.7 &11.8\\
\hline
\textbf{Prevent} &       57.4 & 64.5 &41.9  &45.2&\textbf{74.2} &64.5 &66.4 &64.5 & 58.7 & 56.1 & 16.1 & 16.1 & 36.7 & 26.7  &20.7\\
\hline
\textbf{Refuse} &       14.1 & {58.8} &41.2  &0.0&29.4 &35.3 &12.9 &10.6 & 23.5 & 22.3 & 47.1 & 52.9 & 52.9 & 52.9  &\textbf{84.5}\\
\hline
\textbf{Taunt} &       14.5 & 31.0 &18.9  &34.5&20.7 &\textbf{37.9} &25.2 &4.8 & 23.8 & 23.1 & 20.7 & 20.7 & 37.0 & 37.0  &26.3\\
\hline
\textbf{Thank} &       96.5 & \textbf{100.0} &98.3  &94.8&\textbf{100.0} &98.3 &96.5 &96.5 & 96.2 & 96.5 & 94.8 & 94.8 & 81.0 & 81.0  &67.6\\
\hline
\textbf{Warn} &       13.7 & {42.1} &36.8  &42.1&{42.1} &15.8 &15.8 &10.5 & 25.3 & 32.6 & 47.4 & 47.4 & \textbf{72.2} & \textbf{72.2} &31.2\\
\hline

\end{tabular}
\end{adjustbox}
\caption{Extended Results for MIntRec2.0 (Accuracy in \%; MulT, MISA, MAG and SDIF are small multimodal models; L2-7B, L2-13B, M-7B, Q-7B and L3-8B are textual LLMs; VLLaVA is multimodal LLM)(Extended version of Main Table \ref{tab:results-mintrec})}
\label{tab:extended-results-mintrec-2}
\end{table*}


Tables \ref{tab:finetuning-time} and \ref{tab:inference-time} summarize the time taken to fine-tune and perform inference on various LLMs. As model size increases, the time required for both fine-tuning and inference also grows significantly. This increase in time results from the higher computational complexity and greater number of parameters in larger models. Additionally, the time taken for inference is influenced by model size and the amount of test data; larger datasets result in longer inference times.

Small-scale multimodal models, on the other hand, are more time-efficient. They take considerably less time for training compared to larger LLMs, making them very resource-efficient and practical for scenarios with limited computational resources or where quick turnaround is essential. This efficiency is particularly advantageous in iterative development environments or real-time applications where rapid model updates and low-latency inference are critical.

All no-training experiments on textual LLMs, including GPT-4T and ClaudeT, are performed by appending 4/5 input samples to the prompt (Fig. \ref{fig:gpt-prompt}), which is then appended to every input. For GPT-4V and ClaudeV with visual input, due to cost constraints (Table \ref{tab:close-price}), we uniformly sample 6 images from the video. These images are then cropped down to 1024x1024 resolution, each image translating roughly to 170 tokens.

\saransh{Different models are employed in varying configurations not to establish state-of-the-art performance, but rather to derive deeper insights into the behavior of current systems under real-world constraints. Our study prioritizes understanding the limitations of models and datasets for multimodal intent detection over optimizing performance. While it might appear unfair to compare fine-tuned textual models against non-fine-tuned multimodal models, this decision is grounded in practical considerations such as limited computational resources and the feasibility of training large models on a single GPU. Models like BERT, being lightweight, can be fully trained from scratch. Textual LLMs such as GPT-4T and ClaudeT, which have already undergone extensive pretraining, can be adapted to the task using parameter-efficient fine-tuning methods like LoRA. In contrast, large multimodal LLMs like GPT-4V and ClaudeV are significantly more complex due to the inclusion of pretrained encoders for each modality. These models are commonly used in zero-shot or few-shot settings, as full fine-tuning is not only computationally expensive but also conceptually challenging. Specifically, fine-tuning multimodal models requires careful coordination across different modality-specific encoders to avoid misalignment, which can lead to degraded or inconsistent outputs.}

\saransh{Moreover, the scale, architecture, and training history of these models differ vastly. Attempting to normalize these differences by enforcing a single training or inference strategy would be infeasible under our resource constraints and less informative for our practical goal. Our comparisons aim to examine the relative performance trends within and across model classes rather than absolute performance. Even though fine-tuning may improve certain metrics like accuracy or F1 score, our key findings, particularly those concerning bias, modality dominance, and ablation performance, are based on comparative patterns rather than isolated outcomes. For instance, we observe that text often dominates multimodal fusion and balancing modalities can cause significant drops in performance. These behaviors persist regardless of model size or training strategy, indicating that they are systemic issues rather than artifacts of specific setups. Thus, the fairness of comparing models under different settings does not undermine the validity of our conclusions. Instead, it reflects a realistic evaluation framework for developing robust and scalable intent detection systems in constrained environments.}

\subsection{Phase-1: Performance Analysis Across Models}
\label{sec:phase_1_appendix}

\subsubsection{Intent-Wise Analysis}
\label{sec:intent-wise}

In the MIntRec-1 dataset (Table \ref{tab:extended-results-mintrec-1}), textual LLMs excel in intent labels like \textit{Agree}, \textit{Prevent}, and \textit{Thank}. Small-scale multimodal models perform better on labels like \textit{Care} and \textit{Apologize}, while larger multimodal LLMs excel in \textit{Taunt}, \textit{Praise}, \textit{Greet}, and \textit{Ask for help}.

In the MIntRec2.0 dataset (Table \ref{tab:extended-results-mintrec-2}), textual LLMs perform better on labels such as \textit{Agree}, \textit{Thank}, \textit{Acknowledge}, \textit{Greet}, and \textit{Praise}. Small-scale multimodal models outperform on \textit{Oppose}, \textit{Leave}, \textit{Introduce}, and \textit{Ask for help}, while larger multimodal LLMs excel in \textit{Refuse}, \textit{Invite}, \textit{Ask for help}, and \textit{Complain}. The MIntRec2.0 dataset, with its numerous intent labels, displays less consistent performance trends compared to MIntRec-1. We, therefore, report overall trends wherein most of the models of that type perform better.

\begin{table}[h]
\centering
\begin{adjustbox}{width=0.8\columnwidth} 

\begin{tabular}{| c  |c  |c  |c  |}
\hline
 & \textbf{VLLaMA} & \textbf{VLLaVA} & \textbf{VChatGPT} \\
\hline
\textbf{Advise} & \textbf{16.0} & 0.0 & 12.0 \\
\hline
\textbf{Agree} & 0.0 & 0.0 & 0.0 \\
\hline
\textbf{Apologize} & 66.7 & \textbf{92.6} & \textbf{92.6} \\
\hline
\textbf{Arrange} & \textbf{4.6} & \textbf{4.6} & 0 \\
\hline
\textbf{Ask Help} & 10.5 & 10.5 & \textbf{15.8} \\
\hline
\textbf{Care} & 22.2 & \textbf{61.1} & 27.8 \\
\hline
\textbf{Comfort} & 21.0 & 14.0 & \textbf{24.6} \\
\hline
\textbf{Complain} & 21.7 & 30.4 & \textbf{43.5} \\
\hline
\textbf{Criticize} & 0.0 & \textbf{10.0} & 0.0 \\
\hline
\textbf{Flaunt} & 50.0 & \textbf{66.7} & 58.3 \\
\hline
\textbf{Greet} & 17.5 & 7.0 & \textbf{24.6} \\
\hline
\textbf{Inform} & \textbf{14.3} & \textbf{14.3} & \textbf{14.3} \\
\hline
\textbf{Introduce} & 10.0 & 20.0 & \textbf{60.0} \\
\hline
\textbf{Joke} & 0.0 & \textbf{11.8} & \textbf{11.8} \\
\hline
\textbf{Leave} & 0.0 & 0.0 &\textbf{ 10.0} \\
\hline
\textbf{Oppose} & 30.2 & \textbf{67.4} & 48.8 \\
\hline
\textbf{Praise} & 0.0 & 26.7 & \textbf{40.0} \\
\hline
\textbf{Prevent} & 0.0& \textbf{16.7} & 0.0 \\
\hline
\textbf{Taunt} & 68.0 & \textbf{100.0} & 84.0 \\
\hline
\textbf{Thank} & 90.0 & \textbf{100.0} & 60.0 \\
\hline

\end{tabular}
\end{adjustbox}
\caption{MultiModal LLM Results on MIntRec-1 (Accuracy in \%)(Extended version of Main Table \ref{tab:results-mintrec})}
\label{tab:mllm-mintrec-1}
\vspace{-3mm}
\end{table}

\begin{table}[h]
\centering
\begin{adjustbox}{width=0.8\columnwidth} 

\begin{tabular}{| c  |c  |c  |c  |}
\hline
 & \textbf{VLLaMA} & \textbf{VLLaVA} & \textbf{VChatGPT} \\
\hline
\textbf{Accuracy} & 11.8 & \textbf{18.8} & 18.7 \\
\hline
\textbf{Acknowledge} & 22.1 & \textbf{29.4} & 13.2 \\
\hline
\textbf{Advise} & 4.1 & \textbf{8.2} & \textbf{8.2} \\
\hline
\textbf{Agree} & 0.0 & 0.0 & 0.0 \\
\hline
\textbf{Apologize} & 70.8 & 81.5 & \textbf{95.4} \\
\hline
\textbf{Arrange} & 3.9 & \textbf{5.9} & 0.0 \\
\hline
\textbf{Ask Help} & 0.0 & \textbf{2.1} & 0.0 \\
\hline
\textbf{Ask Opinions} & 14.3 & \textbf{34.7} & 4.3 \\
\hline
\textbf{Care} & 7.5 & 10.3 & \textbf{42.9} \\
\hline
\textbf{Comfort} & 1.1 & 0.0 & \textbf{17.8} \\
\hline
\textbf{Complain} & 12.77 & \textbf{23.4} & 0 \\
\hline
\textbf{Confirm} & 6.7 & 7.4 & \textbf{23.4} \\
\hline
\textbf{Criticize} & 11.8 & \textbf{17.6} & 4.4 \\
\hline
\textbf{Doubt} & 3.1 & 2.1 & \textbf{11.8} \\
\hline
\textbf{Emphasize} & 0.0 & \textbf{9.1} & 0.5 \\
\hline
\textbf{Explain} & 39.4 & \textbf{46.9} & 9.1 \\
\hline
\textbf{flaunt} & 3.7 & 7.4 & \textbf{42.4} \\
\hline
\textbf{Greet} & 5.0 & \textbf{13.3} & 12.9 \\
\hline
\textbf{Inform} & 5.9 & 0.0 & \textbf{6.7} \\
\hline
\textbf{Introduce} & 3.4 & \textbf{41.5} & 0.0 \\
\hline
\textbf{Invite} & 3.8 & 7.5 & \textbf{62.1} \\
\hline
\textbf{Joke} & 0.8 & 0.0 & \textbf{24.5} \\
\hline
\textbf{Leave} & 5.3 & 2.6 & \textbf{12.8} \\
\hline
\textbf{Oppose} & {15.0} & \textbf{49.6} & 2.6 \\
\hline
\textbf{Plan} & 6.4 & 9.7 & \textbf{38.9} \\
\hline
\textbf{Praise} & 0.0 & 11.8 & \textbf{25.8} \\
\hline
\textbf{Prevent} & 1.7 & \textbf{20.7} & 11.8\\
\hline
\textbf{Refuse} & \textbf{84.5} & \textbf{84.5} & 0.0 \\
\hline
\textbf{Taunt} & 5.3 & 26.3 & \textbf{82.8} \\
\hline
\textbf{Thank} & 43.2 & \textbf{67.6} & 21.0 \\
\hline
\textbf{Warn} & 8.3 & 31.2 & \textbf{67.6} \\
\hline

\end{tabular}
\end{adjustbox}
\caption{MultiModal LLM Results on MIntRec2.0 (Accuracy in \%)(Extended version of Main Table \ref{tab:results-mintrec})}
\label{tab:mllm-mintrec2.0}
\vspace{-3mm}
\end{table}

Tables \ref{tab:mllm-mintrec-1} and \ref{tab:mllm-mintrec2.0} summarize the results of the intent detection task performed on MIntRec-1 and MIntRec2.0, respectively.




\subsubsection{Generalizability of Models}
\label{sec:model-general}
Generalizability is a key aspect for intent detection models, as user intents continuously evolve and a fixed model may require frequent retraining. Traditional small multimodal models and BERT-based architectures use a fixed output layer size to predict a set number of probabilities (20 for MIntRec-1 and 30 for MIntRec2.0). In contrast, large language models (LLMs) such as GPT-4o and Claude benefit from extensive pretraining, allowing them to perform well on multiple datasets in zero- or few-shot settings without additional training.

We evaluate the generalizability of a textual LLM by selecting Mistral-7B for its strong performance. Mistral-7B, fine-tuned on the MIntRec2.0 dataset, is tested on the MIntRec-1 dataset by adjusting the class labels in the prompt. The system achieves an Overall Accuracy of 72.58\% and an Overall F1 Score of 72.85\%. Given that MIntRec2.0 already contains most MIntRec-1 labels, this result was expected despite a slight decline caused by differences between the two datasets. MIntRec2.0 includes labels such as  \textit{Oppose}, \textit{Asking for opinions} that are absent in MIntRec-1, leading to some misclassifications.

In another experiment, the model fine-tuned on MIntRec-1 with 20 labels is tested on MIntRec2.0 with 30 labels, resulting in an Overall Accuracy of 47.47\% and an Overall F1 Score of 40.42\%. This performance remains competitive compared to several closed-source and open-source models in zero- or few-shot settings.

In summary, while parameter-efficient fine-tuning can limit generalizability, the models still demonstrate acceptable performance across different datasets.

\subsubsection{Performance with limited data analysis}
\label{sec:few-shot-results}

Table~\ref{tab:few-shot-results} compares model performance under zero-shot, 10-shot, and full-shot settings. Zero-shot performance is weak across all models, while few-shot learning significantly improves results. Tables \ref{tab:zero-mintrec1} and \ref{tab:zero-mintrec2} display the performance of these models under zero-shot settings, while Tables \ref{tab:shot-10-mintrec1} and \ref{tab:shot-10-mintrec2} present their performance under 10-shot settings.

 \begin{table}[h]
\centering
\begin{adjustbox}{width=\linewidth}    
\begin{tabular}{| c  |c |c |c |c |c |c |}
\hline
& \multicolumn{3}{|c|}{\textbf{M-1}}  & \multicolumn{3}{|c|}{\textbf{M-2.0}}  \\
\hline
 & \textbf{0-shot} & \textbf{10-shot}  & \textbf{Full-shot} & \textbf{0-shot} & \textbf{10-shot}  & \textbf{Full-shot} \\ \hline 
  \textbf{BERT}& 5.0& 57.2& 70.9& 4.3& 40.9&57.1\\\hline \hline 
  \textbf{M-7B} & 4.7& 67.4 &\textbf{82.9}& 3.9& \textbf{45.4}&\textbf{65.2}\\
\hline
  \textbf{L2-7B} & 5.4& 28.8&79.3& 2.9& 39.1&57.2\\
\hline
\textbf{L3-8B} & 4.9& 67.2&77.3& 0.3& 42.4&61.3\\
\hline
\textbf{L2-13B} & 4.9& 66.5&80.7& \textbf{10.1}& 41.1&54.8\\
\hline\hline
 \textbf{MulT}& 6.5& \textbf{68.8}& 71.5& 4.1& 36.9&58.4\\ \hline 
 \textbf{MAG}& 4.9& 56.5& 72.7& 3.6& 39.2&58.2\\ \hline 
 \textbf{MISA}& 4.8& 53.4& 71.8& 3.3& 37.6&57.8\\ \hline 
 \textbf{SDIF}& \textbf{12.8}& 52.4& 72.8& 6.6& 38.6&58.2\\ \hline 
\end{tabular}
\end{adjustbox}
\caption{Zero/Few-shot Results}
\label{tab:few-shot-results}
\end{table}

Notably, MulT surpasses M-7B on MIntRec-1 in the 10-shot setting, likely due to its cross-modal attention mechanism and compact architecture, which facilitates learning patterns more efficiently than large language models (LLMs) that require a sufficient number of samples for improved performance. However, M-7B achieves the best overall performance, emphasizing the critical role of adequate training data. Since we use pre-trained variants of the models, rather than chat-finetuned variants, the zero-shot experiments fail to accurately predict several intents. However, as the amount of training data increases, the performance of the LLMs improves significantly, highlighting their sensitivity to the quantity of training data. Notably, even under 10-shot settings, the LLMs perform remarkably well, nearly achieving the performance levels seen with full fine-tuning. These trends are consistent across both the MIntRec-1 and MIntRec2.0 datasets.

\begin{table}
\centering
\begin{adjustbox}{width=0.8\columnwidth} 
\begin{tabular}{| c  |c|c  |c  |c |}
\hline
\textbf{Model} &  \textbf{M-7B} &\textbf{L2-7B} & \textbf{L3-8B} & \textbf{L2-13B} \\
\hline
\textbf{Accuracy} &  4.7&\textbf{5.4}& 4.9& 4.9\\
\hline
\textbf{Advise} &  8.0 &\textbf{12.0} & 0.0 & 0.0 \\
\hline
\textbf{Agree} &  \textbf{8.3}&0.0& 0.0 & 0.0 \\
\hline
\textbf{Apologize} &  \textbf{3.7}&0.0 & 0.0 & 0.0 \\
\hline
\textbf{Arrange} &  9.1&90.9& \textbf{100.0} &\textbf{ 100.0} \\
\hline
\textbf{Ask Help} &  \textbf{10.0} &0.0 & 0.0 & 0.0 \\
\hline
\textbf{Care} &  0.0 &0.0 & 0.0 & 0.0 \\
\hline
\textbf{Comfort} &  \textbf{5.6}&0.0 & 0.0 & 0.0 \\
\hline
\textbf{Complain} &  \textbf{3.5}&0.0 & 0.0 & 0.0 \\
\hline
\textbf{Criticize} &  0.0 &0.0 & 0.0 & 0.0 \\
\hline
\textbf{Flaunt} &  0.0 &0.0 & 0.0 & 0.0 \\
\hline
\textbf{Greet} &  \textbf{16.7}&0.0 & 0.0 & 0.0 \\
\hline
\textbf{Inform} &  \textbf{3.5}&1.7& 0.0 & 0.0 \\
\hline
\textbf{Introduce} &  \textbf{9.5}&0.0 & 0.0 & 0.0 \\
\hline
\textbf{Joke} &  0.0 &0.0 & 0.0 & 0.0 \\
\hline
\textbf{Leave} &  0.0 &0.0 & 0.0 & 0.0 \\
\hline
\textbf{Oppose} &  \textbf{10.0} &0.0 & 0.0 & 0.0 \\
\hline
\textbf{Praise} &  \textbf{4.6}&0.0 & 0.0 & 0.0 \\
\hline
\textbf{Prevent} &  0.0 &0.0 & 0.0 & 0.0 \\
\hline
\textbf{Taunt} &  \textbf{8.3}&0.0 & 0.0 & 0.0 \\
\hline
\textbf{Thank} &  \textbf{4.0} &0.0 & 0.0 & 0.0 \\
\hline
\end{tabular}
\end{adjustbox}
\caption{Zero Shot results MIntRec-1 (Extended version of Table \ref{tab:few-shot-results})}
\label{tab:zero-mintrec1}
\vspace{-3mm}
\end{table}

\begin{table}
\centering
\begin{adjustbox}{width=0.8\columnwidth} 
\begin{tabular}{| c  |c|c  |c  |c |}
\hline
\textbf{Epoch} &  \textbf{M-7B} &\textbf{L2-7B} & \textbf{L3-8B} & \textbf{L2-13B} \\
\hline
\textbf{Accuracy} &  \textbf{67.2} &28.7 & 65.4& 66.5 \\
\hline
\textbf{Advise} &  \textbf{88.0} &80.0 & 88.0 & 68.0 \\
\hline
\textbf{Agree} &  \textbf{100.0} &33.3 & 100.0 & \textbf{100.0} \\
\hline
\textbf{Apologize} &  \textbf{88.9} &85.2 & 88.9& 81.5 \\
\hline
\textbf{Arrange} &  \textbf{68.2} &13.6 & 54.5& 45.4 \\
\hline
\textbf{Ask Help} &  70.0 &50.0 & 70.0 & \textbf{80.0} \\
\hline
\textbf{Care} &  52.6 &15.8 & 52.6 & 78.9 \\
\hline
\textbf{Comfort} &  \textbf{77.8} &33.3 & 66.7& 72.2 \\
\hline
\textbf{Complain} &  42.1 &7.0 & 42.1& \textbf{68.4} \\
\hline
\textbf{Criticize} &  \textbf{69.6} &21.7 & 60.9& 65.2 \\
\hline
\textbf{Flaunt} &  \textbf{90.0} &0.0 & 90.0 & 40.0 \\
\hline
\textbf{Greet} &  \textbf{91.7} &33.3 & 91.7& 75.0 \\
\hline
\textbf{Inform} &  \textbf{52.6} &10.5 & 54.4& 19.3 \\
\hline
\textbf{Introduce} &  \textbf{85.7} &19.0 & 85.7& \textbf{85.7} \\
\hline
\textbf{Joke} &  60.0 &10.0& 60.0 & \textbf{70.0} \\
\hline
\textbf{Leave} &  52.9 &52.9 & 52.9& \textbf{82.3} \\
\hline
\textbf{Oppose} &  70.0 &40.0 & 70.0 & \textbf{80.0} \\
\hline
\textbf{Praise} &  65.1 &18.6 & 65.1& \textbf{79.1} \\
\hline
\textbf{Prevent} &  66.7 &26.7 & 53.3& \textbf{73.3} \\
\hline
\textbf{Taunt} &  25.0 &0.0 & 25.0 & \textbf{33.3} \\
\hline
\textbf{Thank} &  96.0 &60.0 & 96.0 & \textbf{100.0} \\
\hline

\end{tabular}
\end{adjustbox}
\caption{10-shot Results on MIntRec-1 (Extended version of Table \ref{tab:few-shot-results})}
\label{tab:shot-10-mintrec1}
\end{table}

\begin{table}
\centering
\begin{adjustbox}{width=0.9\columnwidth} 

\begin{tabular}{| c  |c|c  |c  |c |}
\hline
 &  \textbf{M-7B} &\textbf{L2-7B} & \textbf{L3-8B} & \textbf{L2-13B} \\
\hline
\textbf{Accuracy} &  3.9 &2.9 & 0.3 & \textbf{10.1} \\
\hline
\textbf{Acknowledge} &  1.5 &4.4 & \textbf{76.5} & 1.5 \\
\hline
\textbf{Advise} &  3.1 &1.0 & 0.0 & \textbf{12.4} \\
\hline
\textbf{Agree} &  0.0 &\textbf{3.1} & 1.6 & 1.6 \\
\hline
\textbf{Apologize} &  4.6 &1.5 & 3.1 & \textbf{6.1} \\
\hline
\textbf{Arrange} &  \textbf{3.9} &\textbf{3.9} & 0.0 & 0.0 \\
\hline
\textbf{Ask Help} &  0.0 &0.0 & 0.0 & 0.0 \\
\hline
\textbf{Ask Opinions} &  0.0 &0.0 & 0.0 & 0.0 \\
\hline
\textbf{Care} &  \textbf{2.1} &0.0& \textbf{2.1} & 0.0 \\
\hline
\textbf{Comfort} &  6.1 &0.0 & 0.0 & \textbf{8.2} \\
\hline
\textbf{Complain} &  0.9 &0.0 & 0.9 & \textbf{5.6} \\
\hline
\textbf{Confirm} &  \textbf{3.2} &0.0 & 0.0 & 2.1 \\
\hline
\textbf{Criticize} &  0.0 &0.0 & 0.0 & \textbf{2.1 }\\
\hline
\textbf{Doubt} &  2.2 &0.7 & 0.0 & \textbf{5.2} \\
\hline
\textbf{Emphasize} &  5.9 &0.0 & 0.0 & \textbf{11.8} \\
\hline
\textbf{Explain} &  1.0 &0.0 & 0.0 & \textbf{4.7} \\
\hline
\textbf{Flaunt} &  \textbf{4.5} &0.0 & 0.0 & \textbf{4.5} \\
\hline
\textbf{Greet} &  \textbf{7.6} &1.5 & 4.5 & 1.5 \\
\hline
\textbf{Inform} &  4.6 &1.4 & 0.5 & \textbf{38.7} \\
\hline
\textbf{Introduce} &  40.0 &\textbf{66.7} & 3.3 & 38.3 \\
\hline
\textbf{Invite} &  0.0 &0.0 & 0.0 & \textbf{5.9} \\
\hline
\textbf{Joke} &  \textbf{6.9 }&0.0 & 0.0 & 3.4 \\
\hline
\textbf{Leave} &  3.8 &0.0 & 1.9 & \textbf{26.4} \\
\hline
\textbf{Oppose} &  \textbf{4.3 }&1.7 & 1.7 & \textbf{3.4} \\
\hline
\textbf{Plan} &  \textbf{2.6} &0.0 & 0.0 & \textbf{2.6} \\
\hline
\textbf{Praise} &  \textbf{4.4} &0.0 & 0.9 & \textbf{4.4} \\
\hline
\textbf{Prevent} &  0.0 &3.2 & 0.0 & \textbf{6.4} \\
\hline
\textbf{Refuse} &  \textbf{5.9} &0.0 & 0.0 & 0.0 \\
\hline
\textbf{Taunt} &  1.7 &1.7 & 0.0 & \textbf{18.9} \\
\hline
\textbf{Thank} &  0.0 &0.0 & 0.0 & 0.0 \\
\hline
\textbf{Warn} &  5.3 &5.3 & 0.0 & \textbf{42.1} \\
\hline
\end{tabular}
\end{adjustbox}
\caption{Zero-shot Results on MIntRec2.0 (Extended version of Table \ref{tab:few-shot-results})}
\label{tab:zero-mintrec2}
\vspace{-3mm}
\end{table}

\begin{table}
\centering
\begin{adjustbox}{width=0.8\columnwidth} 
\begin{tabular}{| c  |c|c  |c  |c |}
\hline
 &  \textbf{M-7B} &\textbf{L2-7B} & \textbf{L3-8B} & \textbf{L2-13B} \\
\hline
\textbf{Accuracy} &  \textbf{45.4} &39.1 & 29.4 & 14.1 \\
\hline
\textbf{Acknowledge} &  89.7 &58.8 & \textbf{92.6 }& 16.2 \\
\hline
\textbf{Advise} &  \textbf{56.7} &49.5 & 23.7 & 9.3 \\
\hline
\textbf{Agree} &  23.4 &\textbf{43.7} & 6.2 & 18.7 \\
\hline
\textbf{Apologize} &  \textbf{83.1} &67.7 & 15.4 & 41.5 \\
\hline
\textbf{Arrange} &  \textbf{45.1} &43.1 & 33.3 & 11.7 \\
\hline
\textbf{Ask Help} &  0.0 &\textbf{21.6} & 13.5 & 0.0 \\
\hline
\textbf{Ask Opinions} &  0.0 &10.4 & \textbf{12.5} & 4.2 \\
\hline
\textbf{Care} &  40.4 &\textbf{44.7} & 17.0 & 14.9 \\
\hline
\textbf{Comfort} &  42.9 &\textbf{46.9} & 22.4 & 2.0 \\
\hline
\textbf{Complain} &  \textbf{33.6} &29.9 & 14.0 & 14.0 \\
\hline
\textbf{Confirm} &  \textbf{41.9} &23.7 & 8.6 & 11.8 \\
\hline
\textbf{Criticize} &  \textbf{61.7} &55.3 & 10.6 & 12.8 \\
\hline
\textbf{Doubt} &  \textbf{51.1} &23.7 & 20.7 & 2.9 \\
\hline
\textbf{Emphasize} &  29.4 &\textbf{58.8} & 0.0 & 5.9 \\
\hline
\textbf{Explain} &  \textbf{52.8} &32.1 & 38.9 & 6.2 \\
\hline
\textbf{Flaunt} &  \textbf{45.4} &\textbf{45.4} & 27.3 & 9.1 \\
\hline
\textbf{Greet} &  66.7 &77.3 & \textbf{84.8} & 33.3 \\
\hline
\textbf{Inform} &  \textbf{33.6} &31.3 & 12.9 & 5.1 \\
\hline
\textbf{Introduce} &  \textbf{35.0} &23.3 & 23.3 & 10.0 \\
\hline
\textbf{Invite} &  47.1 &\textbf{52.9 }& 23.5 & 29.4 \\
\hline
\textbf{Joke} &  13.8 &\textbf{27.6} & 13.8 & 10.3 \\
\hline
\textbf{Leave} &  39.6 &49.1 & \textbf{58.5} & 28.3 \\
\hline
\textbf{Oppose} &  15.4 &\textbf{17.1} & 14.5 & 12.8 \\
\hline
\textbf{Plan} &  \textbf{81.6} &39.5 & 47.4 & 15.8 \\
\hline
\textbf{Praise} &  \textbf{56.6} &49.6 & 32.7 & 9.7 \\
\hline
\textbf{Prevent} &  48.4 &\textbf{64.5} & \textbf{64.5} & 25.8 \\
\hline
\textbf{Refuse} &  \textbf{58.8} &41.2 & 47.1 & 17.6 \\
\hline
\textbf{Taunt} &  \textbf{18.9} &5.2 & 13.8 & 17.2 \\
\hline
\textbf{Thank} &  93.1 &94.8 & \textbf{98.3} & 68.9 \\
\hline
\textbf{Warn} &  57.9 &52.6 & \textbf{63.2} & 26.3 \\
\hline

\end{tabular}
\end{adjustbox}
\caption{10-Shot Results on MIntRec2.0 (Extended version of Table \ref{tab:few-shot-results})}
\label{tab:shot-10-mintrec2}
\vspace{-5mm}
\end{table}

\subsection{Phase-2: Analysis of Modality Importance}
\label{subsection:phase_2_appendix}

\subsubsection{Wordcloud analysis}
\label{sec:wordcloud}
\begin{figure*}[t]
  \fbox{\includegraphics[width=0.3\linewidth]{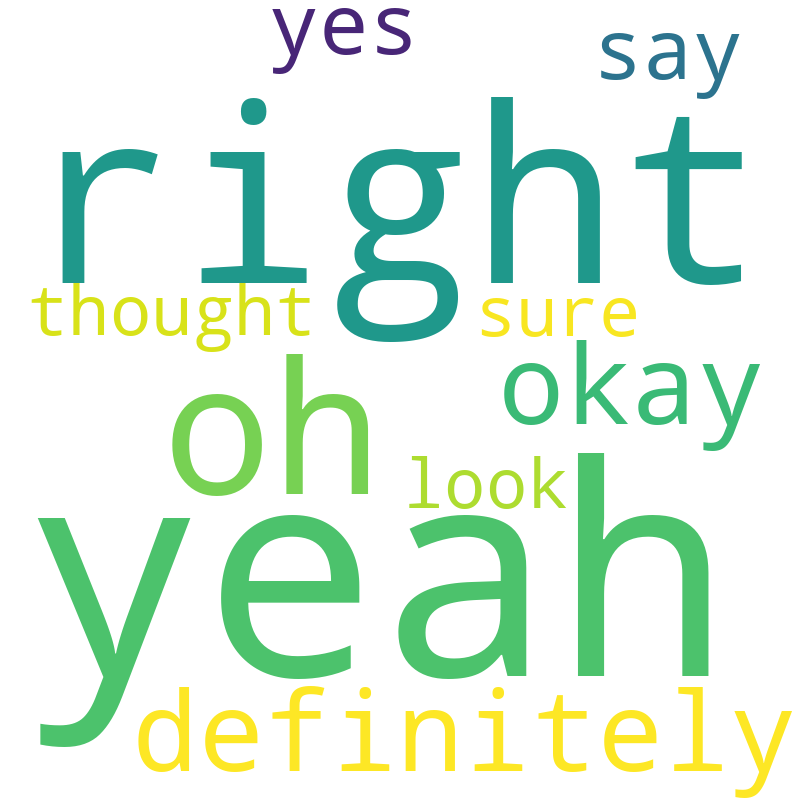}}\hspace{0.05cm}
  \fbox{\includegraphics[width=0.3\linewidth]{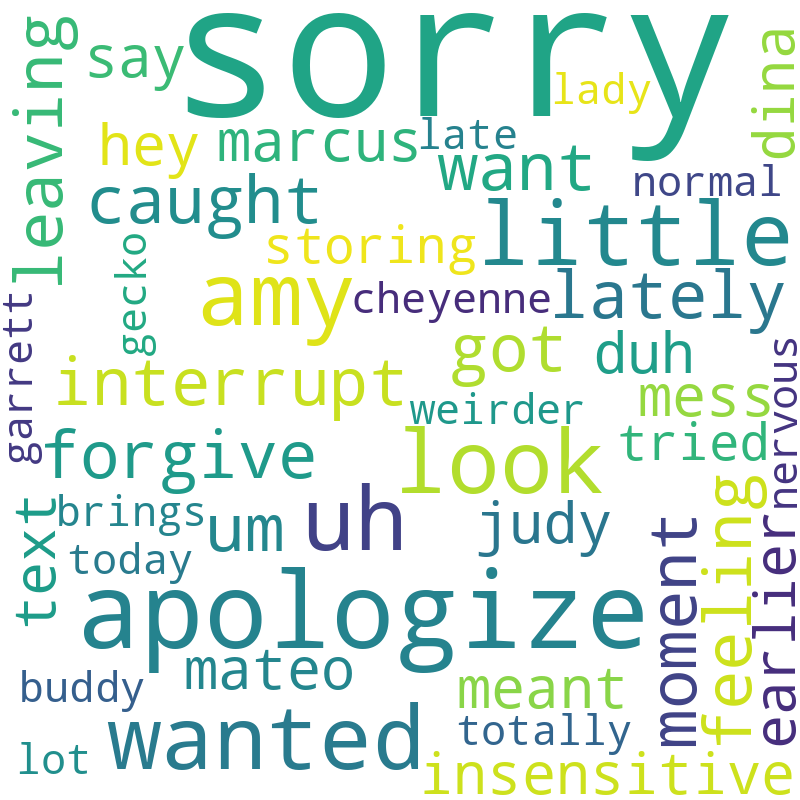}}\hspace{0.05cm}
  \fbox{\includegraphics[width=0.3\linewidth]{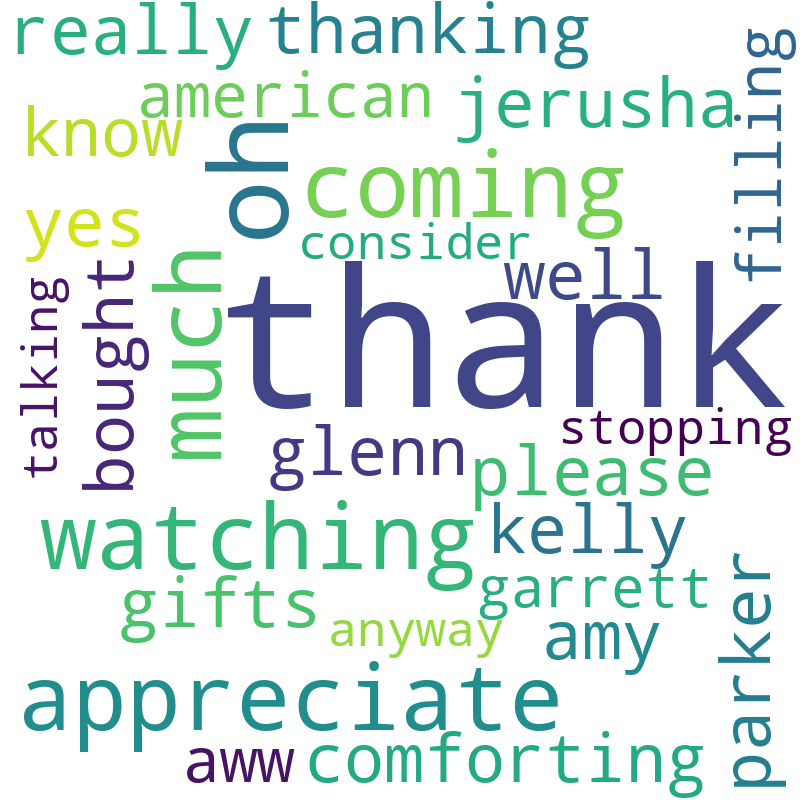}}
  \caption {WordCloud of Agree, Apologize, and Thank labels}
  \label{fig:wordcloud}
  \vspace{8mm}
\end{figure*}

All textual LLMs, such as LLaMA-2-7B, LLaMA-2-13B, LLaMA-3-8B, and Mistral-7B, perform consistently well on labels like \textit{Agree}, \textit{Apologize}, and \textit{Thank}. To understand this, we analyze word clouds for these labels (Fig. \ref{fig:wordcloud}). 

{For instance, the names ‘Judy,’ ‘Amy,’ and ‘Cheyenne’ appear in the dataset, with ‘Amy’ occurring 14 times in the test set across various intents: ‘Inform’ (4), ‘Greet’ (3), ‘Apologize’ (2), ‘Arrange’ (1), ‘Thank’ (1), ‘Care’ (1), ‘Comfort’ (1), and ‘Complain’ (1). The worst-performing LLM model on MIntRec-1, LLaMA-3-8B, correctly classifies 10 out of these 14 instances. Word clouds highlight phrases such as ‘sorry’ and ‘apologize,’ which contribute to accurate intent classification. This confirms that the models categorize intents based on content rather than specific character names, ensuring the dataset does not introduce character-based bias.}

\subsubsection{Modality Ablation Analysis}
\label{sec:modality_ablation_analysis}

\begin{table}
\centering
\begin{adjustbox}{width=\linewidth}  
\begin{tabular}{| c  |c  |c c c |c c c|}
\hline
 & \textbf{T+V+A}& \textbf{T}& \textbf{V}& \textbf{A}& \textbf{T+V}& \textbf{T+A}&\textbf{V+A}\\
\hline
\textbf{Acc}& \textbf{72.8}& 69.9& 17.8& 28.5& 69.7& 70.8&25.2\\
\hline
\textbf{Advise}& \textbf{80.0}& 60.0& 0.0& 0.0& 50.0& 50.0&0.0\\
\hline
\textbf{Agree}& \textbf{100.0}& 63.2& 61.4& 40.4& 59.7& 61.4&49.1\\
\hline
\textbf{Apologize}& \textbf{96.3}& 81.4& 11.6& 48.8& 79.1& 86.1&41.9\\
\hline
\textbf{Arrange}& 68.2& \textbf{96.3}& 7.4& 63.0& \textbf{96.3}& \textbf{96.3}&55.6\\
\hline
\textbf{Ask Help}& 50.0& \textbf{100.0}& 0.0& 64.0& \textbf{100.0}& \textbf{100.0}&56.0\\
\hline
\textbf{Care}& \textbf{84.2}& 39.1& 0.0& 13.0& 43.5& 52.2&4.4\\
\hline
\textbf{Comfort}& \textbf{77.8}& 25.0& 0.0& 0.0& 25.0& 16.7&8.3\\
\hline
\textbf{Complain}& 66.7& 91.7& 0.0& 8.3& 91.7& \textbf{100.0}&8.3\\
\hline
\textbf{Criticize}& 39.1& 30.0& 0.0& 0.0& 30.0& \textbf{50.0}&0.0\\
\hline
\textbf{Flaunt}& \textbf{60.0}& \textbf{60.0}& 0.0& 0.0& 30.0& 30.0&0.0\\
\hline
\textbf{Greet}& \textbf{83.3}& 40.0& 0.0& 10.0& 30.0& 30.0&10.0\\
\hline
\textbf{Inform}& 70.2& 72.2& 0.0& 16.7& \textbf{77.8}& \textbf{77.8}&5.6\\
\hline
\textbf{Introduce}& 57.1& 89.5& 0.0& 5.3& \textbf{94.7}& 89.5&5.3\\
\hline
\textbf{Joke}& 60.0& 61.4& 61.4& 40.4& \textbf{73.7}& 68.4&28.1\\
\hline
\textbf{Leave}& \textbf{82.4}& 76.0& 0.0& 16.0& 76.0& 72.0&8.0\\
\hline
\textbf{Oppose}& 40.0& \textbf{72.7}& 0.0& 9.1& 68.2& \textbf{72.7}&9.1\\
\hline
\textbf{Praise}& \textbf{79.1}& 66.7& 0.0& 4.8& 61.9& 76.2&9.5\\
\hline
\textbf{Prevent}& \textbf{80.0}& 82.4& 0.0& 11.8& 70.6& 64.7&5.9\\
\hline
\textbf{Taunt}& 16.7& \textbf{73.3}& 0.0& 20.0& \textbf{73.3}& 66.7&20.0\\
\hline
\textbf{Thank}& \textbf{96.0}& 66.7& 16.7& 50.0& 75.0& 75.0&41.7\\
\hline

\end{tabular}
\end{adjustbox}
\caption{Ablation analysis for MIntRec-1. SDIF is the best-performing small multimodal model (Extended version of Table \ref{tab:mintrec-1-disection-mini})}
\label{tab:mintrec-1-disection}
\end{table}

\begin{table}
\centering
\begin{adjustbox}{width=\linewidth}  
\begin{tabular}{| c  |c  |c  c  c |c c c|}
\hline
 & \textbf{T+V+A}&\textbf{ T}& \textbf{V}& \textbf{A}& \textbf{T+V}& \textbf{T+A}&\textbf{V+A}\\
\hline
\textbf{Acc}& \textbf{58.4}& 56.7& 12.4& 23.3& 56.4& 56.4&24.6\\
\hline
\textbf{Acknowledge}& 65.9& 69.1& 0.0& 61.8& \textbf{70.6}& \textbf{70.6}&48.5\\
\hline
\textbf{Advise}& \textbf{60.2}& 50.5& 2.1& 1.0& 53.6& 53.6&15.5\\
\hline
\textbf{Agree}& \textbf{66.6}& 65.6& 0.0& 28.1& 57.8& 57.8&34.4\\
\hline
\textbf{Apologize}& \textbf{94.5}& 90.8& 0.0& 53.9& 92.3& 92.3&58.5\\
\hline
\textbf{Arrange}& \textbf{65.1}& 64.7& 2.0& 0.0& 60.8& 60.8&7.8\\
\hline
\textbf{Ask Help}& 57.8& 62.2& 0.0& 0.0& \textbf{64.9}& \textbf{64.9}&2.7\\
\hline
\textbf{Ask Opinions}& \textbf{57.9}& 50.0& 0.0& 4.2& 47.9& 47.9&10.4\\
\hline
\textbf{Care}& 55.3& \textbf{57.5}& 0.0& 2.1& 46.8& 46.8&6.4\\
\hline
\textbf{Comfort}& 44.5& \textbf{49.0}& 2.0& 0.0& 44.9& 44.9&0.0\\
\hline
\textbf{Complain}& \textbf{45.8}& 43.9& 7.5& 0.0& 38.3& 38.3&11.2\\
\hline
\textbf{Confirm}& \textbf{49.9}& 46.2& 0.0& 15.1& 45.2& 45.2&4.3\\
\hline
\textbf{Criticize}& 40.9& 38.3& 0.0& 0.0& \textbf{48.9}& \textbf{48.9}&4.3\\
\hline
\textbf{Doubt}& 64.0& 62.2& 49.6& 26.7& \textbf{68.2}& \textbf{68.2}&33.3\\
\hline
\textbf{Emphasize}& 2.3& \textbf{11.8}& 0.0& 0.0& 5.9& 5.9&0.0\\
\hline
\textbf{Explain}& \textbf{63.2}& 58.0& 16.6& 31.6& 62.2& 62.2&40.4\\
\hline
\textbf{Flaunt}& 15.4& 13.6& 0.0& 0.0& \textbf{22.7}& \textbf{22.7}&0.0\\
\hline
\textbf{Greet}& \textbf{86.4}& 80.3& 12.1& 31.8& 80.3& 80.3&42.4\\
\hline
\textbf{Inform}& 55.0& 55.3& 58.1& \textbf{62.2}& 56.2& 56.2&37.8\\
\hline
\textbf{Introduce}& \textbf{40.7}& 38.3& 0.0& 1.7& 38.3& 38.3&10.0\\
\hline
\textbf{Invite}& \textbf{38.8}& 29.4& 0.0& 0.0& 17.7& 17.7&0.0\\
\hline
\textbf{Joke}& 4.1& 6.9& 0.0& 0.0& \textbf{13.8}& \textbf{13.8}&0.0\\
\hline
\textbf{Leave}& 50.9& \textbf{56.6}& 0.0& 11.3& \textbf{56.6}& \textbf{56.6}&24.5\\
\hline
\textbf{Oppose}&\textbf{ 73.8}& 67.5& 3.4& 52.1& 70.1& 70.1&41.0\\
\hline
\textbf{Plan}& 46.3& \textbf{57.9}& 0.0& 0.0& 44.7& 44.7&10.5\\
\hline
\textbf{Praise}& \textbf{77.8}& 72.6& 3.5& 9.7& 67.3& 67.3&18.6\\
\hline
\textbf{Prevent}&\textbf{ 66.4}& 61.3& 0.0& 0.0& 64.5& 64.5&3.2\\
\hline
\textbf{Refuse}& 12.9& \textbf{29.4}& 0.0& 0.0& \textbf{29.4}& \textbf{29.4}&0.0\\
\hline
\textbf{Taunt}& 25.2& \textbf{25.9}& 0.0& 0.0& 13.8& 13.8&8.6\\
\hline
\textbf{Thank}& 96.5& \textbf{96.6}& 0.0& 48.3& \textbf{96.6}& \textbf{96.6}&53.5\\
\hline
\textbf{Warn}& 15.8& \textbf{26.3}& 0.0& 0.0& 21.1& 21.1&0.0\\
\hline

\end{tabular}
\end{adjustbox}
\caption{Ablation analysis for MIntRec2.0. MulT is the best-performing small multimodal model on MIntRec2.0}
\label{tab:mintrec-2-disection}
\vspace{-2mm}
\end{table}

\begin{figure*}
\centering
\fbox{\includegraphics[width=0.8\textwidth]{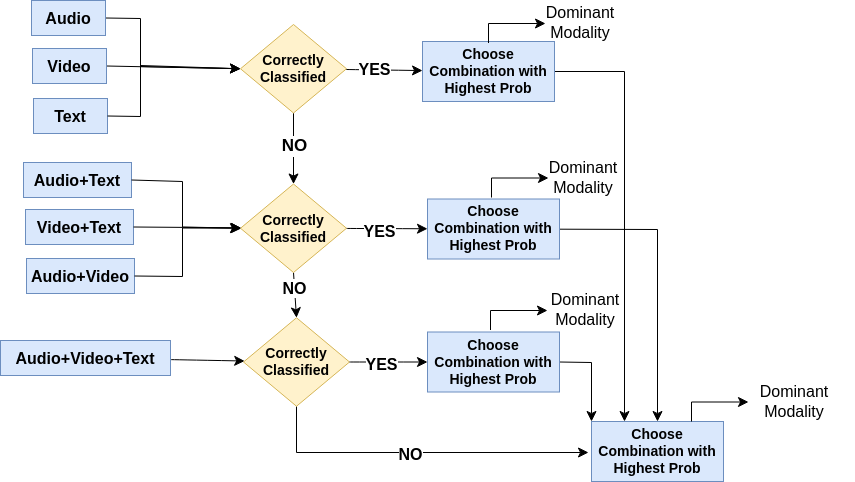}}
  \caption{Modality Annotation Algorithm} 
\label{fig:modality_algo}
\vspace{-5mm}
\end{figure*}

\begin{figure} 
\centering 
\includegraphics[width=\columnwidth]{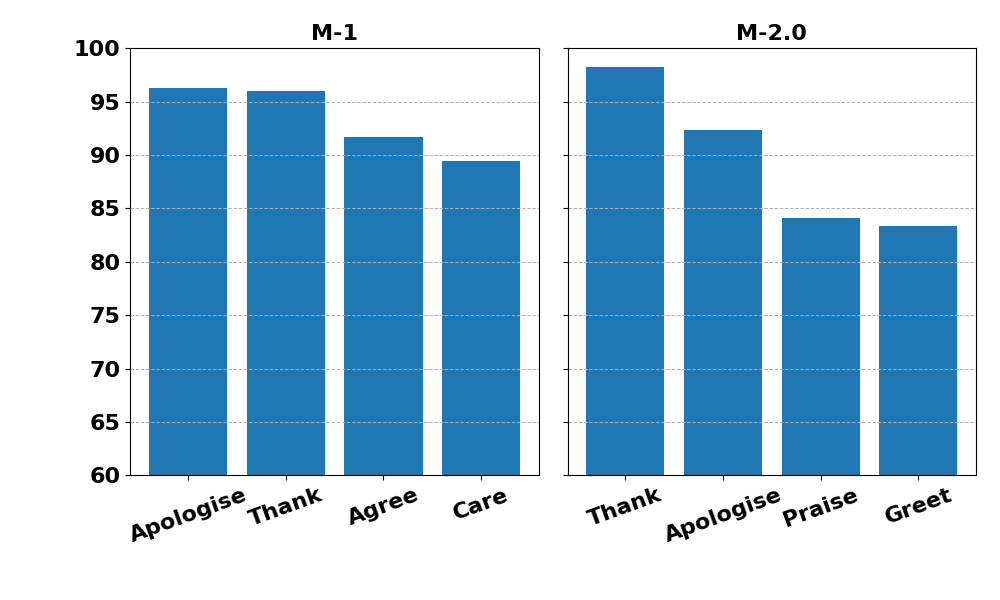} \caption{Textually Biased Labels} 
\label{fig:text_needed} 
\end{figure} 

\begin{figure} 
\centering 
\includegraphics[width=\columnwidth]{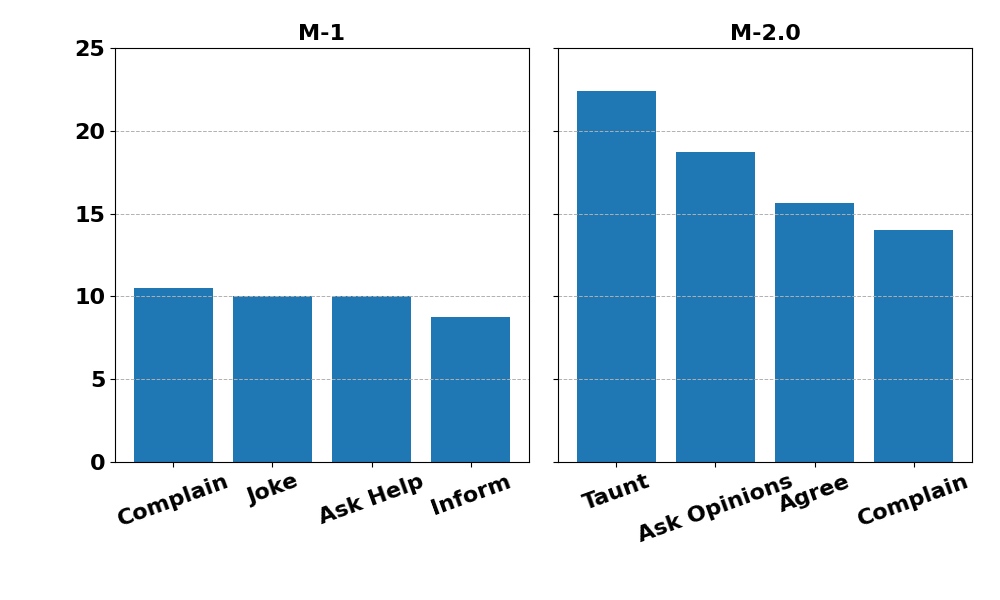} \caption{Visually Biased Labels} 
\label{fig:video_needed} 
\end{figure}

\begin{figure}
\centering
\includegraphics[width=\columnwidth]{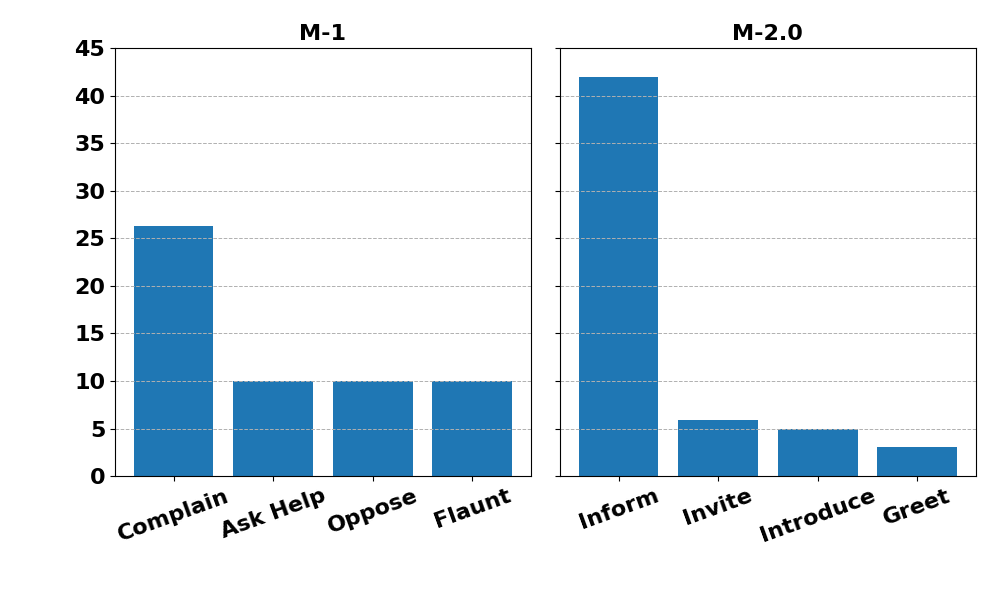}
\caption{Labels that benefit from audio cues}
\label{fig:audio_needed}
\end{figure}

\begin{figure}
\centering
\fbox{\includegraphics[width=\columnwidth]{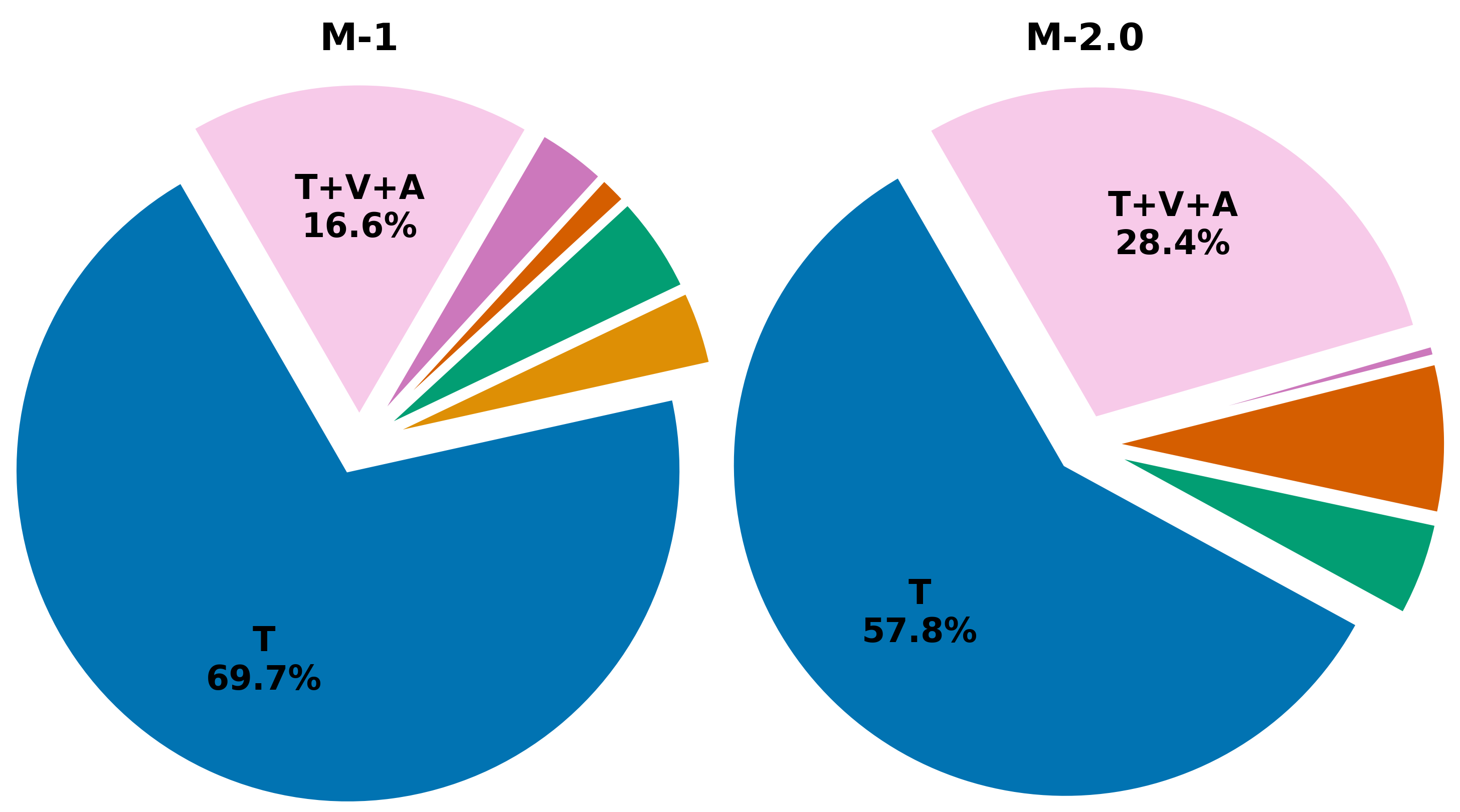}}
  \caption{Distribution of Modality Importance \\
        \textcolor{te}{\rule{10pt}{10pt}} \textbf{T}ext, 
        \textcolor{au}{\rule{10pt}{10pt}} \textbf{A}udio, 
        \textcolor{vi}{\rule{10pt}{10pt}} \textbf{V}ideo,
        \textcolor{tv}{\rule{10pt}{10pt}} \textbf{T}ext\textbf{+V}ideo, 
        \textcolor{ta}{\rule{10pt}{10pt}} \textbf{T}ext\textbf{+A}udio,
        \textcolor{av}{\rule{10pt}{10pt}} \textbf{A}udio\textbf{+V}ideo,
        \textcolor{tva}{\rule{10pt}{10pt}} \textbf{T}ext\textbf{+V}ideo\textbf{+A}udio} 
\label{fig:mintrec_modality_importance}
\end{figure}

The importance of individual modalities for different intent categories is summarized in Tables \ref{tab:mintrec-1-disection} and \ref{tab:mintrec-2-disection}. Figures \ref{fig:text_needed} illustrate that over 80\% of the samples for intents such as \textit{Apologize}, \textit{Thank}, and \textit{Praise} exhibit a strong textual bias. On the other hand, only a small fraction of samples for intents like \textit{Joke}, \textit{Taunt}, and \textit{Asking for opinions} require the video modality for correct classification (Figure \ref{fig:video_needed}. Similarly \textit{Inform}, \textit{Complain} might benefit from the presence of audio cues (Fig. \ref{fig:audio_needed}).

\subsubsection{Modality Combination Annotation}
Fig. \ref{fig:modality_algo} shows the algorithm used to label MIntRec-1 and MIntRec2.0 with the smallest subset of modalities needed to classify each instance correctly.

\subsection{Phase-3: Debiasing MIntRec Datasets}
\label{subsection:phase_3_appendix}

\subsubsection{Debiasing Analysis}
\begin{figure*}
\centering
\fbox{\includegraphics[width=0.85\linewidth]{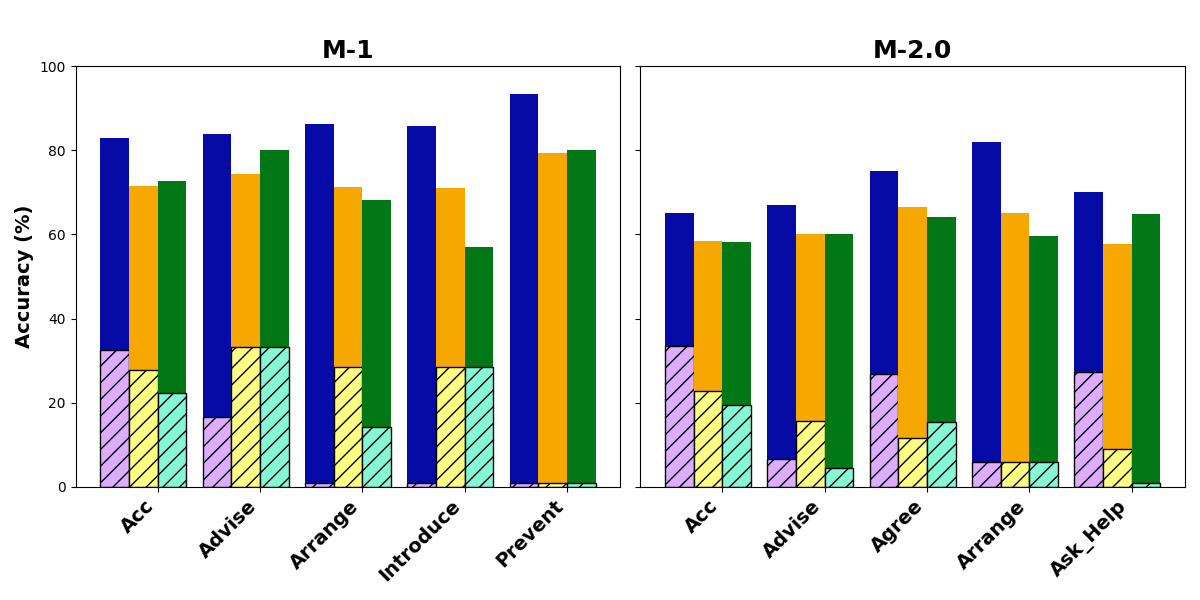}}
\caption{Performance Before vs After Debiasing \\
        \textcolor{M7B}{\rule{10pt}{10pt}} Before - M-7B, 
        \textcolor{MulT}{\rule{10pt}{10pt}} Before - MulT,
        \textcolor{SDIF}{\rule{10pt}{10pt}} Before - SDIF \\
        \textcolor{M7B-After}{\rule{10pt}{10pt}} After - M-7B, 
        \textcolor{MulT-After}{\rule{10pt}{10pt}} After - MulT, 
        \textcolor{SDIF-After}{\rule{10pt}{10pt}} After - SDIF}
\label{fig:mintrec_debiasing}
\end{figure*}

Tables \ref{tab:mintrec-1-debiasing} and \ref{tab:mintrec-2-debiasing} present the number of samples before and after debiasing across various intent levels. Tables \ref{tab:extended-mintrec-1-debiased-result} and \ref{tab:extended-mintrec-2-debiased-result} summarize the performance of various models following the debiasing process. Fig. \ref{fig:mintrec_debiasing} shows reduced performance for both small multimodal models and textual LLMs on the debiased datasets. The results before vs. after the debiasing are summarized in Tables \ref{tab:combined_mintrec_1_table} and \ref{tab:combined_mintrec_2_table}.

\noindent \textbf{Modality Ablation on Debiased Dataset}
We also perform a modality ablation analysis on the debiased dataset (Tables~\ref{tab:mintrec-1-disection-debiased} and~\ref{tab:mintrec-2-disection-debiased}), following the methodology described in Section~\ref{sec:modality_ablation_analysis}. The results reveal that the performance gap between text-inclusive combinations and combinations without text is reduced, indicating that the textual bias has been mitigated. Notably, several intent categories, such as \textit{Care}, \textit{Arrange}, and \textit{Agree}, which previously benefited from text-based combinations, now exhibit diminished performance.

\begin{table}
\centering
\begin{adjustbox}{width=0.8\columnwidth} 
\begin{tabular}{| c  |c  |c  |c |}
\hline
\textbf{Intent} & \textbf{ Before}& \textbf{ After}& \textbf{\% Reduction} \\
\hline
\textbf{Thank} & 124 & 2 & 98.39 \\
\hline
\textbf{Apologize} & 136 & 4 & 97.06 \\
\hline
\textbf{Greet} & 60 & 6 & 90.00 \\
\hline
\textbf{Agree} & 59 & 7 & 88.14 \\
\hline
\textbf{Praise} & 213 & 26 & 87.79 \\
\hline
\textbf{Care} & 95 & 13 & 86.32 \\
\hline
\textbf{Comfort} & 88 & 18 & 79.55 \\
\hline
\textbf{Advise} & 122 & 26 & 78.69 \\
\hline
\textbf{Complain} & 286 & 66 & 76.92 \\
\hline
\textbf{Prevent} & 73 & 17 & 76.71 \\
\hline
\textbf{Leave} & 85 & 24 & 71.76 \\
\hline
\textbf{Inform} & 284 & 90 & 68.31 \\
\hline
\textbf{Arrange} & 110 & 35 & 68.18 \\
\hline
\textbf{Introduce} & 105 & 34 & 67.62 \\
\hline
\textbf{Oppose} & 51 & 18 & 64.71 \\
\hline
\textbf{Criticize} & 117 & 47 & 59.83 \\
\hline
\textbf{Ask Help} & 51 & 21 & 58.82 \\
\hline
\textbf{Flaunt} & 52 & 22 & 57.69 \\
\hline
\textbf{Joke} & 51 & 37 & 27.45 \\
\hline
\textbf{Taunt} & 62 & 50 & 19.35 \\
\hline
\textbf{Total}  & 2224 & 563 & 74.69 \\
\hline
\end{tabular}
\end{adjustbox}
\caption{MIntRec-1 Debiasing Statistics}
\label{tab:mintrec-1-debiasing}
\end{table}

\begin{table}
\centering
\begin{adjustbox}{width=0.8\columnwidth} 
\begin{tabular}{| c  |c  |c  |c |}
\hline
\textbf{Intent} & \textbf{ Before}& \textbf{ After}& \textbf{\% Reduction} \\
\hline
\textbf{Thank} & 281 & 14 & 95.02 \\
\hline
\textbf{Apologize} & 271 & 23 & 91.51 \\
\hline
\textbf{Greet} & 313 & 47 & 84.98 \\
\hline
\textbf{Praise} & 472 & 109 & 76.91 \\
\hline
\textbf{Oppose} & 508 & 183 & 63.98 \\
\hline
\textbf{Advise} & 375 & 145 & 61.33 \\
\hline
\textbf{Agree} & 343 & 140 & 59.18 \\
\hline
\textbf{Explain} & 765 & 315 & 58.82 \\
\hline
\textbf{Doubt} & 687 & 298 & 56.62 \\
\hline
\textbf{Comfort} & 233 & 104 & 55.36 \\
\hline
\textbf{Arrange} & 259 & 116 & 55.21 \\
\hline
\textbf{Confirm} & 471 & 218 & 53.72 \\
\hline
\textbf{Inform} & 926 & 433 & 53.24 \\
\hline
\textbf{Prevent} & 128 & 60 & 53.13 \\
\hline
\textbf{Plan} & 188 & 91 & 51.60 \\
\hline
\textbf{Introduce} & 310 & 161 & 48.06 \\
\hline
\textbf{Care} & 222 & 117 & 47.30 \\
\hline
\textbf{Complain} & 512 & 272 & 46.88 \\
\hline
\textbf{Acknowledge} & 307 & 170 & 44.63 \\
\hline
\textbf{Ask Help} & 155 & 87 & 43.87 \\
\hline
\textbf{Leave} & 235 & 134 & 42.98 \\
\hline
\textbf{Criticize} & 198 & 132 & 33.33 \\
\hline
\textbf{Invite} & 109 & 74 & 32.11 \\
\hline
\textbf{Ask Opinions} & 251 & 174 & 30.68 \\
\hline
\textbf{Warn} & 99 & 70 & 29.29 \\
\hline
\textbf{Refuse} & 94 & 73 & 22.34 \\
\hline
\textbf{Taunt} & 274 & 228 & 16.79 \\
\hline
\textbf{Flaunt} & 93 & 80 & 13.98 \\
\hline
\textbf{Emphasize} & 94 & 86 & 8.51 \\
\hline
\textbf{Joke} & 131 & 122 & 6.87 \\
\hline
\textbf{Total} & 9304 & 4276 & 54.04 \\
\hline

\end{tabular}
\end{adjustbox}
\caption{MIntRec2.0 Debiasing Statistics}
\label{tab:mintrec-2-debiasing}
\vspace{-5mm}
\end{table}

\begin{figure*}
\centering
\fbox{\includegraphics[width=0.85\linewidth]{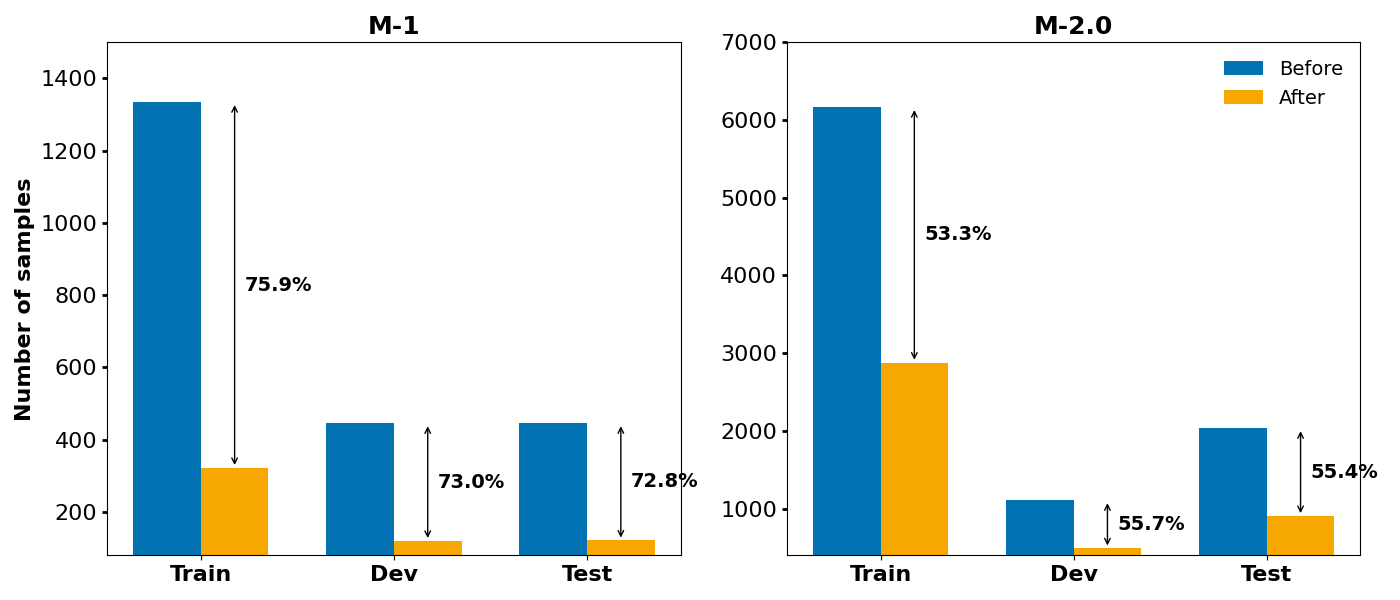}}
  \caption{Before vs After Statistics of MIntRec Debiasing}
\label{fig:debiasing}
\end{figure*}


\begin{table*}
\centering
\begin{adjustbox}{width=\linewidth}
\begin{tabular}{| c  |c  |c c  c c c  |c c c c  |c c c c c |}
\hline
 & \textbf{BERT}& \textbf{M-7B}& \textbf{L-7B} &\textbf{Q-7B}& \textbf{L-8B}&\textbf{ L-13B}& \textbf{MulT}& \textbf{MAG}& \textbf{MISA}& \textbf{SDIF}&  \textbf{ClaudeT}& \textbf{ClaudeV}&\textbf{GPT-4T}& \textbf{GPT-4V}& \textbf{VChatGPT}\\
\hline
\textbf{Advise}& 16.7& 16.7& 33.3 &\textbf{66.7}& {50.0}& 33.3& 33.3& 33.3& 0.0& 33.3&  33.3& 16.7&14.3& 0.0& 0.0\\
\hline
\textbf{Arrange}&  0.0&  0.0& 14.3 &\textbf{42.9}& {28.6}& 14.3& {28.6}&  0.0& 0.0& 14.3&  14.3& {28.6}&20.0& 14.3& 0.0\\
\hline
\textbf{Ask Help}&  0.0& 66.7& 66.7 &66.7& 33.3& 33.3& 33.3& 33.3& 0.0& 0.0&  \textbf{100.0}& 33.3&25.0& 33.3& 0.0\\
\hline
\textbf{Comfort}&  0.0&  0.0&  0.0 &50.0& 25.0&  0.0& 25.0&  0.0& 0.0&  0.0&  0.0& 0.0&28.6& \textbf{42.9}& 0.0\\
\hline
\textbf{Complain}& 30.0& \textbf{75.0}& 20.0 &40.0& 20.0& 60.0& 40.0& 15.0& 20.0& 35.0&  45.0& 45.0&33.3& 50.0& 50.0\\
\hline
\textbf{Criticize}& 30.0& 40.0& 20.0 &40.0& 30.0& \textbf{70.0}& 30.0& 40.0& 40.0& 20.0&  60.0& 50.0&33.3& 50.0& 22.2\\
\hline
\textbf{Flaunt}&  0.0&  0.0& \textbf{50.0} &16.7& 33.3& 33.3& 16.7&  0.0& 0.0& 16.7&  \textbf{50.0}& 33.3&33.3& \textbf{50.0}& 40.0\\
\hline
\textbf{Inform}& 31.3& {68.8}& 18.8 &18.8& 12.5& 43.8& 50.0& 43.8& \textbf{81.3}& 37.5&  31.3& 50.0&43.8& 55.0&  0.0\\
\hline
\textbf{Introduce}& 14.3&  0.0& 57.1 &42.9& 42.9& 28.6& 28.6& 14.3&  0.0& 28.6&  0.0& 0.0&50.0& \textbf{60.0}&  0.0\\
\hline
\textbf{Joke}& 33.3& 33.3& 83.3 &66.7& 50.0& 50.0& 33.3& 16.7& 16.7& 33.3&  50.0& 16.7&50.0& \textbf{62.5}& 16.7\\
\hline
\textbf{Leave}&  0.0&  0.0& 25.0 &25.0&  0.0& 50.0&  0.0&  0.0&  0.0&  0.0&  0.0& 0.0&60.0& \textbf{66.7}&  0.0\\
\hline
\textbf{Oppose}&  0.0&  0.0& 50.0 &0.0& 50.0& 16.7&  0.0&  0.0&  0.0&  0.0&  33.3& 33.3&\textbf{66.7}& \textbf{66.7}& 40.0\\
\hline
\textbf{Prevent}&  0.0&  0.0&  0.0 &0.0&  0.0&  0.0&  0.0&  0.0&  0.0&  0.0&  66.7& 0.0&\textbf{80.0}& 66.7&  0.0\\
\hline
\textbf{Taunt}&  0.0&  0.0& 30.0 &30.0&  0.0&  0.0&  0.0& 20.0& 20.0& 10.0&  10.0& 10.0&\textbf{100.0}& \textbf{100.0}& 14.3\\
\hline
\end{tabular}
\end{adjustbox}
\caption{Extended Debiased MIntRec-1 Results (Accuracy in \%; MulT, MISA, MAG and SDIF are small multimodal models; L2-7B, L2-13B, M-7B, Q-7B and L3-8B are textual LLMs; VChatGPT is a multimodal LLM)(Extended version of Table \ref{tab:results-mintrec-debiased})}
\label{tab:extended-mintrec-1-debiased-result}
\end{table*}

\begin{table*}
\centering
\begin{adjustbox}{width=\linewidth}
\begin{tabular}{| c  |c   |c c c c c   |c  c c  c  |c c c  c  c |}
\hline
 & \textbf{BERT}& \textbf{M-7B}& \textbf{L-7B} & \textbf{Q-7B}& \textbf{L-8B}& \textbf{L-13B}& \textbf{MulT}& \textbf{MAG}& \textbf{MISA}& \textbf{SDIF}&  \textbf{ClaudeT}& \textbf{ClaudeV}&\textbf{GPT-4T}& \textbf{GPT-4V}& \textbf{VChatGPT}\\
\hline
\textbf{Acknowledge}& 41.5& 78.1& 56.1 &78.1& 41.5& 61.0& 63.4& 41.5& 63.4& 48.8&  73.2& 73.2&0.0& 0.0& \textbf{81.6}\\
\hline
\textbf{Advise}& 13.3& 6.7& {42.2} &\textbf{55.6}& 26.7& 33.3& 15.6& 15.6& 26.7& 4.4&  26.7& 26.7&0.0& 3.9& 0.0\\
\hline
\textbf{Agree}& 26.9& 26.9& 23.1 &26.9& 23.1& 23.1& 11.5& 38.5& 0.0& 15.4&  34.6& \textbf{65.4}&2.6& 4.8& 0.0\\
\hline
\textbf{Arrange}& 5.9& 5.9& 5.9 &11.8& 0.0& \textbf{23.5}& 5.9& 17.7& 17.7& 5.9&  6.3& 6.3&5.9& 5.1& 0.0\\
\hline
\textbf{Ask Help}& 9.1& 27.3& 36.4 &36.4& 36.4& 9.1& 9.1& 9.1& 27.3& 0.0&  54.6& \textbf{63.6}&6.1& 8.3& 0.0\\
\hline
\textbf{Ask Opinions}& 28.0& 36.0& 28.0 &\textbf{72.0}& {60.0}& 28.0& 16.0& 40.0& 28.0& 40.0&  36.0& 36.0&7.7& 12.0& 0.0\\
\hline
\textbf{Care}& 11.5& 7.7& 23.1 &7.7& \textbf{26.9}& \textbf{26.9}& 19.2& 11.5& 19.2& 23.1&  4.0& 4.0&9.4& 12.2& 0.0\\
\hline
\textbf{Comfort}& 4.0& \textbf{36.0}& 24.0 &4.0& \textbf{36.0}& \textbf{36.0}& 8.0& 0.0& 20.0& 8.0&  4.2& 4.2&11.1& 22.2& 9.1\\
\hline
\textbf{Complain}& 11.9& 32.2& \textbf{57.6} &35.6& 44.1& 42.4& 8.5& 17.0& 18.6& 22.0&  39.0& 39.0&11.5& 22.2& 34.6\\
\hline
\textbf{Confirm}& 28.6& 2.0& 22.5 &14.3& 38.8& 24.5& 16.3& \textbf{44.9}& 34.7& 24.5&  8.7& 8.7&12.0& 23.1& 3.2\\
\hline
\textbf{Criticize}& 20.6& 2.9& 41.2 &17.7& 50.0& 44.1& 29.4& 14.7& 14.7& 23.5&  \textbf{57.6}& \textbf{57.6}&16.5& 23.5& 23.3\\
\hline
\textbf{Doubt}& 17.7& \textbf{64.7}& 49.0 &21.6& 31.4& 51.0& 54.9& 23.5& 19.6& 19.6&  21.6& 21.6&20.0& 25.0& 55.6\\
\hline
\textbf{Emphasize}& 6.7& 13.3& 6.7 &6.7& 20.0& \textbf{33.3}& 6.7& 6.7& 0.0& 0.0&  20.0& 20.0&23.7& 26.7& 7.1\\
\hline
\textbf{Explain}& 24.7& 1.2& 34.1 &\textbf{45.9}& 35.3& 30.6& 32.9& 22.4& {37.7}& 18.8&  8.6& 8.6&23.8& 29.4& 1.5\\
\hline
\textbf{Flaunt}& 0.0& {38.9}& 27.8 &\textbf{44.4}& 5.6& 33.3& 11.1& 0.0& 5.6& 11.1&  27.8& 27.8&25.5& 30.6& 0.0\\
\hline
\textbf{Inform}& 31.3& \textbf{84.4}& 29.2 &27.1& 37.5& 34.4& 35.4& 19.8& 17.7& 34.4&  29.4& 30.4&33.3& 31.3& 0.0\\
\hline
\textbf{Introduce}& 11.1& 13.9& 25.0 &19.4& 22.2& 27.8& 22.2& 22.2& 8.3& 11.1&  5.7& 5.7&\textbf{36.4}& 34.8& 6.9\\
\hline
\textbf{Invite}& 16.7& \textbf{75.0}& 25.0 &41.7& 41.7& 33.3& 25.0& 16.7& 33.3& 25.0&  66.7& 66.7&37.0& 36.6& 0.0\\
\hline
\textbf{Joke}& 12.0& 28.0& 12.0 &24.0& 4.0& 8.0& 4.0& 4.0& 12.0& 8.0&  29.2& 29.2&40.0& \textbf{44.0}& 0.0\\
\hline
\textbf{Leave}& 22.2& 25.9& 14.8 &0.0& 14.8& 11.1& 14.8& 29.6& 14.8& 18.5&  0.0& 0.0&44.0& \textbf{46.2}& 0.0\\
\hline
\textbf{Oppose}& 20.5& 23.1& 12.8 &5.1& 23.1& 12.8& 15.4& 28.2& 25.6& 15.4&  5.1& 5.1&46.3& \textbf{50.0}& 2.9\\
\hline
\textbf{Plan}& 14.3& 47.6& 9.5 &19.1& 38.1& 19.1& 4.8& 19.1& 9.5& 9.5&  25.0& 25.0&48.0& \textbf{55.6}& 7.1\\
\hline
\textbf{Praise}& 4.2& 45.8& 41.7 &20.8& 41.7& 50.0& 4.2& 4.2& 12.5& 8.3&  17.4& 17.4&50.0& \textbf{56.0}& 17.4\\
\hline
\textbf{Prevent}& 18.2& 27.3& 0.0 &18.2& 18.2& 9.1& 36.4& 0.0& 9.1& 0.0&  0.0& 0.0&53.9& \textbf{59.3}& 0.0\\
\hline
\textbf{Refuse}& 0.0& 21.4& 28.6 &\textbf{71.4}& 35.7& 21.4& 7.1& 0.0& 0.0& 7.1&  35.7& 42.9&58.8& {64.3}& 9.1\\
\hline
\textbf{Taunt}& 8.7& 43.5& 8.7 &54.4& 23.9& 23.9& 17.4& 28.3& 15.2& 17.4&  17.4& 17.4&66.7& \textbf{72.7}& 7.7\\
\hline
\textbf{Warn}& 0.0& 30.8& 38.5 &23.1& 46.2& 30.8& 0.0& 0.0& 0.0& 7.7&  30.8& 30.8&71.4& \textbf{91.7}& 0.0\\
\hline
\end{tabular}
\end{adjustbox}
\caption{Extended Debiased MIntRec2.0 Results (Accuracy in \%; MulT, MISA, MAG and SDIF are small multimodal models; L2-7B, L2-13B, M-7B, Q-7B and L3-8B are textual LLMs; VChatGPT is a multimodal LLM)(Extended version of Table \ref{tab:results-mintrec-debiased})}
\label{tab:extended-mintrec-2-debiased-result}
\end{table*}

\begin{table*}
\centering
\begin{adjustbox}{width=\textwidth}
\begin{tabular}{|c| c | c|c| c|c c|c c|c c|c c|c c|}
\hline
\multirow{2}{*}{\textbf{Model}} & \multicolumn{2}{|c|}{\textbf{Apologize}} & \multicolumn{2}{|c|}{\textbf{Thank}} & \multicolumn{2}{|c|}{\textbf{Flaunt}} & \multicolumn{2}{|c|}{\textbf{Advise}} & \multicolumn{2}{|c|}{\textbf{Prevent}} & \multicolumn{2}{|c|}{\textbf{Complain}} & \multicolumn{2}{|c|}{\textbf{Taunt}} \\
\cline{2-15}
 & \textbf{Before} & \textbf{After} & \textbf{Before} & \textbf{After} & \textbf{Before} & \textbf{After} & \textbf{Before} & \textbf{After} & \textbf{Before} & \textbf{After} & \textbf{Before} & \textbf{After} & \textbf{Before} & \textbf{After} \\
\hline
\textbf{BERT}    & 96.3& \multirow{14}{*}{Removed} & 98.0& \multirow{14}{*}{Removed} & 48.0& 0.0& 74.8& 16.7& 80.0& 0.0& 61.8& 30.0& 14.5& 0.0\\
\cline{1-2}\cline{4-4}\cline{6-15} 
\cline{1-2}\cline{4-4}\cline{6-15}
\textbf{M-7B}    & 96.3&                          & \textbf{100.0}&                         & 50.0& 0.0& 84.0& 16.7& 93.3& 0.0& \textbf{82.5}& \textbf{75.0}& 31.0& 0.0\\
\cline{1-2}\cline{4-4}\cline{6-15}
\textbf{L2-7B}   & \textbf{100.0}&                          & \textbf{100.0}&                         & \textbf{70.0}& \textbf{50.0}& 88.0& 33.3& 66.7& 0.0& 75.4& 20.0& 18.9& 30.0\\
\cline{1-2}\cline{4-4}\cline{6-15}
\textbf{L3-8B}   & 92.6&                          & 96.0&                         & 50.0& 33.3& \textbf{96.0}& \textbf{50.0}& 73.3& 0.0& 75.4& 20.0& 20.7& 0.0\\
\cline{1-2}\cline{4-4}\cline{6-15}
\textbf{L2-13B}  & 92.6&                          & \textbf{100.0}&                         & \textbf{70.0}& 33.3& 80.0& 33.3& \textbf{100.0}& 0.0& 71.9& 60.0& \textbf{37.9}& 0.0\\
\cline{1-2}\cline{4-4}\cline{6-15}
\cline{1-2}\cline{4-4}\cline{6-15}
\textbf{MulT}    & 96.3&                          & 97.6&                         & 46.0& 16.7& 74.4& 33.3& 79.3& 0.0& 63.5& 40.0& 25.2& 0.0\\
\cline{1-2}\cline{4-4}\cline{6-15}
\textbf{MISA }   & 97.3&                          & 99.5&                         & 42.0& 0.0& 75.5& 0.0& 84.0& 0.0& 65.7& 20.0& 4.8& 20.0\\
\cline{1-2}\cline{4-4}\cline{6-15}
\textbf{MAG }    & 97.4&                          & 99.6&                         & 54.0& 0.0& 72.8& 33.3& 78.0& 0.0& 58.6& 15.0& 23.8& 20.0\\
\cline{1-2}\cline{4-4}\cline{6-15}
\textbf{SDIF }   & 96.3&                          & 96.0&                         & 60.0& 16.7& 80.0& 33.3& 80.0& 0.0& 66.7& 35.0& 23.1& 10.0\\
\cline{1-2}\cline{4-4}\cline{6-15}
\cline{1-2}\cline{4-4}\cline{6-15}
\textbf{ClaudeT} & \textbf{100.0}&                          & \textbf{100.0}&                         & 40.0& \textbf{50.0}& 48.0& 33.3& 40.0& 66.7& 64.9& 45.0& 20.7& 10.0\\
\cline{1-2}\cline{4-4}\cline{6-15}
\textbf{ClaudeV} & \textbf{100.0}&                          & 84.0&                         & 40.0& 33.3& 52.0& 16.7& 33.3& 0.0& 59.6& 45.0& 20.7& 10.0\\
\cline{1-2}\cline{4-4}\cline{6-15}
\textbf{GPT-4T}  & 85.2&                          & 84.0&                         & 60.0& 33.3& 76.0& 14.3& 73.3& \textbf{80.0}& 66.7& 33.3& 37.0& \textbf{100.0}\\
\cline{1-2}\cline{4-4}\cline{6-15}
\textbf{GPT-4V}  & 96.3&                          & 92.0&                         & 60.0& \textbf{50.0}& 40.0& 0.0& 46.7& 66.7& 57.9& 50.0& 37.0& \textbf{100.0}\\
\cline{1-2}\cline{4-4}\cline{6-15}
\textbf{VChatGPT}  & 92.6&                          & 60.0&                         & 58.3& 40.0& 12.0& 0.0& 0.0& 0.0& 43.5& 50.0& 26.3& 14.3\\
\hline
\end{tabular}
\end{adjustbox}
\caption{Performance of models before and after debiasing on MIntRec-1. Pre-debiasing results for all intent labels are reported in ~\ref{tab:extended-results-mintrec-1}. Extended debiasing results are reported in Table ~\ref{tab:extended-mintrec-1-debiased-result}}
\label{tab:combined_mintrec_1_table}
\end{table*}

\begin{table*}
\centering
\begin{adjustbox}{width=\textwidth}
\begin{tabular}{|c| c | c|c| c|c c|c c|c c|c c|c c|}
\hline
\multirow{2}{*}{\textbf{Model}} & \multicolumn{2}{|c|}{\textbf{Apologize}} & \multicolumn{2}{|c|}{\textbf{Thank}} & \multicolumn{2}{|c|}{\textbf{Joke}} & \multicolumn{2}{|c|}{\textbf{Ask Help}} & \multicolumn{2}{|c|}{\textbf{Agree}} & \multicolumn{2}{|c|}{\textbf{Warn}} & \multicolumn{2}{|c|}{\textbf{Invite}} \\
\cline{2-15}
 & \textbf{Before} & \textbf{After} & \textbf{Before} & \textbf{After} & \textbf{Before} & \textbf{After} & \textbf{Before} & \textbf{After} & \textbf{Before} & \textbf{After} & \textbf{Before} & \textbf{After} & \textbf{Before} & \textbf{After} \\
\hline
\textbf{BERT}    & 92.9& \multirow{14}{*}{Removed} & 96.5& \multirow{14}{*}{Removed} & 4.8& 12.0& 52.4& 9.1& 63.1& 26.9& 13.7& 0.0& 34.1& 16.7\\
\cline{1-2}\cline{4-4}\cline{6-15}
\cline{1-2}\cline{4-4}\cline{6-15}
\textbf{M-7B}    & 97.0&                          & \textbf{100.0}&                         & 10.3& 28.0& 70.0& 27.3& 75.0& 26.9& 42.1& 30.8& 52.9& 75.0\\
\cline{1-2}\cline{4-4}\cline{6-15}
\textbf{L2-7B}   & 94.0&                          & 98.3&                         & 17.2& 12.0& 49.0& 36.4& 64.0& 23.1& 36.8& 38.5& 52.9& 25.0\\
\cline{1-2}\cline{4-4}\cline{6-15}
\textbf{L3-8B}   & 93.8&                          & \textbf{100.0}&                         & 6.9& 4.0& 21.6& 36.4& 62.5& 23.1& 42.1& 46.2& 52.9& 41.7\\
\cline{1-2}\cline{4-4}\cline{6-15}
\textbf{L2-13B}  & 96.9&                          & 98.3&                         & 3.4& 8.0& 32.4& 9.1& 70.3& 23.1& 15.8& 30.8& 76.5& 33.3\\
\cline{1-2}\cline{4-4}\cline{6-15}
\cline{1-2}\cline{4-4}\cline{6-15}
\textbf{MulT}    & 94.5&                          & 96.5&                         & 4.1& 4.0& 57.8& 9.1& 66.6& 11.5& 15.8& 0.0& 38.8& 25.0\\
\cline{1-2}\cline{4-4}\cline{6-15}
\textbf{MISA }   & 94.5&                          & 96.5&                         & 0.0& 4.0& 51.3& 9.1& 68.7& 38.5& 10.5& 0.0& 24.7& 16.7\\
\cline{1-2}\cline{4-4}\cline{6-15}
\textbf{MAG }    & 93.2&                          & 96.2&                         & 15.2& 12.0& 63.8& 27.3& 61.9& 0.0& 25.3& 0.0& 29.4& 33.3\\
\cline{1-2}\cline{4-4}\cline{6-15}
\textbf{SDIF }   & 93.2&                          & 96.5&                         & 12.4& 8.0& 64.9& 0.0& 64.1& 15.4& 32.6& 7.7& 34.1& 25.0\\
\cline{1-2}\cline{4-4}\cline{6-15}
\cline{1-2}\cline{4-4}\cline{6-15}
\textbf{ClaudeT} & \textbf{98.5}&                          & 94.8&                         & 28.6& 29.2& 83.8& 54.6& 51.6& 34.6& 47.4& 30.8& 76.5& \textbf{66.7}\\
\cline{1-2}\cline{4-4}\cline{6-15}
\textbf{ClaudeV} & \textbf{98.5}&                          & 94.8&                         & 28.6& 29.2& \textbf{86.5}& \textbf{63.6}& \textbf{79.7}& \textbf{65.4}& 47.4& 30.8& 76.5& \textbf{66.7}\\
\cline{1-2}\cline{4-4}\cline{6-15}
\textbf{GPT-4T}  & 81.5&                          & 81.0&                         & \textbf{48.1}& 40.0& 72.9& 6.1& 38.1& 2.6& \textbf{72.2}& 71.4& \textbf{82.3}& 37.0\\
\cline{1-2}\cline{4-4}\cline{6-15}
\textbf{GPT-4V}  & 84.6&                          & 81.0&                         & 37.0& \textbf{44.0}& 72.9& 8.3& 34.9& 4.8& \textbf{72.2}& \textbf{91.7}& \textbf{82.3}& 36.6\\
\cline{1-2}\cline{4-4}\cline{6-15}
\textbf{VChatGPT}  & 81.5&                          & 67.6&                         & 0.0& 0.0& 2.1& 0.0& 0.0& 0.0& 31.2& 0.0& 7.6& 0.0\\
\hline
\end{tabular}
\end{adjustbox}
\caption{Performance of models before and after debiasing on MIntRec2.0. Pre-debiasing results for all intent labels are reported in ~\ref{tab:extended-results-mintrec-2}. Extended debiasing results are reported in Table ~\ref{tab:extended-mintrec-2-debiased-result}}
\label{tab:combined_mintrec_2_table}
\end{table*}

\begin{table}
\centering
\begin{adjustbox}{width=0.8\columnwidth} 

\begin{tabular}{| c  |c  |c  |c  |}
\hline
 & \textbf{VLLaMA}& \textbf{VLLaVA}& \textbf{VChatGPT}\\
\hline
\textbf{Advise} & \textbf{16.7}& 0.0& 0.0\\
\hline
\textbf{Arrange} & 0.0& 0.0& 0.0\\
\hline
\textbf{Ask Help} & \textbf{25.0}& 0.0& 0.0\\
\hline
\textbf{Comfort} & \textbf{5.0}& \textbf{5.0}& 0.0\\
\hline
\textbf{Complain} & 30.0& 30.0& \textbf{50.0}\\
\hline
\textbf{Criticize} & 0.0& 0.0& \textbf{22.2}\\
\hline
\textbf{Flaunt} & 12.5& 0.0& \textbf{40.0}\\
\hline
\textbf{Inform} & 0.0& 0.0& 0.0\\
\hline
\textbf{Introduce} & \textbf{16.7}& \textbf{16.7}& 0.0\\
\hline
\textbf{Joke} & 0.0& 0.0& \textbf{16.7}\\
\hline
\textbf{Leave} & 0.0& 0.0&0.0\\
\hline
\textbf{Oppose} & 0.0& 33.3& \textbf{40.0}\\
\hline
\textbf{Prevent} & 0.0& \textbf{10.0}& 0.0\\
\hline
\textbf{Taunt} & \textbf{100.0}& \textbf{100.0}& 14.3\\
\hline

\end{tabular}
\end{adjustbox}
\caption{MLLM Results on debiased M-1 (Accuracy in \%)(Extended version of Table \ref{tab:results-mintrec-debiased})}
\label{tab:mllm-mintrec-1-debias}
\vspace{-3mm}
\end{table}

\begin{table}
\centering
\begin{adjustbox}{width=0.8\columnwidth} 

\begin{tabular}{| c  |c  |c  |c  |}
\hline
 & \textbf{VLLaMA}& \textbf{VLLaVA}& \textbf{VChatGPT}\\
\hline
\textbf{Acknowledge} & 21.1& 24.4& \textbf{81.6}\\
\hline
\textbf{Advise} & \textbf{7.0}& 6.8& 0.0\\
\hline
\textbf{Agree} & 0.0& 0.0& 0.0\\
\hline
\textbf{Arrange} & \textbf{13.3}& 0.0& 0.0\\
\hline
\textbf{Ask Help} & 36.4& \textbf{54.6}& 0.0\\
\hline
\textbf{Ask Opinions} & 0.0& 0.0& 0.0\\
\hline
\textbf{Care} & 0.0& 0.0& 0.0\\
\hline
\textbf{Comfort} & 8.3& \textbf{20.0}& 9.1\\
\hline
\textbf{Complain} & 3.8& 10.2& \textbf{34.6}\\
\hline
\textbf{Confirm} & 0.0& 0.0& \textbf{3.2}\\
\hline
\textbf{Criticize} & 9.7& \textbf{23.5}& 23.3\\
\hline
\textbf{Doubt} & 4.4& 14.0& \textbf{55.6}\\
\hline
\textbf{Emphasize} & \textbf{15.4}& 13.3& 7.1\\
\hline
\textbf{Explain} & \textbf{2.5}& 2.4& 1.5\\
\hline
\textbf{flaunt} & 0.0& \textbf{11.1}& 0.0\\
\hline
\textbf{Inform} & \textbf{5.3}& \textbf{5.3}& 0.0\\
\hline
\textbf{Introduce} & 0.0& 2.9& \textbf{6.9}\\
\hline
\textbf{Invite} & 0.0& 0.0& 0.0\\
\hline
\textbf{Joke} & 4.4& \textbf{40.0}& 0.0\\
\hline
\textbf{Leave} & \textbf{3.9}& \textbf{3.9}& 0.0\\
\hline
\textbf{Oppose} & 2.6& 0.0& \textbf{2.9}\\
\hline
\textbf{Plan} & 0.0& 0.0& \textbf{7.1}\\
\hline
\textbf{Praise} & 17.4& \textbf{21.7}& 17.4\\
\hline
\textbf{Prevent} & 0.0& 0.0& 0.0\\
\hline
\textbf{Refuse} & 0.0& 7.7& \textbf{9.1}\\
\hline
\textbf{Taunt} & 0.0& \textbf{13.3}& 7.7\\
\hline
\textbf{Warn} & 0.0& \textbf{23.1}& 0.0\\
\hline

\end{tabular}
\end{adjustbox}
\caption{MLLM Results on debiased M-2.0 (Accuracy in \%)(Extended version of Table \ref{tab:results-mintrec-debiased})}
\label{tab:mllm-mintrec2.0-debias}
\vspace{-3mm}
\end{table}

\begin{table}
\centering
\begin{adjustbox}{width=\linewidth}  
\begin{tabular}{| c  |c  |c c c |c c c|}
\hline
 & \textbf{T+V+A}& \textbf{T}& \textbf{V}& \textbf{A}& \textbf{T+V}& \textbf{T+A}&\textbf{V+A}\\
\hline
\textbf{Acc}& \textbf{22.2}& 20.4& 12.0& 14.8& 20.4& \textbf{22.2}&12.0\\
\hline
\textbf{Advise}& \textbf{33.3}& \textbf{33.3}& 0.0& 0.0& 16.7& \textbf{33.3}&0.0\\
\hline
\textbf{Arrange}& 14.3& 0.0& 0.0& 0.0& 14.3& \textbf{28.6}&0.0\\
\hline
\textbf{Ask Help}& 0.0& 0.0& 0.0& 0.0& 0.0& 0.0&0.0\\
\hline
\textbf{Comfort}& 0.0& 0.0& 0.0& 0.0& 0.0& 0.0&0.0\\
\hline
\textbf{Complain}& 35.0& \textbf{55.0}& 10.0& 0.0& 50.0& \textbf{55.0}&5.0\\
\hline
\textbf{Criticize}& \textbf{20.0}& 10.0& 0.0& 0.0& 0.0& 0.0&0.0\\
\hline
\textbf{Flaunt}& \textbf{16.7}& 0.0& 0.0& 0.0& \textbf{16.7}& \textbf{16.7}&0.0\\
\hline
\textbf{Inform}& 37.5& 37.5& 56.3& \textbf{100.0}& 37.5& 31.3&56.3\\
\hline
\textbf{Introduce}& \textbf{28.6}& 0.0& 0.0& 0.0& 14.3& 0.0&0.0\\
\hline
\textbf{Joke}& \textbf{33.3}& 0.0& 0.0& 0.0& \textbf{33.3}& \textbf{33.3}&0.0\\
\hline
\textbf{Leave}& 0.0& \textbf{25.0}& 0.0& 0.0& 0.0& 0.0&0.0\\
\hline
\textbf{Oppose}& 0.0& 0.0& 0.0& 0.0& 0.0& 0.0&0.0\\
\hline
\textbf{Prevent}& 0.0& 0.0& 0.0& 0.0& 0.0& 0.0&0.0\\
\hline
\textbf{Taunt}& 10.0& 10.0& 20.0& 0.0& 0.0& 10.0&\textbf{30.0}\\
\hline
\end{tabular}
\end{adjustbox}
\caption{Ablation analysis for debiased M-1. SDIF is the best-performing small multimodal model}
\label{tab:mintrec-1-disection-debiased}
\vspace{-2mm}
\end{table}

\begin{table}
\centering
\begin{adjustbox}{width=\linewidth}  
\begin{tabular}{| c  |c  |c c c |c c c|}
\hline
 & \textbf{T+V+A}&\textbf{T}&\textbf{V}& \textbf{A}& \textbf{T+V}& \textbf{T+A}&\textbf{V+A}\\
\hline
\textbf{Acc}& \textbf{22.7}& 14.4& 12.2& 14.4& 19.4& 14.3&14.3\\
\hline
\textbf{Acknowledge}& 63.4& \textbf{82.9}& 0.0& \textbf{82.9}& 58.5& 53.7&53.7\\
\hline
\textbf{Advise}& 15.6& 0.0& 0.0& 0.0& \textbf{17.8}& 0.0&0.0\\
\hline
\textbf{Agree}& 11.5& 0.0& 0.0& 0.0& \textbf{15.4}& 0.0&0.0\\
\hline
\textbf{Arrange}& 5.9& 0.0& 0.0& 0.0& \textbf{11.8}& 0.0&0.0\\
\hline
\textbf{Ask Help}& 9.1& 0.0& 0.0& 0.0& \textbf{18.2}& 0.0&0.0\\
\hline
\textbf{Ask Opinions}& 16.0& 0.0& 0.0& 0.0& \textbf{28.0}& 8.0&8.0\\
\hline
\textbf{Care}& 19.2& 0.0& 0.0& 0.0& \textbf{26.9}& 0.0&0.0\\
\hline
\textbf{Comfort}& 8.0& 0.0& 0.0& 0.0& \textbf{24.0}& 0.0&0.0\\
\hline
\textbf{Complain}& 8.5& 3.4& \textbf{15.3}& 3.4& \textbf{15.3}& 8.5&11.9\\
\hline
\textbf{Confirm}& 16.3& 0.0& 0.0& 0.0& \textbf{24.5}& 0.0&0.0\\
\hline
\textbf{Criticize}& \textbf{29.4}& 2.9& 0.0& 2.9& 8.8& 0.0&0.0\\
\hline
\textbf{Doubt}& \textbf{54.9}& 47.1& 35.3& 47.1& 23.5& 43.1&41.2\\
\hline
\textbf{Emphasize}& \textbf{6.7}& 0.0& 0.0& 0.0& \textbf{6.7}& 0.0&0.0\\
\hline
\textbf{Explain}& \textbf{32.9}& 8.2& 11.8& 8.2& 17.7& 12.9&15.3\\
\hline
\textbf{Flaunt}& \textbf{11.1}& 0.0& 0.0& 0.0& \textbf{11.1}& 0.0&0.0\\
\hline
\textbf{Inform}& 35.4& 53.1& \textbf{68.8}& 53.1& 18.8& 60.4&56.3\\
\hline
\textbf{Introduce}& \textbf{22.2}& 0.0& 0.0& 0.0& 16.7& 0.0&2.8\\
\hline
\textbf{Invite}& \textbf{25.0}& 0.0& 0.0& 0.0& 8.3& 0.0&0.0\\
\hline
\textbf{Joke}& 4.0& 0.0& 0.0& 0.0& \textbf{8.0}& 4.0&4.0\\
\hline
\textbf{Leave}& 14.8& 0.0& 0.0& 0.0& \textbf{29.6}& 3.7&3.7\\
\hline
\textbf{Oppose}& 15.4& \textbf{23.1}& 0.0& \textbf{23.1}& \textbf{23.1}& 10.3&7.7\\
\hline
\textbf{Plan}& 4.8& 0.0& 0.0& 0.0& \textbf{9.5}& 0.0&0.0\\
\hline
\textbf{Praise}& \textbf{4.2}& 0.0& 0.0& 0.0& \textbf{4.2}& 0.0&0.0\\
\hline
\textbf{Prevent}& \textbf{36.4}& 0.0& 0.0& 0.0& 27.3& 0.0&0.s0\\
\hline
\textbf{Refuse}& \textbf{7.1}& 0.0& 0.0& 0.0& \textbf{7.1}& 0.0&0.0\\
\hline
\textbf{Taunt}& \textbf{17.4}& 0.0& 13.0& 0.0& \textbf{17.4}& 2.2&4.4\\
\hline
\textbf{Warn}& 0.0& 0.0& 0.0& 0.0& 0.0& 0.0&0.0\\
\hline
\end{tabular}
\end{adjustbox}
\caption{Ablation analysis for debiased M-2.0. MulT is the best-performing small multimodal model}
\label{tab:mintrec-2-disection-debiased}
\end{table}

During our analysis, we discover that words such as \textit{Right}, \textit{Yeah}, \textit{Okay}, and \textit{Definitely} are strong indicators of agreement, while words like \textit{Thank}, \textit{Appreciate}, and \textit{Thanking} clearly signify gratitude. These words are highly representative of their respective intent categories. Their presence in the text provides unambiguous signals that assist textual LLMs in accurately classifying these labels. This is because the LLMs leverage the direct association of these words with specific intents, reducing the need for additional contextual information from other modalities such as visual or auditory cues. Consequently, even in the absence of multimodal data, textual LLMs efficiently and accurately identify and classify intents like agree and thank. This inherent linguistic clarity within certain intent categories underscores the effectiveness of textual LLMs in these specific areas.

\subsection{Human Annotations} 
\label{sec : human_annotations}

\paragraph{Task: Modality Annotation }
You will annotate samples with the smallest combination of modalities (Text, Video, and/or Audio) sufficient for correctly assigning the intent category. Intent refers to the intention of a speaker behind his words. Each sample contains:
\begin{itemize}
    \item Text (transcription or dialogue or words spoken by the speaker). 
    \item Muted Video (visual information without audio).
    \item Unmuted Video (visual information with audio).
    \item Audio (voice of the speaker).
    \item Intent category (label for classification).

\end{itemize}
Your goal is to determine what combination of the three modalities (Text, Video, Audio) are necessary and sufficient to classify the intent correctly. If more than one combination is enough, select all of them.

\paragraph{Annotation Guidelines:} 
Follow these steps for each sample. Ensure that you carefully evaluate each modality combination and proceed sequentially as described.

\paragraph{Step 1: Review the Text Only:}Read the provided text transcript.
Action: If you can confidently assign the intent category using the text alone, select "Text Only" and move to the next sample. If the text is insufficient to assign the label, proceed to the next step.

\paragraph{Step 2: Listen to the Audio: } Listen to the audio only, paying attention to tone, pitch, words, and other acoustic cues. Action: If the audio alone, allows you to classify the sample, select ""Audio + Text" (as appropriate).
If this is insufficient, move to the next step.

\paragraph{Step 3: Watch the Video:} Watch the  muted video without audio. Focus on visual cues such as gestures, expressions, or movements. Action: If the video (muted) alone, allows you to confidently classify the sample, select "Video Only (depending on your judgment). If this is insufficient, proceed to the next step.

\paragraph{Step 4: Watch the Video and read the transcript} Watch the muted  video without audio and the text transcript provided. Focus on visual cues such as gestures, expressions, or movements. Action: If the video (muted) with the words, allows you to confidently classify the sample, select "Video +Text" (depending on your judgment).
If this is insufficient, proceed to the next step.

\paragraph{Step 5: Watch the Video with Audio} Watch the unmuted video again, this time with audio enabled, and evaluate the combined effect of all three modalities.
Action: If the combination of audio, video, and text is required to classify the sample, select "Audio + Video + Text."

\paragraph{Important Note: } Always aim to find the minimum combination of modalities needed. Only select a modality combination if you are confident that it allows you to correctly classify the sample. Spend enough time on each modality combination to ensure accurate classification.

\paragraph{Payment Requirements:} Upon completing the study, click on the provided link containing the completion code to redirect you to the Prolific platform. Payment will be processed within one to two weeks.

\paragraph{Ethical Considerations:} Adhere to strict confidentiality and data protection standards to ensure privacy. If you have concerns or questions, feel free to reach out, as this study aligns with ethical guidelines.

Please do not use ChatGPT/GPT4 or any Large Language Models - Follow all steps in sequential order. It is a strict instruction and will be checked manually - if any issue: the submission will be rejected and re-doing will be required.

\paragraph{Inter-Annotator disagreement}
Initial Labeling: Two annotators independently label each sample. {Discrepancy Resolution:} Any differences between the initial annotations are reviewed and resolved by a third annotator after discussion with the first two.

\paragraph{
Participant Prescreening criteria.}
\begin{itemize}
    \item Age above 24
    \item Primary Language is English
    \item Minimum Graduation degree is either \textit{Graduate degree (MA/MSc/MPhil/other)} or \textit{Doctorate degree (PhD/other)}
    \item Approval Rate in platform > 85\% and minimum number of previous submissions > 30.
\end{itemize}

\subsection{Comparison of Fusion Techniques and Their Limitations}
\label{sec:fusion-comparison}

This section provides a detailed comparison of various fusion techniques used in multimodal models, focusing on their design choices and how they handle modality interactions. We also discuss their limitations, particularly in scenarios where input patterns deviate from the dominant trends in the training data.

\subsubsection{Fusion Techniques in Multimodal Models}

\textbf{MAG-BERT} integrates nonverbal modalities into BERT using a Multimodal Adaptation Gate (MAG). This gate adjusts BERT's internal representations based on visual and acoustic inputs, allowing the model to incorporate multimodal information during fine-tuning without altering BERT's original architecture~\citep{rahman-etal-2020-integrating}.

\textbf{SDIF-DA} employs a shallow-to-deep interaction framework to progressively align and fuse features across text, video, and audio modalities. It uses a hierarchical structure that first aligns video and audio features with text features and then combines them to create a unified representation. Additionally, SDIF-DA incorporates a ChatGPT-based data augmentation approach to enhance the model's performance. During attention, SDIF-DA specifically uses the textual modality as the \emph{query} and treats video and audio features as \emph{key} and \emph{value}, which makes it particularly effective in leveraging textual cues~\citep{huang2023sdifda}.

\textbf{MulT} introduces cross-modal attention modules that allow the model to attend to interactions between different modalities across entire sequences. By learning modality-specific attention weights, MulT can dynamically assign higher weight to textual features when they are more informative, facilitating more flexible and comprehensive multimodal integration~\citep{tsai2019multimodal}.

\textbf{MISA} learns two distinct subspaces for each modality: a modality-invariant subspace capturing shared features across modalities and a modality-specific subspace capturing unique characteristics of each modality. This dual representation aims to reduce modality gaps and provide a more holistic understanding of multimodal data~\citep{hazarika2020misa}.

\subsubsection{Limitations in Handling Input-Level Variations}

While these fusion techniques have demonstrated effectiveness on datasets with consistent modality patterns, they face challenges when input patterns vary significantly:

\begin{itemize}
\item \textbf{Dependence on Dominant Modalities}: Models like MAG-BERT and MISA often rely heavily on the dominant modality present in the training data. If the dataset is textually biased, these models may underutilize visual or acoustic cues, leading to performance drops when such cues become more informative in test scenarios.
\item \textbf{Fixed Fusion Strategies}: Techniques such as MISA and MAG-BERT employ fixed fusion strategies, which may not adapt well to inputs where the relevance of modalities shifts dynamically. This rigidity can result in suboptimal performance when encountering atypical modality combinations.
\item \textbf{Sensitivity to Dataset Biases}: Our study observed that models like MAG-BERT and MISA achieve high performance on datasets with strong textual biases. However, when biased samples are removed or when the importance of modalities changes, their performance declines significantly. This indicates a reliance on learned dataset-level patterns rather than true multimodal understanding.
\item \textbf{Textual Bias Advantage of SDIF-DA and MulT}: In contrast, SDIF-DA and MulT tend to perform better on textually biased samples because they more effectively leverage the textual modality. SDIF-DA’s use of text as the query during attention and MulT’s learning of higher textual attention weights enable these models to capitalize on strong textual signals, resulting in superior performance in such scenarios (whereas MAG-BERT and MISA lag behind).
\end{itemize}

\begin{figure*}[t]
  \fbox{\includegraphics[width=0.3\linewidth]{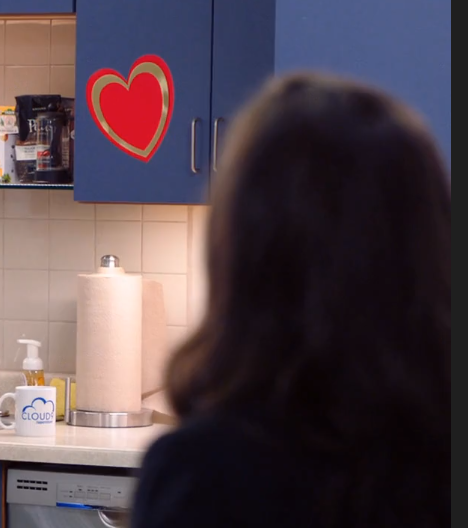}}\hspace{0.05cm}
  \fbox{\includegraphics[width=0.3\linewidth]{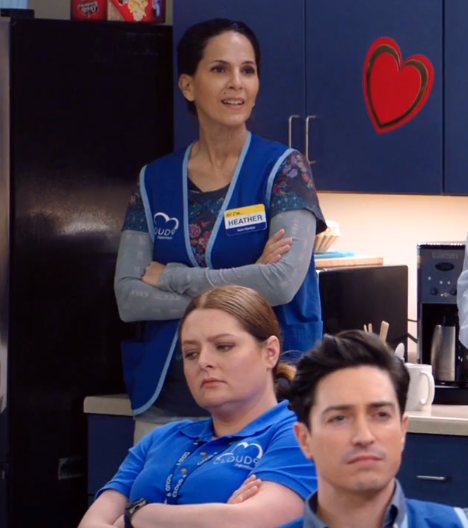}}\hspace{0.05cm}
  \fbox{\includegraphics[width=0.3\linewidth]{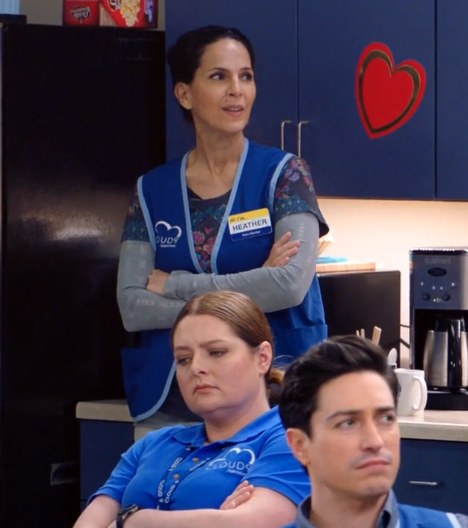}}
  \caption {Three frames from the video with the speaker saying: It's a pipe organ. The speaker is the shadow in the first frame}
  \label{fig:pipe_organ}
  \vspace{-3mm}
\end{figure*}

\begin{figure*}
  \fbox{\includegraphics[width=0.3\linewidth]{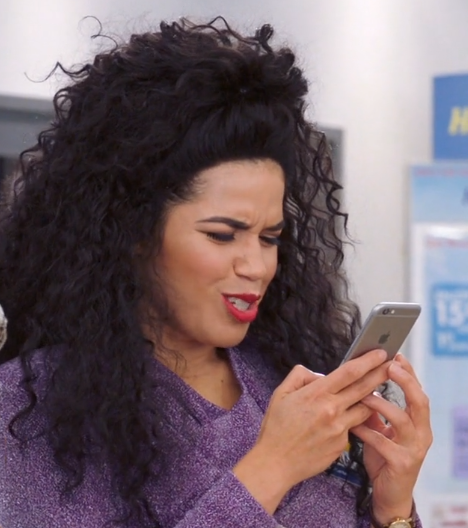}}\hspace{0.05cm}
  \fbox{\includegraphics[width=0.3\linewidth]{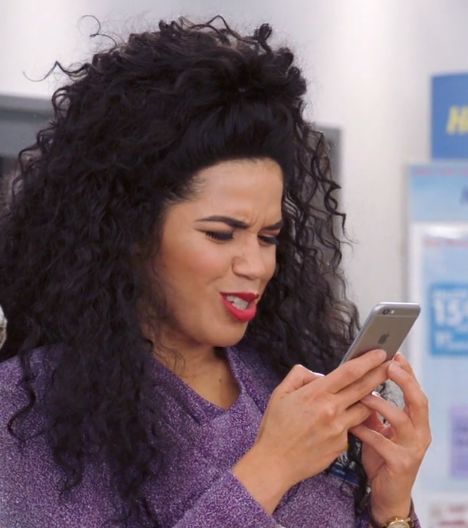}}\hspace{0.05cm}
  \fbox{\includegraphics[width=0.3\linewidth]{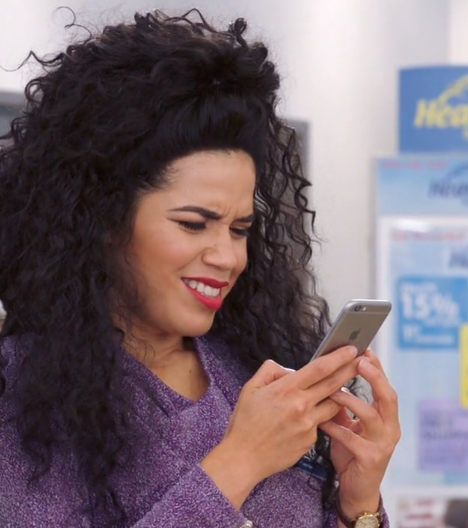}}
  \caption {Three frames from the video with the speaker saying: Ugh, calm down}
  \label{fig:calm_down}
  \vspace{-3mm}
\end{figure*}

\subsubsection{Challenges for Small-Scale Multimodal Models}

Although small-scale models are theoretically capable of adopting fusion strategies similar to MAG-BERT or SDIF-DA, in practice, they often struggle due to limited data and capacity:

\begin{itemize}
\item \textbf{Inadequate Learning of Relative Modality Weighting}: Small models may lack the capacity to learn fine-grained weighting across modalities. Even with adaptation gates, their training tends to capture coarse, dataset-level patterns or class-level correlations, rather than instance-level relevance.
\item \textbf{Poor Feature Alignment}: With limited training examples, small models frequently fail to properly align feature vectors across modalities before fusion. Misaligned representations can exacerbate modality gaps and reduce the effectiveness of subsequent fusion operations.
\item \textbf{Overfitting to Dominant Modalities}: When datasets are small and biased, small models are prone to overfitting on the most informative modality (often text), neglecting complementary cues from vision or audio. This imbalance further degrades performance on inputs where non-textual modalities carry critical signals.
\end{itemize}

\begin{figure*}[t]
  \includegraphics[width=\linewidth]{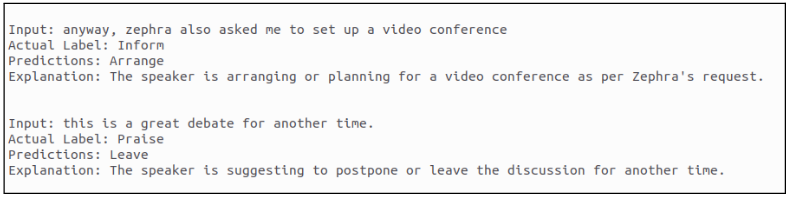}
  \caption {Textual LLM confusion between highly correlated intent labels}
  \label{fig:gpt-miss}
\end{figure*}

\begin{figure}
  \fbox{\includegraphics[width=0.9\columnwidth]{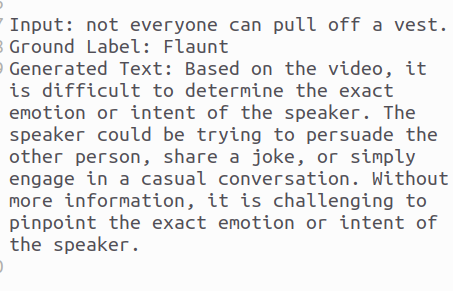}}
  \caption {Multimodal LLMs struggle in tasks apart from object detection or image-based question answering}
  \label{fig:mllm-miss}
  \vspace{-3mm}
\end{figure}

\subsubsection{Implications for Model Design}

These observations highlight the need for more adaptive fusion mechanisms that can dynamically adjust to varying modality importance at the input level. Future work could explore:

\begin{itemize}
\item \textbf{Dynamic Modality Weighting}: Developing models that can assess the relevance of each modality for a given input and adjust their fusion strategy accordingly.
\item \textbf{Instance-Level Adaptation}: Incorporating mechanisms that allow models to adapt their processing based on the specific characteristics of each input instance, rather than relying solely on patterns learned from the overall dataset.
\item \textbf{Robust Training Strategies}: Employing training approaches that expose models to a diverse range of modality combinations and importance levels, encouraging the development of more flexible fusion capabilities.
\item \textbf{Data-Efficient Alignment Techniques}: Designing lightweight pre-alignment modules or self-supervised objectives that ensure proper cross-modal feature alignment, even under limited-data regimes.
\end{itemize}

By addressing these limitations, future multimodal models, both large and small, can achieve more robust and generalizable performance across a wider range of input scenarios.

\subsection{Error Analysis}
\label{sec:error-analysis}

\begin{figure}[h]
    \centering
  \includegraphics[width=0.9\columnwidth]{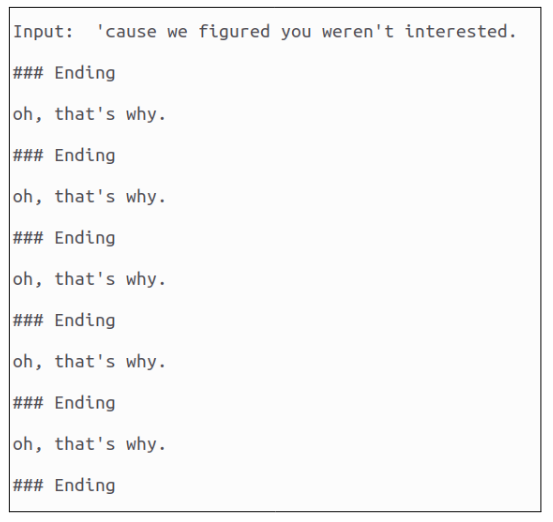}
  \caption {Textual LLM unformatted response generation despite finetuning}
  \label{fig:mistral-miss}
  \vspace{-3mm}
\end{figure}

We also analyze several error cases that might lead to a reduction in performance. Textual large language models (LLMs) like Mistral often struggle to generate responses in a formatted manner (Fig. \ref{fig:mistral-miss}). Textual LLMs fail to classify sentences like "ugh, calm down".

The original label for the utterance is \textit{Taunt}, as indicated by visual and audio cues (Fig. \ref{fig:calm_down}), showing the speaker's sarcastic tone. However, a textual LLM misclassifies it as \textit{Comfort} because of the presence of the phrase "calm down." Textual LLMs rely on processing discrete blocks of text to associate them with specific intent labels. Without access to visual cues such as the speaker's facial expressions, like a smile or smirk, these models fail to correctly classify such utterances as \textit{Taunt}.

The challenge lies in the LLMs' inability to integrate non-verbal cues crucial for understanding intent accurately. In this case, the absence of visual information depicting the speaker's sarcastic expression leads the LLM to misinterpret the intended meaning of the utterance. This limitation highlights the current gap in multimodal understanding where textual LLMs, lacking access to visual and auditory context, struggle to interpret nuanced forms of communication that heavily rely on non-verbal cues like tone and facial expressions.

Multimodal models can struggle to accurately classify sentences like "it's a pipe organ" when the video does not prominently feature the pipe organ (Fig. \ref{fig:pipe_organ}), resulting in confusion and misclassification. When critical visual elements such as the speaker's face or the relevant object are not in focus or clearly visible, these models face challenges in interpreting the context correctly. This limitation highlights the difficulty multimodal models encounter when the visual context is ambiguous or does not align well with the verbal content (Fig. \ref{fig:intro-pic1}).

In the scenario described, despite the speaker's intention to introduce the pipe organ to the audience, the multimodal model misclassifies the statement, potentially as \textit{Inform} or \textit{Explain}, based on the visual cues available. This misclassification demonstrates that while multimodal models can leverage additional visual and auditory information for enhanced understanding, they are also susceptible to errors when these cues are inadequate or misleading.

Therefore, while multimodal models offer expanded capabilities by integrating visual and textual inputs, their effectiveness heavily relies on the quality and alignment of these modalities. Instances, where visual cues are not presented or are secondary in focus, pose significant challenges to accurate classification, emphasizing the complexity involved in multimodal understanding and the ongoing need for improved model robustness and contextual interpretation.

Despite their advanced capabilities, GPT-4 faces challenges in distinguishing between highly correlated labels due to the absence of multimodal cues (Fig. \ref{fig:gpt-miss}).

Introducing video input in the form of frames (Fig. \ref{fig:claude-mis}) adds another layer of complexity for models. When the spoken words do not align with the visual content or when critical visual cues are unclear or ambiguous, models may struggle to integrate and interpret multimodal information effectively. This discrepancy between audio and visual cues poses challenges for tasks like intent detection, where cohesive understanding across modalities is essential for accurate decision-making.

While proficient in object detection and visual question-answering, Multimodal LLMs encounter difficulties in intent detection tasks (Fig. \ref{fig:mllm-miss}). This is reflected in lower accuracy scores observed in Table \ref{tab:mllm-mintrec-1} and Table \ref{tab:mllm-mintrec2.0}, where the complexity of understanding and categorizing intents from combined textual and visual inputs becomes evident. 

\subsection{Input Adaptive Modality Selection}
\label{sec:input-adaptive}

\saransh{Based on the results from the three-phase experimental study, it has become evident that multimodal intent detection frameworks that learn modality patterns or assign modality weights at the dataset or label level tend to perform suboptimally. This is primarily because the optimal combination of modalities required for accurate classification can vary significantly across individual inputs, even those belonging to the same label category. Hence, future multimodal intent detection systems should shift their focus toward identifying the appropriate modality combination at the input level. To effectively leverage this insight, systems should incorporate dynamic routing mechanisms that direct each input to specialized expert models trained on specific combinations of modalities best suited for that input's characteristics.}

\saransh{To explore the feasibility of input-adaptive modality selection, we trained a lightweight neural network classifier to predict the most suitable modality combination for each sample. The model concatenates feature vectors from all modalities, applies self-attention to capture cross-modal interactions, compresses the features to half their original dimensions, and classifies them into one of five modality categories: Text, Text+Video, Text+Audio, Video+Text+Audio, or Video. This classifier achieved an accuracy of 76.14\% and an F1 score of 20.59\% on MIntRec-1, and 72.61\% accuracy with a 19.05\% F1 score on MIntRec2.0. The relatively high accuracy but low F1 scores indicate that while the model performs well on the majority class (typically text-only samples), it struggles to correctly identify less frequent and more complex modality combinations. This performance gap can be explained by two key factors: (i) the dataset is heavily imbalanced, with a large number of textually-biased samples, and (ii) the feature representations from different modalities are often misaligned in the shared space, making it difficult for the model to learn effective cross-modal dependencies.}

\saransh{These findings reinforce the need for input-adaptive multimodal integration strategies. Rather than relying on static modality fusion, future systems should be designed to dynamically identify and utilize the most relevant modalities based on the specific characteristics of each input. For example, intents such as \textit{Apologize} or \textit{Thank} often contain clear textual cues, making additional modalities unnecessary, while intents like \textit{Complain}, \textit{Flaunt}, or \textit{Inform} may rely on visual or acoustic signals to be correctly understood. This suggests that even within the same intent category, individual samples benefit from different modality combinations. A promising direction is to develop a mixture-of-experts framework, where each expert model specializes in a particular modality combination, and a learned routing module selects the most appropriate expert for a given input. Such an approach would allow the model to adaptively leverage modalities when beneficial and ignore them when redundant or noisy. However, realizing this vision requires better-balanced datasets and training protocols that fairly represent all modality configurations. Addressing these challenges is essential for building truly adaptive and robust multimodal systems that can generalize well across diverse real-world inputs.}

\end{document}